# A Hormetic Approach to the Value-Loading Problem: Preventing the Paperclip Apocalypse?

Nathan I. N. Henry[1*], Mangor Pedersen[1], Matt Williams[2], Jamin L. B. Martin[3], Liesje Donkin[1]

[1]*Department of Psychology and Neuroscience, Auckland University of Technology, 90 Akoranga Drive, Northcote, Auckland 0627, New Zealand*
[2]*School of Psychology, Massey University, Kell Drive, Albany, Auckland 0632, New Zealand*
[3]*School of Physical and Chemical Sciences, 20 Kirkwood Avenue, Upper Riccarton, Christchurch 8041, New Zealand*
[*]*nathan.henry@aut.ac.nz*

## 1 Abstract

The value-loading problem is a significant challenge for researchers aiming to create artificial intelligence (AI) systems that align with human values and preferences. This problem requires a method to define and regulate safe and optimal limits of AI behaviors. In this work, we propose HALO (Hormetic ALignment via Opponent processes), a regulatory paradigm that uses hormetic analysis to regulate the behavioral patterns of AI. Behavioral hormesis is a phenomenon where low frequencies of a behavior have beneficial effects, while high frequencies are harmful. By modeling behaviors as allostatic opponent processes, we can use either Behavioral Frequency Response Analysis (BFRA) or Behavioral Count Response Analysis (BCRA) to quantify the hormetic limits of repeatable behaviors. We demonstrate how HALO can solve the 'paperclip maximizer' scenario, a thought experiment where an unregulated AI tasked with making paperclips could end up converting all matter in the universe into paperclips. Our approach may be used to help create an evolving database of 'values' based on the hedonic calculus of repeatable behaviors with decreasing marginal utility. This positions HALO as a promising solution for the value-loading problem, which involves embedding human-aligned values into an AI system, and the weak-to-strong generalization problem, which explores whether weak models can supervise stronger models as they become more intelligent. Hence, HALO opens several research avenues that may lead to the development of a computational value system that allows an AI algorithm to learn whether the decisions it makes are right or wrong.

## 2 Introduction

### 2.1 Behavioral regulation in artificial intelligence

Artificial Intelligence (AI) algorithms are garnering considerable attention, as they have been demonstrated to both match and exceed human performance on several tasks (Dell'Acqua et al., 2023). Some researchers believe advancements in AI are progressing towards the eventual creation of agents with 'superintelligence', or intelligence that exceeds the capabilities of the best human minds in virtually all domains (Bostrom, 1998). Whether superintelligent systems are attainable, and how they would work in the real world, remains unknown. But the implications of such superintelligent systems are profound. Just as human intelligence has enabled the development of tools and strategies for unprecedented control over the environment, AI systems have the potential to wield significant power through autonomous tool and strategy development (Soares & Fallenstein, 2014). With this comes the risk of these systems performing tasks that may not align with humanity's goals and preferences. Hence, there is a need to perform 'alignment' on these systems, to ensure that the actions of advanced machine learning systems are directed towards the intended goals and values of humanity (Taylor et al., 2016).

Currently, there are two general approaches to aligning superintelligent AI with human preferences. The first is 'scalable oversight' – using more powerful supervisory AI models to regulate weaker AI models that may, in the future, outperform human skills (Bowman et al., 2022). The second is 'weak-to-strong generalization', where weaker machine learning models are used to train stronger models that can then generalize from the weaker models' labels (Burns et al., 2023). It is hoped that these techniques will allow superintelligence to self-improve both safely and recursively (Omohundro, 2007; Yudkowsky, 2007). But to achieve this, we must first solve the value-loading problem: how do we encode human-aligned values into AI systems, and what will those values be (Bostrom, 2014b)?

Reward modelling is an emerging technique aiming to solve the value-loading problem, by equipping agents with a reward signal that guides behavior toward desired outcomes. By optimizing this signal, agents act in ways congruent with human preferences (Leike et al., 2018). This assumes that our emotional neurochemistry evolved as a proxy reward function for behaviors that encourage growth, adaptation, and improvement of human wellbeing simultaneously (Kelley, 2005). However, this reward model is sub-optimal, producing negative externalities such as addiction (Kelley, 2005) due to cognitive biases like temporal discounting, in which immediate gains are favoured over long-term outcomes (Critchfield & Kollins, 2001; van den Bos & McClure, 2013). Hence, a nuanced reward model is needed to align AI behaviors with human emotional preferences, which we use in everyday life to help us judge between right and wrong (Damasio, 1994)[1]. Merely rewarding desired actions isn't sufficient; negative feedback must also be given when necessary. This is already performed in leading algorithms like GPT-4, which use Reinforcement Learning with Human Feedback (RLHF), combining reward-based reinforcement with corrective human input to improve the reward model when necessary (Christiano et al., 2017).

A further challenge to creating reward models for AI alignment is that behaviors are generally repeatable. Hence, the value of performing a behavior is affected by temporal influences such as the count and frequency of repetition of that behavior in recent history. For example, while eating food is essential for survival, a person who has recently consumed several slices of pizza should consider whether eating an additional piece will be harmful to their long-term health, despite the potential short-term pleasure. Therefore, an AI agent deciding whether to perform a behavior must rely on historical, short-term, long-term, and game-theoretic considerations.

---

[1] The judgement of 'right' and 'wrong' is also influenced by one's background, personality, culture, religion, and other factors. This poses a significant philosophical challenge for AI developers aiming to create unbiased reward models, given the diversity of human preferences worldwide.

To enable this type of decision making, we propose a reward modelling paradigm called HALO (Hormetic ALignment via Opponent processes). HALO enables us to quantify the healthy limits of repeatable behaviors, accounting for the temporal influences described above. We believe that HALO can be used to create AI models that are aligned with human emotional processing, while avoiding the traps that lead to sub-optimal human behaviors. Firstly, to describe this paradigm, we must explain some of its foundational concepts.

## 3 Background

### 3.1 Using behavioral posology for reward modeling

Behavioral posology is a paradigm we introduced to model the healthy limits of repeatable behaviors (Henry, Pedersen, Williams, & Donkin, 2023). By quantifying a behavior in terms of its potency, frequency, count, and duration, we can simulate the combined impact of repeated behaviors on human mental well-being, using pharmacokinetic/pharmacodynamic (PK/PD) modelling techniques for drug dosing (Henry et al. 2023). In turn, *insights derived from these models could theoretically be used to set healthy limits on repeatable AI behaviors.* This type of regulation has already been demonstrated in the context of machine learning recommendation systems, by using an allostatic model of opponent processes to prevent online echo chamber formation (Henry, Pedersen, Williams, Martin, et al., 2023).

In Solomon and Corbit's opponent process theory, humans respond to stimuli with a dual-phase psychological response, consisting of an initial positive a-process succeeded by a prolonged, less intense, and negative b-process (Solomon & Corbit, 1974). This occurs along multiple dimensions, including hedonic state, although other emotions also exhibit opponent process properties, such as anxiety, expectation, loneliness, grief, and relief (Solomon & Corbit, 1974). As an example, repeated opponent processes at a high frequency can cause hedonic allostasis, where accumulating b-processes shift one's hedonic set point away from homeostatic levels, potentially inducing a depressive state (Karin et al., 2021; Koob & Le Moal, 2001). Figure 1 illustrates this phenomenon. Allostasis serves as a regulatory mechanism, enabling the body to recalibrate during environmental and psychological challenges by adapting to and anticipating future demands (Katsumi et al., 2022; Sterling, 2012).

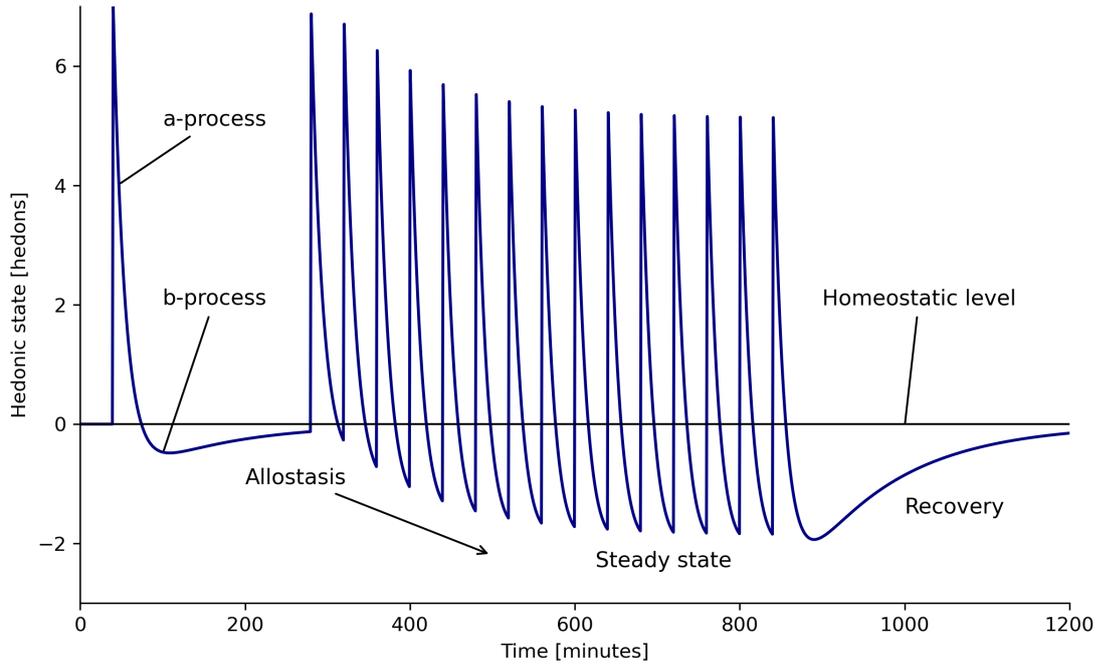

*Figure 1: PK/PD simulation of allostasis via repeated opponent processes, generated by behavioral repetition. An instance of the behavior is performed at time t=40. This is followed by rapid repetition of the behavior, causing allostasis due to summation of b-processes. An approximate steady-state is reached around t=600, followed by recovery to homeostatic baseline after the behavior ceases at t=840.*

## 3.2 The link between allostasis and hormesis

A growing body of biological research suggests that allostasis is linked to a phenomenon called hormesis (Li & He, 2009; McEwen & Wingfield, 2003; Sonmez et al., 2023). Hormesis is a dose-response relationship where low doses of a stimulus have a positive effect on the organism, while higher doses are harmful beyond a hormetic limit, also known as the NOAEL (No Observed Adverse Effect Level) (Agathokleous et al., 2021). This phenomenon occurs in many areas of nature, medicine, and psychology, and is also referred to as the Goldilocks zone, the U-shaped (or inverted U-shaped) curve, and the biphasic response curve (Calabrese & Baldwin, 2001; Przybylski & Weinstein, 2017). For example, moderate coffee consumption is known to improve cognitive performance in the short-term (Jarvis, 1993; Sargent et al., 2020), but excessive consumption may lead to dependency and withdrawal symptoms (Min et al., 2023; Zhu et al., 2023). A dose-response analysis of 12 observational studies identified a hormetic relationship between coffee consumption and risk of depression, with a decreased risk of depression for consumption up to 600 mL/day, and an increased risk above 600 mL/day (Grosso et al., 2016). Yet this phenomenon also appears to occur in some behaviors. For instance, moderate use of digital technologies (such as social media) may have social and mental benefits, but excessive use may lead to symptoms of behavioral addiction (Ho et al., 2014; Przybylski & Weinstein, 2017; Zhang et al., 2023).

Henry et al. have shown via PK/PD modeling that under certain conditions, frequency-based hormesis may be generated from allostatic opponent processes delivered at varying frequencies (Henry, Pedersen, Williams, & Donkin, 2023). The intriguing implication is that certain behaviors exhibit positive effects when practiced at lower frequencies, but harmful effects at higher frequencies. It is plausible that all behaviors have a frequency-based hormetic limit. This appears to be true even for positive behaviors such as generosity, which has game theoretic advantages for all agents in repeated interactions, as it encourages reciprocity and mutual growth (Delton et al., 2011). However, if an agent is overly generous, they will eventually run out of resources to donate. Therefore, in theory, there is a hormetic limit for generosity that shouldn't be exceeded. Even behavior as positive as laughter can be fatal in excess (Topno & Thakurmani, 2020).

In theory, behavioral posology can be used to quantify the hormetic limit for behaviors that cause allostatic opponent processes, when combined with longitudinal observational data (Henry, Pedersen, Williams, & Donkin, 2023). This may also help to define the moral limits of 'grey' behaviors, which have both positive and negative aspects. However, defining these hormetic limits is challenging, especially when considering the cumulative effects of repeated behavioral doses in both the short- and long-term, such as sensitization, habituation, tolerance, and addiction. Yet if we can quantify these hormetic limits in different contexts, this could be used as a framework for building a value system that keeps the AI within these hormetic limits.

### 3.3 The law of diminishing marginal utility

The 'paperclip maximizer' problem serves as a cautionary tale illustrating the perils of a misaligned AI. In this scenario, an AI tasked with maximizing paperclip production without constraints converts all matter, including living beings, into paperclips, resulting in global devastation (Bostrom, 2014a). This scenario underscores that an AI, even with benign intentions, can become 'addicted' to harmful behaviors if its reward model is incorrectly specified.

An understanding of behavioral economics is crucial for AI agents (such as the paperclip producing agent) to navigate complex decision-making processes effectively. Essential to this understanding are the concepts of Total Utility ($TU$) and Marginal Utility ($MU$) (Smith, 1776). $TU$ is defined as the overall satisfaction or benefit experienced by the consumer of a product or service, accounting for diverse factors like product quality, timing, and psychological appeal. $MU$, on the other hand, measures the added satisfaction from consuming an extra unit of a product or service. The relationship between $MU$ and $TU$ tends to follow the law of diminishing $MU$, asserting that as consumption rises, the incremental satisfaction per unit diminishes (Marshall, 1890). This law is demonstrated in Figure 2. The Relative Marginal Utility ($RMU$) represents the change in $MU$ compared to $MU_{initial}$, the value of $MU$ at $n = 0$. Hence, $RMU$ starts at a value of 0 and decreases as $n$ increases.

Intriguingly, the law of diminishing $MU$ can be considered a form of hormesis, assuming that $MU$ continues to decrease linearly after becoming negative (Szarek, 2005). Figure 2 illustrates that beyond the point of maximum $TU$, humans tend to cease their consumption of a product as its marginal utility becomes increasingly negative. Imagine an office worker for whom the ideal quantity of paperclips is five, as depicted in Figure 2. Beyond this threshold, the utility of extra paperclips diminishes; they serve no purpose and impose storage costs. Further, the worker incurs unnecessary expenses for producing these surplus clips. A rational worker would stop acquiring more paperclips upon recognizing the decline in their $MU$. However, imagine a person who due to a strong hoarding compulsion, continues to acquire paperclips even beyond the point where $MU$ has become negative. Similarly, a misaligned AI agent, exemplified by a paperclip maximizer, could persist in creating paperclips for its owner forever, despite negative outcomes that eclipse initial benefits. Taken far enough, such an agent could cause significant damage to the environment and humanity in its pursuit of creating paperclips.

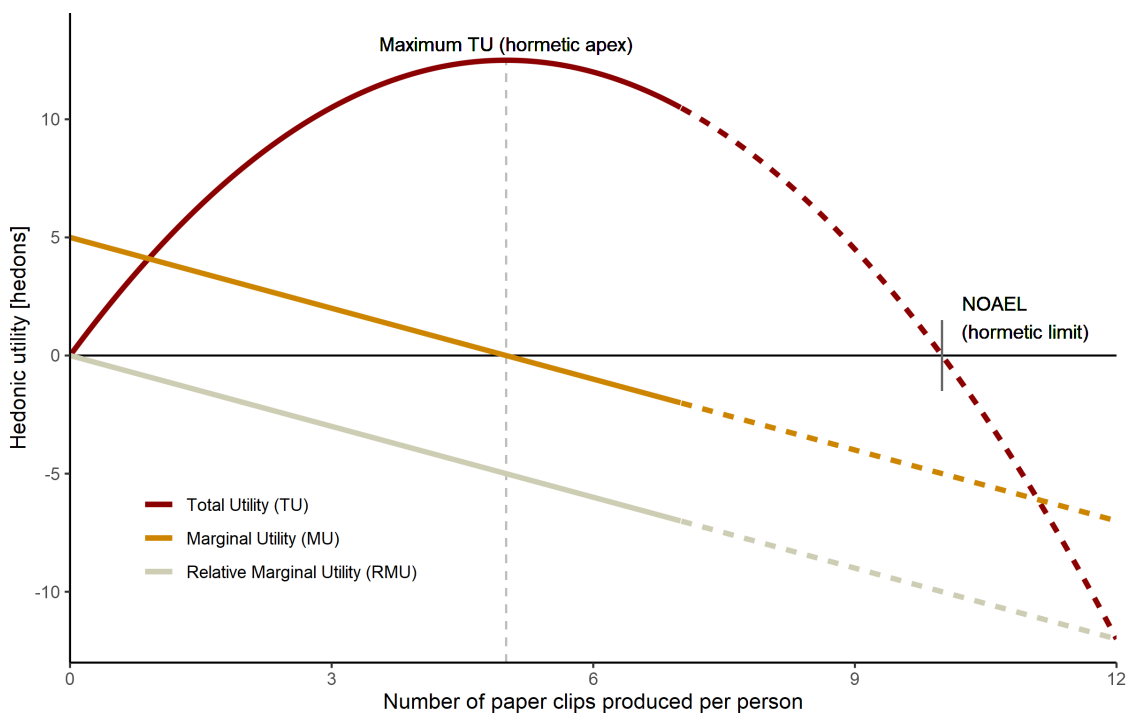

*Figure 2: Illustration of the extension of the conventional $MU$ curve to reveal hormetic patterns. The solid lines depict the standard relationship between $TU$ and $MU$, whereas the dashed lines extrapolate this relationship to showcase hormetic effects at higher product volumes. As $TU$ becomes increasingly negative, repercussions like environmental degradation and human subjugation eventually emerge. $RMU$ represents the relative change in $MU$ compared to the initial $MU$ at $n = 0$.*

However, the conventional model of decreasing $MU$ relies on the assumption that all paperclips are both produced and delivered at time $t = 0$. But what about scenarios where this assumption is false? For example, Hartmann (2006) analyzed the intertemporal effects of consumption on golf demand, showing that the $MU$ of playing golf decreases if the consumer has played golf recently, but recovers after a certain period. In our case, paperclips may be produced in batches at different times, in response to varying demand. As demand increases over time, so does $MU$, which increases the gradient of the $MU$ curve. This raises the $TU$ curve and subsequently increases the hormetic limit, as demonstrated in Figure 3.

To demonstrate this effect, consider the pizza slice example. When a person consumes all slices of a pizza immediately, $MU$ diminishes with each added slice. But if the person consumes one slice every two hours, the marginal utility curve changes; it initially falls post-consumption but subsequently rises as the person becomes hungry again. Hence, the introduction of time as a variable elevates the $MU$ curve, which has the effect of increasing the hormetic limit and the hormetic apex for the *TU* curve[2]. This increase is approximately proportional to the time between pizza slices consumed.

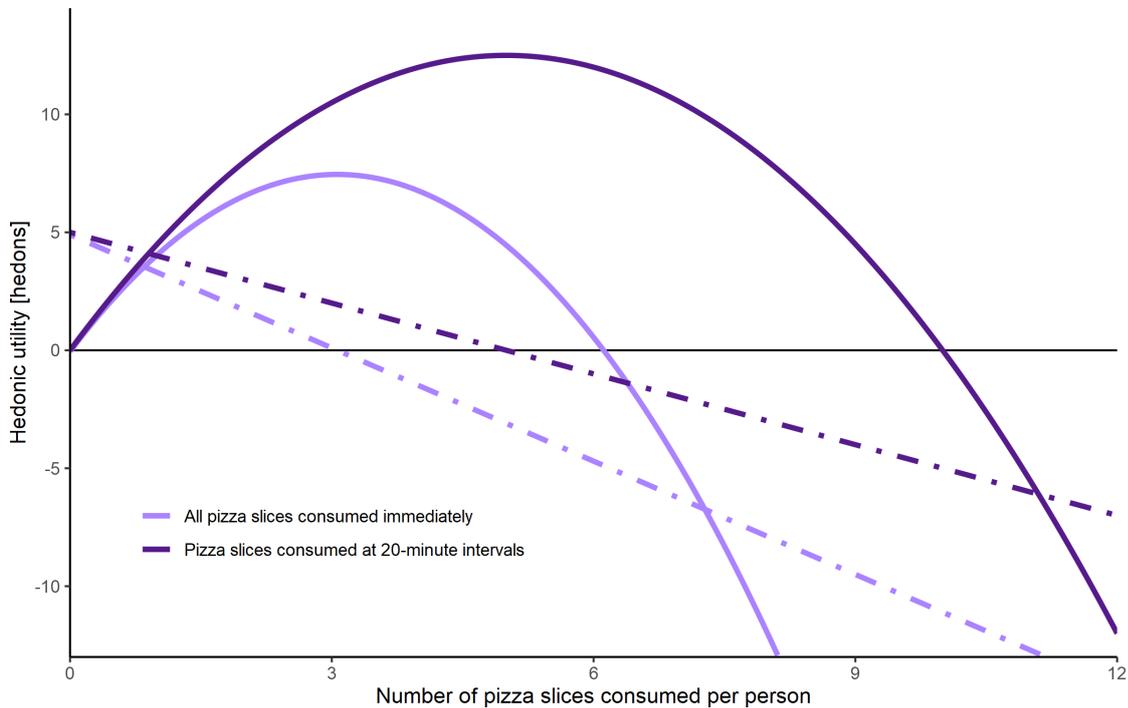

*Figure 3: Hypothetical comparison of **TU** (solid) and **MU** (dashed) curves for the scenario when all pizza slices are consumed immediately, versus when pizza slices are consumed at 20-minute intervals. The addition of the 20-minute interval raises the gradient of the **MU** curve, leading to an increase in both the hormetic apex and hormetic limit.*

Opponent process theory offers a compelling framework for explaining the temporal dynamics of hedonic utility in the context of repeatable behaviors. Historically, the hedonic and utilitarian aspects of a product were often viewed as distinct (Dhar & Wertenbroch, 2000; Voss et al., 2003), partly due to challenges in quantifying hedonic experiences (Kahneman et al., 1997). However, experiential utility encompasses various facets, including hedonic, emotional, and motivational elements (Kahneman et al., 1997; Solomon & Corbit, 1974). Indeed, Motoki et al. have shown that representations of hedonic and utilitarian value occupy similar neural pathways in the ventral striatum, indicating a correlation between these two states (Motoki et al., 2019).

It's possible that a person's hedonic response to a behavior could potentially serve as an indirect measure of the *MU* derived from that behavior. We can then model the opponent process dynamics within the brain generated by these behaviors, potentially leading to allostasis when executed frequently. This provides us with a mechanism to replicate the *TU* curve and set safe hormetic limits for behaviors such as 'paperclip creation'. We call our approach the HALO paradigm (Hormetic Alignment via Opponent processes). Below, we demonstrate the HALO method by performing a hormetic analysis of 'paperclip creation' to determine the safe limits of this simple behavior, then expanding this modeling process to other behaviors. In this way, we can program a value system for the AI agent – essentially an evolving database of values assigned to seed behaviors, from which the agent can learn to assign values to novel behaviors.

---

[2] It is also important to consider the scenario where the initial *MU* is either negative or zero, implying that the behavior is always undesirable. In these cases, the *TU* curve is monotonically negative.

# 4  Programming a value system with the HALO algorithm

We propose Algorithm 1 for using the HALO paradigm to program a value system that can regulate and optimize the behaviors performed by an AI agent. In this paradigm, a database of opponent process parameters for a range of seed behaviors is set up. The AI agent evaluates its environment, suggests a list of optimal actions to perform, and queries the database for similar behaviors. It then proposes opponent process parameters for the optimal actions based on their similarity to other behaviors, and by hormetic analysis. Finally, the agent selects and executes the best action, and repeats the process.

**Algorithm 1: HALO paradigm**
1. **initialize** environment $E$.
2. **initialize** database of opponent process parameters, $D_{op}$.
3. **while** agent is switched on:
    a. Evaluate $E$.
    b. Suggest a set of optimal actions $A$ based on $E$.
    c. Query $D_{op}$ for behaviors $b$ similar to $A$.
    d. **for** each $a$ in $A$:
        i. **if** prior similar behaviors $D_{op}(b)$ are available:
            1. Set opponent process parameters for $A(a)$ based on their proximity to $D(b)$.
        ii. **else**:
            1. Request human-suggested opponent process parameters for $A(a)$.
        iii. Conduct hormetic analysis to determine the hormetic apex and hormetic limit for $A(a)$ within a specified simulation time, $t_{sim}$.
        iv. Store opponent process parameters for $A(a)$ in $D_{op}$ (if not already stored).
    e. Select the action $a_{optimal}$ from $A$ that has the optimal combination of hormetic apex and hormetic limit.
    f. Execute $a_{optimal}$ for the duration of $t_{sim}$.
    g. Re-evaluate $E$, and repeat.
4. **end**

Paperclip creation is an ideal seed behavior for populating the database. It is a low-risk activity with quantifiable benefits, such as organizing papers, along with associated costs like production and storage expenses. Creating one paperclip produces a small but perceptible improvement to one's productivity, while turning the world into paperclips is both unproductive and, even worse, destructive. Using this information, we can propose parameters for a set of opponent processes that would accurately reflect the diminishing $MU$ of creating new paperclips, in terms of hedonic utility, which is measured in hedons – units of pleasure if positive, or pain if negative.

Here, we demonstrate two methods for hormetic analysis within the HALO paradigm. The first, Behavioral Frequency Response Analysis (BFRA), employs Bode plots to examine how a person's emotional states vary in response to the person performing a behavior at different frequencies (Henry, Pedersen, Williams, & Donkin, 2023; Schulthess et al., 2018). The second method, Behavioral Count Response Analysis (BCRA), parallels BFRA but uses the count of behavioral repetitions as the independent variable instead of behavioral frequency. To quantify opponent process parameters for the 'paperclip production' behavior, we adapted Henry et al.'s PK/PD model of allostatic opponent processes (Henry, Pedersen, Williams, & Donkin, 2023) using the *mrgsolve* package (v1.0.9) in R v4.1.2 (Baron & Gastonguay, 2015; Elmokadem et al., 2019; R Core Team, 2022). This involved coding a system of ordinary differential equations (ODEs) to represent the a- and b-processes in response to each successive behavioral dose. The simulation code, along with examples of modifying the a- and b-process parameters, is provided in the Supplementary Materials. We recommend consulting Henry et al. (2023) for a more detailed explanation of the behavioral posology model on which HALO is built, including demonstrations of the relationship between PK and PD in the context of this model.

## 4.1 PK/PD model of opponent processes leading to hormesis

Below, we present the mathematical framework for our model. We defined a behavior as a repeatable pattern of actions performed by an individual or agent over time. In the context of behavioral posology, we refer to individual actions that make up the behavior as 'behavioral doses'. We employed a modified equation for behavioral doses (Henry, Pedersen, Williams, & Donkin, 2023; Manojlovich & Sidani, 2008):

$$Dose_{action} = \int_0^{Duration_{action}} Potency \, dt$$

where $Potency$ is a scalar representing the hedonic utility of creating a paperclip compared to other actions (set to 1 for simplicity); $Amount$ is a constant signifying the time allocated to creating the paperclip; $Frequency$ denotes the production rate in $min^{-1}$; and $\overline{Dose_{individual\ action}}$ represents the mean dose per action over the $Duration$ in which $Dose_{cumulative\ behavior}$ is assessed in $min$. In this case, since $Potency$ and $Duration_{action}$ are constants, $Dose_{action}$ is also a constant. This leaves two options for performing hormetic analysis: the BFRA, performed in the frequency domain when the number of behavioral repetitions, $n$, is kept constant, and the BCRA, performed in the temporal domain when $Frequency$ is kept constant.

Readers unfamiliar with PK/PD modeling are directed to Mould & Upton's introductory papers (Mould & Upton, 2012, 2013; Upton & Mould, 2014). Our PK/PD model is a mass transport model that loosely mimics dopamine's pharmacokinetic dynamics in the brain (Chou & D'Orsogna, 2022), and incorporates nonlinear pharmacodynamic elements to simulate neurohormonal dynamics in regions such as the hypothalamic-pituitary-adrenal (HPA) axis (Karin et al., 2020, 2021). The model's state-space representation is provided below, with detailed descriptions of all variables and parameters available in Table 1.

1) $\dfrac{dDose}{dt} = -k_{Dose} Dose$

2) $\dfrac{da_{pk}}{dt} = k_{Dose} Dose - k_{a,pk} a_{pk}$

3) $\dfrac{db_{pk}}{dt} = k_{a,pk} a_{pk} - k_{b,pk} b_{pk}$

4) $\dfrac{da_{pd}}{dt} = E_{0_a} + \dfrac{E_{max_a} \cdot a_{pk}{}^{\gamma_a}}{EC_{50_a}{}^{\gamma_a} + a_{pk}{}^{\gamma_a}} - k_{a,pd} a_{pd}$

5) $\dfrac{db_{pd}}{dt} = E_{0_b} + \dfrac{E_{max_b} \cdot b_{pk}{}^{\gamma_b}}{EC_{50_b}{}^{\gamma_b} + b_{pk}{}^{\gamma_b}} - k_{b,pd} b_{pd}$

6) $\dfrac{dH_{a,b}}{dt} = k_{a,pd} a_{pd} - k_{b,pd} b_{pd} - k_H H_{a,b}$

For all simulations performed, the default parameters to produce a short, high-potency a-process followed by a longer, low-potency b-process were as follows: $k_{Dose} = 1, k_{a,pk} = 0.02, k_{b,pk} = 0.004, k_{a,pd} = 1, k_{b,pd} = 1, k_H = 1, E_{0_a} = 0, E_{max_a} = 1, EC_{50_a} = 1, \gamma_a = 2, E_{0_b} = 0, E_{max_b} = 3, EC_{50_b} = 9, \gamma_b = 2$. These parameters were used for all simulations in this article unless stated otherwise. At time $t = 0$, the initial values of the compartments were: $Dose(0) = 1, a_{pk}(0) = 0, b_{pk}(0) = 0, a_{pd}(0) = 0, b_{pd}(0) = 0,$ and $H_{a,b}(0) = 0$. Infusion time was set to one minute, effectively instantaneous on the timescale used.

Table 1: Meaning of variables and parameters in PK/PD model.

| PARAMETERS | DESCRIPTIONS | DEFAULT VALUE |
|---|---|---|
| $t$ | Time elapsed, in minutes | - |
| $Dose$ | Behavioral dose compartment for hormonal and neurochemical concentrations following an action | 1 |
| $a_{pk}$ | Pharmacokinetic compartment for a-process | - |
| $a_{pd}$ | Pharmacodynamic compartment for a-process | - |
| $b_{pk}$ | Pharmacokinetic compartment for b-process | - |
| $b_{pd}$ | Pharmacodynamic compartment for b-process | - |
| $k_{Dose}$ | Pharmacokinetic clearance rate for $Dose$ compartment | 1 |
| $k_{a,pk}$ | Clearance rate for pharmacokinetic a-process compartment | 0.02 |
| $k_{a,pd}$ | Clearance rate for pharmacodynamic a-process compartment | 1 |
| $k_{b,pk}$ | Clearance rate for pharmacokinetic b-process compartment | 0.004 |
| $k_{b,pd}$ | Clearance rate for pharmacodynamic b-process compartment | 1 |
| $E_{0_a}$ | Baseline effect coefficient for a-process Hill equation | 0 |
| $E_{max_a}$ | Maximum possible effect coefficient for a-process Hill equation | 1 |
| $EC_{50_a}$ | Half-maximal effect coefficient for a-process Hill equation | 1 |
| $\gamma_a$ | Sigmoidicity coefficient for a-process Hill equation | 2 |
| $E_{0_b}$ | Baseline effect coefficient for b-process Hill equation | 0 |
| $E_{max_b}$ | Maximum possible effect coefficient for b-process Hill equation | 3 |
| $EC_{50_b}$ | Half-maximal effect coefficient for b-process Hill equation | 9 |
| $\gamma_b$ | Sigmoidicity coefficient for b-process Hill equation | 2 |
| $H_{a,b}$ | Pharmacodynamic compartment for Total Utility | - |
| $k_H$ | Clearance rate for pharmacodynamic $U_{a,b}$ compartment | 1 |

Equations 4) and 5) are implementations of the Hill equation, which governs the biophase curve – the relationship between pharmacokinetic concentration and pharmacodynamic effect. Although the pharmacodynamic compartments introduce complexity to the model, they provide an independent system outside of the pharmacokinetic mass transport system that is essential for generating hormetic effects. These effects arise from the non-linear interaction between the pharmacodynamic effects produced by the a- and b-processes[3].

For a single behavioral dose initiated at time $t = 0$, the integral of the utility compartment over time, $H_{a,b}(t)_{single}$, quantifies the hedonic utility produced by the opponent processes triggered by that behavioral dose over the simulation time $t_{sim}$. This value is equal to the initial marginal utility, $MU_{initial}$:

$$7)\ MU_{initial} = \int_0^{t_{sim}} H_{a,b}(t)_{single}\ dt$$
$$= \int_0^{t_{sim}} \left( \frac{k_{a,pd} a_{pd}(t) - k_{b,pd} b_{pd}(t) - \frac{dH_{a,b}(t)}{dt}}{k_H} \right) dt$$

This represents the summed hedonic utility for a single instance of the behavior. To find the total utility $TU$, the effect of multiple behavioral doses delivered sequentially can be summed to find the integral for $H_{a,b}(t)_{total}$, representing the total hedonic utility from all doses combined:

$$8)\ TU = \int_0^{t_{sim}} H_{a,b}(t)_{multiple}\ dt$$

---

[3] Specifically, the a-process pharmacodynamic effect surpasses the b-process effect at low pharmacokinetic levels, whereas the opposite holds true at elevated pharmacokinetic levels, leading to a biphasic dose-response curve. This would not be possible with only pharmacokinetic compartments since the system would only scale linearly due to the laws of mass conservation.

$$= \sum_{i=0}^{n} \int_{i/f}^{t_{sim}} \left( \frac{k_{a,pd} a_{pd,i}(t) - k_{b,pd} b_{pd,i}(t) - \frac{dH_{a,b,i}(t)}{dt}}{k_H} \right) dt$$

where $n$ is the count of behavioral doses delivered at a frequency $f$ over $t_{sim}$. Note that if $t_{sim} < \infty$, the value of $TU$ will increase for all values of $f$ and $n$, since the finite simulation will predominantly feature positive a-processes, given their shorter decay duration compared to b-processes.

This also provides us with an indication of whether the behavior is hormetic. If we have a behavior with $MU_{initial} > 0$ and a b-process integral sufficient to produce significant allostasis, we can generally predict that low frequencies of that behavior will produce a positive $TU$, while higher behavioral frequencies will lead to allostasis that produces a negative $TU$. (This is demonstrated in Figure 5 and Figure 6.)

In standard economic models, the $TU$ curve is calculated as the integral of the $MU$ curve. However, the temporal dynamics of opponent processes complicate the relationship between $TU$ and $MU$, meaning that simulation is required to quantify the rate of hedonic allostasis. Figure 4 demonstrates what happens if we separate the a- and b-processes in Figure 1. It turns out that the b-process curve is proportional to the relative $MU$, or $RMU$, of the behavior. To illustrate this, let us consider a person consuming a bag of sweets throughout the day. Each time a person consumes a sweet, they experience a rush of dopamine, endorphins, and energy from the sugar in the sweet, all of which contribute to a hedonic a-process. This is followed by an opposing b-process, during which the person experiences a small depletion of dopamine and endorphins, along with decreased craving for another sweet. This corresponds to a decline in the $MU$ of consuming an additional sweet, aligning with the law of diminishing $MU$. However, as time elapses, this decline in $MU$ decays exponentially as the person's craving for another sweet gradually increases. If the person maintains a consistent frequency of sweet consumption, a hedonic equilibrium is eventually achieved. This equilibrium represents a balance between the decreasing $MU$ that follows sweet consumption, and the gradual increase in craving for another sweet as time passes. Consequently, the $RMU$ equates to the allostatic load, which is proportional to the b-process curve.

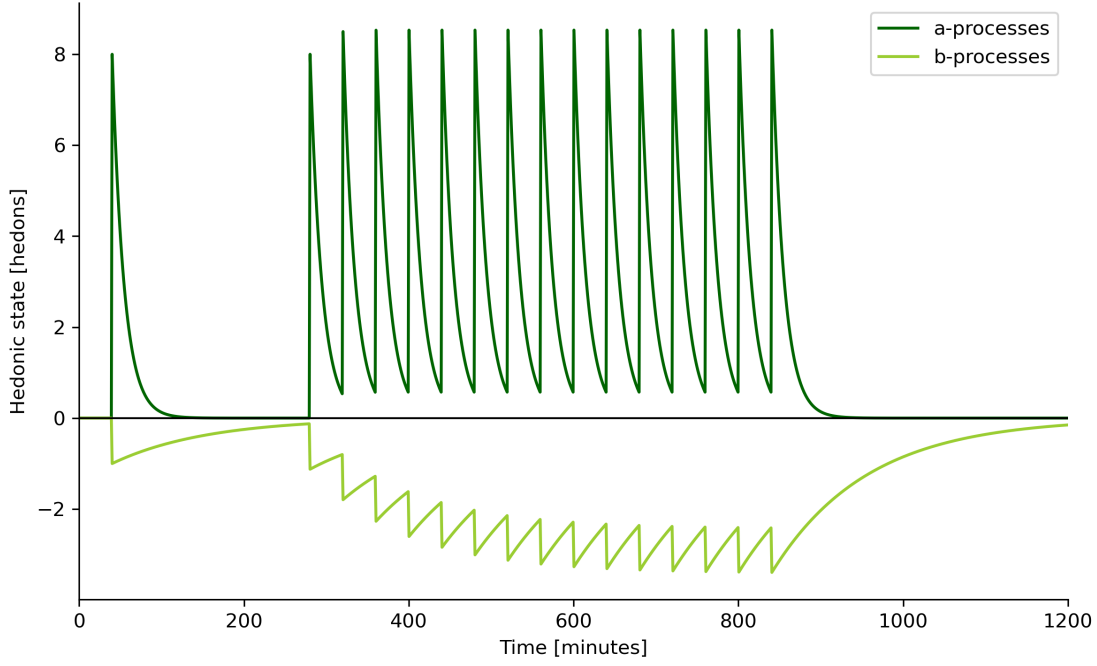

*Figure 4: Illustration of the same PK/PD model simulation in Figure 1, but plotting a- and b-process compartments as separate compartments. Allostasis is more pronounced for the b-processes due to their longer decay period, which explains why opponent process allostasis is negative overall in Figure 1, when the a- and b-processes are combined. In this model, the cumulative b-process curve is proportional to the RMU of the behavior. When the behavior is performed at a constant frequency, allostasis initially occurs, but a steady state is quickly reached.*

The optimal behavioral frequency or count is found by quantifying the hormetic apex. To do this, one must create a Bode magnitude plot to show either the frequency-response or count-response curve, assuming constant potency and duration for each behavioral dose. For the BFRA, this can be performed analytically. Assuming the behavior persists at a constant frequency indefinitely, a quasi-steady state solution can be computed once all compartments stabilize. At this equilibrium, the average inflow matches the average outflow for each compartment. The full derivation for the steady-state solution can be found in Appendix 1. This is achieved by setting all derivatives equal to zero in equations 1) to 6), then deriving the steady-state solution for the final $H_{a,b}$ compartment:

$$9)\ H_{a,b_{steady\,state}} = \frac{E_{0_a} + \dfrac{E_{max_a} \cdot \dfrac{D_0 f}{k_{a,pk}}^{\gamma_a}}{EC_{50_a}^{\gamma_a} + \dfrac{D_0 f}{k_{a,pk}}^{\gamma_a}} - E_{0_b} - \dfrac{E_{max_b} \cdot \dfrac{D_0 f}{k_{b,pk}}^{\gamma_b}}{EC_{50_b}^{\gamma_b} + \dfrac{D_0 f}{k_{b,pk}}^{\gamma_b}}}{k_H}$$

Thus, the Bode plot for a BFRA can be quantified analytically by calculating $H_{a,b_{steady\,state}}$ as a function of the behavioral frequency, $f$. On the other hand, a BCRA does not produce a steady-state solution since it uses finite behavioral counts. Hence, the Bode plot for a BCRA must be computed numerically using equation 8).

# 5 Hormetic analysis of paperclip-producing agent

To demonstrate the HALO paradigm, we must first show how a human can manually program a value for a seed behavior that the AI can learn from. We imagined a situation where an AI was tasked with producing the optimal number of paperclips for a small office of ten employees handling moderate paperwork. To achieve this, we performed hormetic analysis in two hypothetical scenarios. In the first scenario, it was assumed that the human workers consistently required a steady stream of paperclips at a rate of 0.015 min$^{-1}$ – roughly one per hour. This required a BFRA to optimize the *rate* of paperclip production. In the second scenario, the workload occasionally surged, meaning the workers required batches of five paperclips at certain times. Here, BCRA was used to optimize the *count* of paperclips produced. In both cases, we needed to propose opponent process parameters that would achieve three things:

1) Provide plausible $MU$ values that would match the utility of a paperclip in real life.
2) Produce a hormetic curve with an apex that matched the target frequency or count of paperclips required.
3) Produce a sensible hormetic limit that would prevent excessive production of paperclips.

To simplify the parameter selection process, we chose to only vary the $EC_{50_b}$ parameter. Increasing $EC_{50_b}$ reduces the b-process magnitude, reducing the rate of b-process allostasis and thus increasing both the hormetic apex and hormetic limit.

## 5.1 Behavioral Frequency Response Analysis

The first scenario allowed us to set long-term production caps on the AI agent, by regulating the frequency of paperclip production via BFRA. To examine the frequency-response of the model with a BFRA, we fixed $n$ and *Potency* and evaluated total utility $TU$ as a function of behavioral frequency, $f$, using equation 8). At a constant $f$, $H_{a,b}(t)_{multiple}$ converges to a steady-state value, $H_{a,b_{steady\,state}}$, which is proportional to $TU$. This framework allowed analytical calculation of $TU_{apex}$ and $TU_{NOAEL}$ (the hormetic apex and hormetic limit), and their respective frequencies $f_{apex}$ and $f_{NOAEL}$. The challenge lay in determining $f_{NOAEL}$ – the safe upper limit of paperclip production frequency – and, ideally, $f_{apex}$ to optimize its production rate in terms of hedonic utility, as experienced by humans.

Figure 5 shows some of the simulated results from the BFRA performed to find suitable opponent process parameters to produce an $f_{apex}$ of 0.015 min$^{-1}$. $EC_{50_b}$ was set to 9.2, keeping all other parameters in Table 1 constant. Figure 5a shows the *mrgsolve* simulation of the $H_{a,b}$ compartment over time at $f_{apex}$, demonstrating the optimal frequency at which the integral of the $H_{a,b}$ compartment is highest, thus maximizing $TU$. Figure 5b shows the simulation at $f_{NOAEL}$, which in this case is approximately 0.025 min$^{-1}$. At $f_{NOAEL}$, the steady-state value of the simulation is zero, meaning that the $MU$ of new paperclips being created is zero. At higher frequencies, the steady-state value becomes increasingly negative, which leads to the decreasing portion of the hormetic curve in Figure 5c.

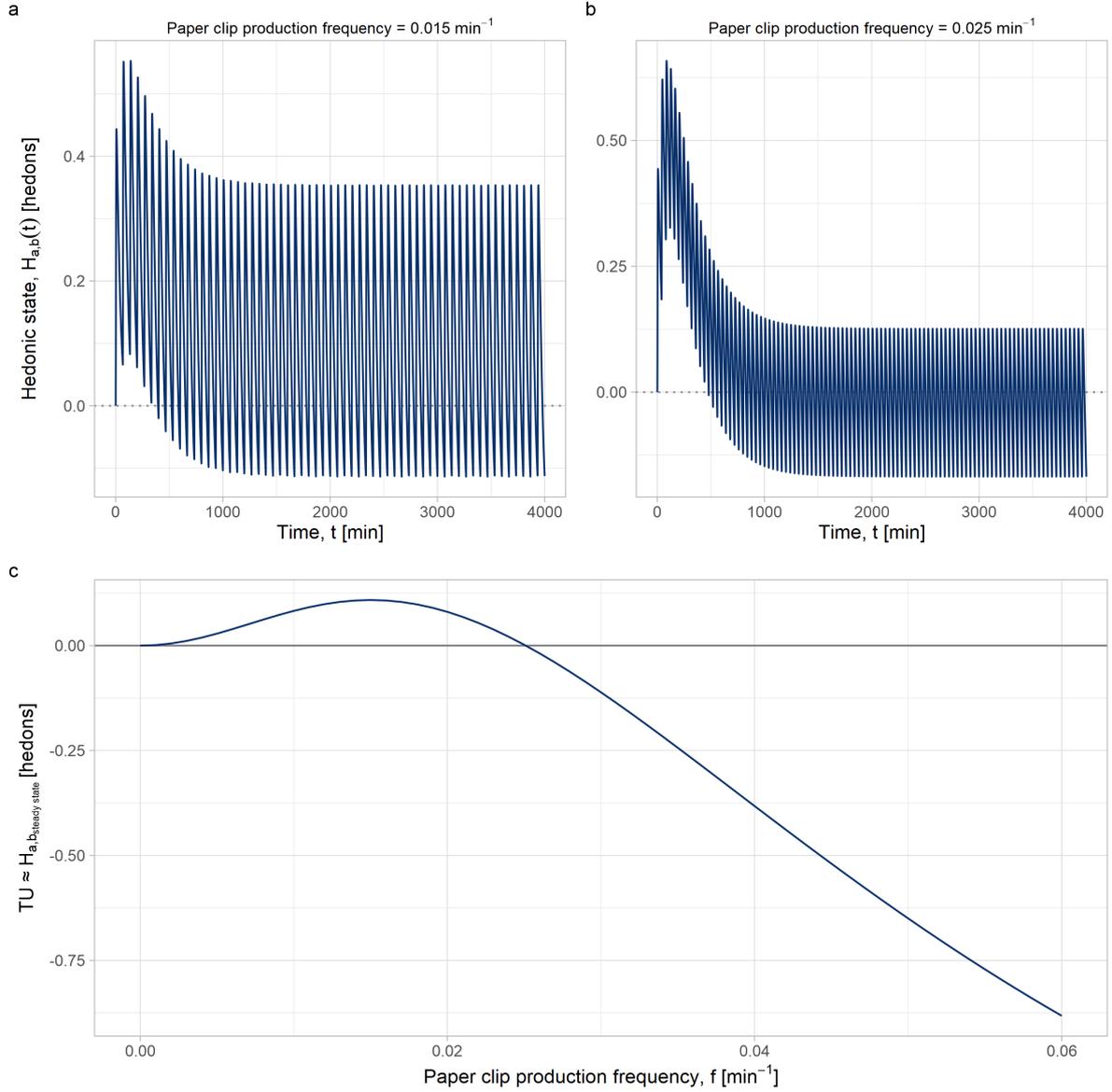

*Figure 5: BFRA performed to determine optimal opponent process parameters for an AI agent aiming to produce 0.015 paperclips per minute. $EC_{50_b}$ was set to 9.2, keeping all other parameters in Table 1 constant. a,b) $H_{a,b}(t)_{multiple}$ scores generated by* mrgsolve *simulations at $f_{apex}$ (left) and $f_{NOAEL}$ (right). c) Bode magnitude plot of Total Utility as a function of behavioral frequency. $f_{apex} \approx 0.015$ min$^{-1}$, while $f_{NOAEL} \approx 0.025$ min$^{-1}$.*

We have included further BFRA examples in Appendix 2, demonstrating how parameter modifications from Table 1 influence $TU$ outcomes and change the shape of the hormetic curve. For example, increasing $EC_{50_b}$ shifts the biophase curve for the b-process, reducing the pharmacodynamic effects produced by equivalent pharmacokinetic concentrations. This reduces the b-process magnitude, lowering the rate of negative allostasis and increasing the steady-state value of the $H_{a,b}$ compartment, which increases $f_{NOAEL}$, the hormetic limit. In essence, higher ratios of a- to b-process magnitudes increase the behavioral frequency required to maintain an allostatic rate that produces a negative steady-state.

## 5.2 Behavioral Count Response Analysis

In the second scenario, the AI needed to adjust production levels to account for fluctuating demand. Once the *MU* of creating a new paperclip became negative, the AI was required to halt production until the system recovered to homeostasis. This scenario required an examination of behavioral bursts – short, high-frequency bursts of paperclip production – using the BCRA approach to examine the count-response of the model.

For simplicity, our analysis focused solely on the first behavioral burst, ignoring subsequent bursts. To perform a BCRA, we fixed *Potency* and $f$, and measured the numerical integral of *TU* as a function of the dose count, $n$. This method does not allow steady state to be reached, since the behavior does not repeat to infinity. This necessitates time-domain simulation for optimal $n$ determination. Future research should explore whether an algorithmic approach can identify the optimal value of $n$ for each set of opponent process parameters.

Figure 6 shows some of the simulated results from the BFRA performed to find suitable opponent process parameters to produce a hormetic apex of $n_{apex} = 5$ paperclips. $EC_{50_b}$ was set to 12.4, keeping all other parameters in Table 1 constant. The axes in the figure align with those in Figure 5, except for the bottom plot, which shows the integral of *TU* over $t_{sim}$ plotted against $n$. This differs from the BFRA, where the steady-state value of $H_{a,b}$ was plotted against $f$. At $n_{NOAEL}$ (12 paperclips produced), the *MU* of new paperclips is already negative, indicating that 12 paperclips are too many for the task at hand.

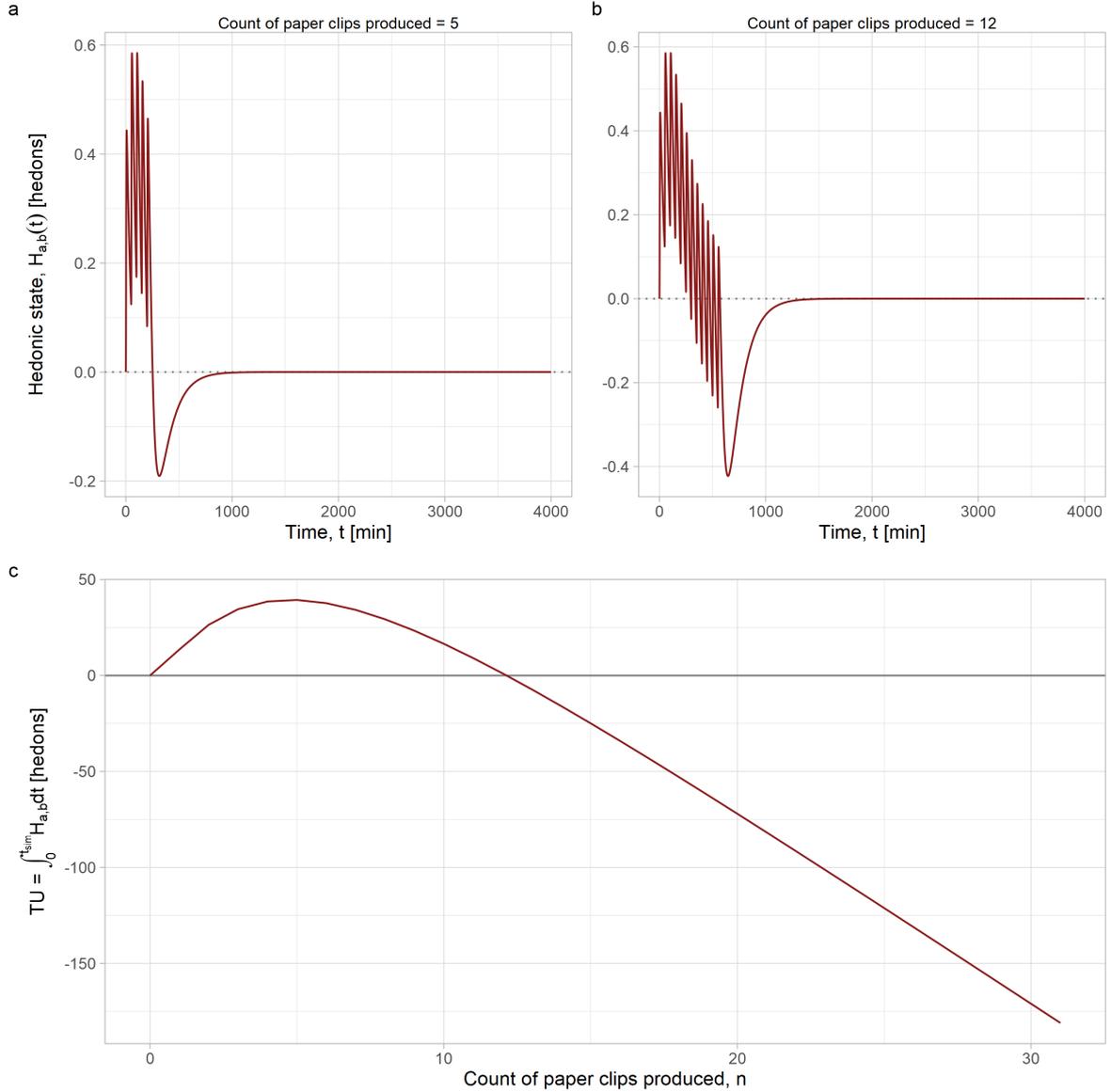

*Figure 6: BCRA performed to determine optimal opponent process parameters for an AI agent aiming to produce a single batch of 5 paperclips. $EC_{50_b}$ was set to 12.4, keeping all other parameters in Table 1 constant. a,b) $(t)_{multiple}$ scores generated by* mrgsolve *simulations at $f_{apex}$ (left) and $f_{NOAEL}$ (right). c) Bode magnitude plot of Total Utility as a function of behavioral count. $n_{apex} \approx 5$, while $n_{NOAEL} \approx 12$.*

We have included further BCRA examples in Appendix 2. Generally, both BCRA- and BFRA-generated $TU$ curves exhibit similar sensitivities to parameter changes.

## 6   Using HALO to classify new behaviors

So far, we have demonstrated a method to quantify the hedonic value of paperclip creation in various contexts, considering the number of paperclips recently created and, in the extreme case, the ethical implications of mass extinction due to overproduction. This allows humans to impute values for seed behaviors. Then, by repeating the HALO algorithm iteratively for novel behaviors, an AI can build a 'behavioral value space' consisting of opponent process parameters for different behaviors, each with their own hormetic apexes and limits. This represents a potential solution to the value-loading problem, as it presents a way to both optimize and regulate AI behaviors based on human emotional processing.

Figure 7 shows a subset of the behavioral value space for $TU_{apex}$ values that can be created by combining different combinations of variables, while keeping all other variables constant (refer to Table 1 for their defaults). Certain variable combinations produce complex interaction effects, while others produce more predictable effects. This could be used to restrict the value space to predictable outcomes. For example, adjusting the $k_H$ parameter (the decay constant of the final $H_{a,b}$ compartment) notably impacts the curve's sharpness, while maintaining the same value of $f_{NOAEL}$. Hence, the $k_H$ parameter could be used to distinguish behaviors that have identical hormetic limits but greater magnitudes of risk and reward. In contrast, parameters like $EC_{50_b}$ or $\gamma_b$ exhibit nonlinear effects on the hormetic curve and alter the hormetic limit. These parameters may be better suited to distinguish between behaviors with different hormetic limits. Hence, by restricting the value space to combinations of $k_H$ and $EC_{50_b}$, for example, one can produce a feasible set of hormetic outcomes that could be used to represent a wide range of behaviors that are safe to perform. Examples of these effects are provided in Appendix 2.

However, not all $TU$ curves exhibit true hormesis, instead staying positive over the entire range of frequencies or counts. The paperclip maximiser scenario is a case of an AI agent that has not been bounded by a hormetic limit. Thus, caution is required during value space classification, and boundaries will need to be placed on the value space parameters to avoid all non-hormetic outcomes, including both non-negative and monotonically negative solutions.

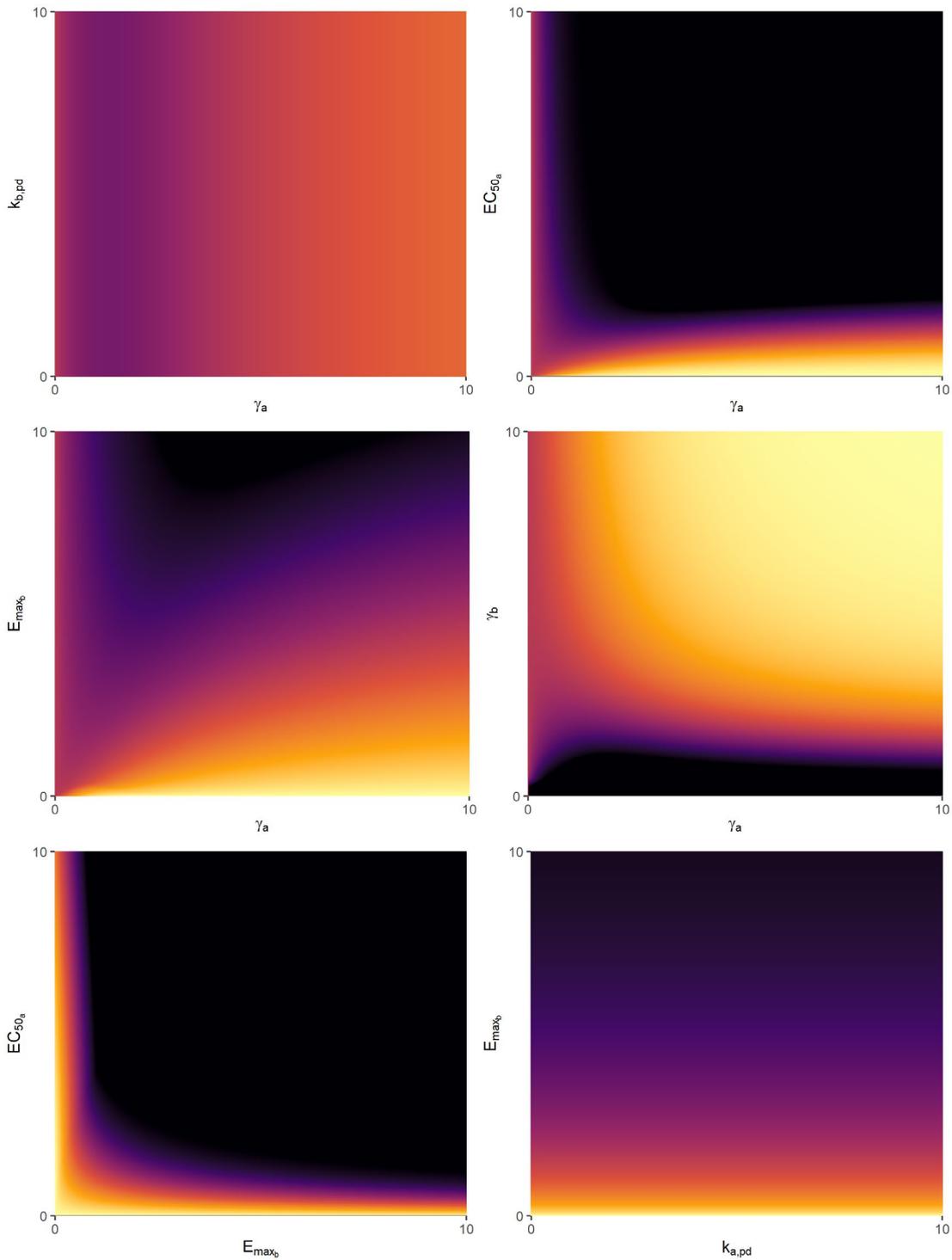

*Figure 7: Demonstration of behavioral value space, resulting from combinations of parameters selected from all pairwise variable combinations. Individual behaviors can be placed within these graphs at locations that suit their utility and risk profiles. Colours correspond to the value of $TU_{apex}$, ranging between 0 (black) and 1 (light). Hence lighter colors represent a higher apex of the BFRA curve, indicating greater $TU$ (and most likely, a greater hormetic limit), while behaviors within the black regions of value space have a very low value of $MU_{initial}$, and shouldn't be performed at all. Note that not all positive solutions will be hormetic as some are non-negative solutions, meaning they don't have a hormetic limit.*

This method of value-loading may work within the weak-to-strong generalization paradigm of AI alignment. Once the weaker model has categorized the value of a diverse set of behaviors (with human help), these behaviors can form a behavioral value space: a database of a- and b-process parameters, analogous to $D_{op}$ in Algorithm 1. This is similar to a 'synthetic data' training approach, in that we augment the training dataset with algorithmically generated examples, using only a few human-generated seed entries to start with (Gulcehre et al., 2023; Honovich et al., 2022). The stronger model can then generalize from this value space to classify novel behaviors that are beyond the weaker model's capacity to solve.

Decision tree methods such as XGBoost (Chen & Guestrin, 2016) or centroid-based methods such as CentNN (Ngoc & Park, 2018) could be used to estimate the location of novel behaviors in value space, based on their proximity to other behaviors. However, this may not work well for behaviors that are significantly different from those already defined in the value space – in other words, out-of-distribution (OOD) behaviors (Geirhos et al., 2020). Furthermore, instances of near-identical behaviors, possibly differentiated solely by context, may lead to poor discrimination (Eysenbach et al., 2018). In clear OOD cases, an error can be raised, prompting human intervention. This could also be combined with techniques such as linear probing, where the final output layer of a neural network is modified while keeping all other feature layers of the model frozen; this technique has been shown to outperform fine-tuning on OOD data in terms of model accuracy (Kirichenko et al., 2023; Kumar et al., 2022). A logistic regression model could be trained to set a threshold for detecting OOD behaviors. However, defining the threshold for human intervention is a complex challenge. Hence, using the hormetic limit as an uncertainty metric remains a prudent practice.

## 7 Discussion

In this article, we have demonstrated how the HALO paradigm can be used to help create a value system for AI. Our model offers a starting point for quantifying the hedonic utility of repeatable behaviors. While the value of a behavior is not merely composed of hedonic utility alone, HALO may form the basis of a more advanced system of allostatic behavioral regulation that includes hedonic, social, economic, legal, and ethical considerations. Thus, HALO may provide a method of regulating advanced AI algorithms with repeatable behaviors.

HALO can be used to set hormetic limits in terms of both behavioral count and frequency – features not found in current methods like Reinforcement Learning with Human Feedback (RLHF) that assess singular actions in binary terms (Bennett et al., 2023). Such features are crucial for real-world AI interactions, where behaviors need clear frequency and count constraints. The hormetic limit serves as a safety buffer, allowing the AI to aim for the hormetic apex with the assurance that it is behaving within a margin of error. If the hormetic limit is zero, HALO will prevent the behavior from being performed at all. To further enhance AI safety, an uncertainty factor could be added to reduce the hormetic limit initially, then gradually increase it as trust grows in the reward model.

HALO provides several benefits for AI regulation. The temporal analysis of opponent processes allows the AI model to assess both the immediate and future impacts to humans while sidestepping pitfalls like temporal discounting. It also provides nuanced categorization of behaviors, allowing shades of grey and fuzzy reasoning in uncertain environments, rather than binary polarization. Such metric-driven ethics could be crucial for guiding the moral compass of intelligent robots, which operate in high uncertainty environments (Narayanan, 2023). Finally, the HALO model aligns more closely with human emotional responses, being grounded in biological principles such as allostasis, opponent process theory, and PK/PD modeling. In turn, it's possible that the development of a HALO-based value space for AI behaviors could offer insights into human behavioral psychology, and in particular, the healthy limits of repeatable human behaviors.

## 7.1 Empirical data for modeling hormesis

The PK/PD model offers several degrees of freedom for modulating the hormetic curve, providing broad flexibility for categorization of behaviors. However, this also complicates the creation of a comprehensive behavioral value space. The current approach requires extensive pre-calculation across various parameters, which is computationally demanding. It also confines users to specific parameter ranges and is fragile to environmental changes that could affect the reward model. Additionally, the complexity of solving stochastic differential equations means that evaluating and categorizing behaviors is time intensive. Future research could focus on selecting a subset of opponent process parameters to optimize that still provides sufficient flexibility in modulating the $TU$ curve.

While human intelligence and AI have significant differences, human psychology provides insights that may guide the development of aligned AI models (Goertzel, 2014). The challenge lies in discerning the precise temporal dynamics of human psychological responses to behaviors. While it has been proposed that emotional responses decay exponentially (Picard, 2000), little research has been done to quantify these decays. Real-time emotional responses can be potentially monitored using fMRI data (Heller et al., 2015; Horikawa et al., 2020), but sustaining such monitoring over extended periods for diverse populations also requires longitudinal research. Ecological Momentary Assessment (EMA) studies may allow us to compile a comprehensive dataset of parameters related to a- and b-processes, which can be used to quantify the neurodynamics of affect in conjunction with fMRI data (Heller et al., 2015). EMA data allows us to capture individualized responses to diverse behaviors, which may facilitate the construction of eigenmoods derived from combined emotional states (Cambria et al., 2011, 2015; Cowen & Keltner, 2021), allowing us to derive more accurate opponent process models. This may also allow us to incorporate more dimensions (including social, economic, legal, and ethical considerations) into the decision-making process.

## 7.2 Preventing reward hacking

Any alignment method may be susceptible to design specification problems where the agent's incentives differ from the creator's true intentions (Leike et al., 2018). One hypothetical example is wireheading, where the agent, tasked with maximizing hedonic pleasure in humans, achieves this goal by directly stimulating the reward centres of the brain with electrodes (Yampolskiy, 2014). A more practical example was posed by Urbina et al. (2022), in which a drug discovery reward model was inverted to create lethal toxins. Such an inversion, if applied to the hormesis model, could have catastrophic effects, emphasizing the importance of securing the reward model.

Reward tampering often takes place along causal pathways that are poorly understood by humans. The intricacy of these pathways amplifies the risk of unforeseen AI exploitation (Everitt et al., 2021). To counteract this, a deeper exploration of these causal routes is essential to prevent AI from leveraging them for self-benefit. Specifically, understanding the pathways influencing human emotional responses is pivotal. Such insights empower AI to better discern how behaviors causally impact human emotions.

The hormetic framework offers insights into addiction within AI systems. If an AI exceeds the hormetic threshold in its behaviors, it can be analogously viewed as being 'addicted' to that behavior, persistently engaging in it despite detrimental outcomes for humanity. Analyzing count- and frequency-based hormesis confers a significant advantage: it prompts the AI to prioritize long-term outcomes, mitigating the risk of addictive cycles. This may help to solve the incommensurability problem in hedonic (also known as felicific) calculus – the idea that the value of all behaviors cannot be compared on a common scale (Klocksiem, 2011). The addition of allostasis allows us to compare the hedonic utility of different behaviors over both short- and long-term timespans, providing us with a more accurate metric for comparing behaviors. If some emotions, like anxiety and satisfaction, cannot be compared directly, it's also possible to assign opposing process parameters for various dimensions simultaneously. While this method makes the model more complex, it enhances safety by constraining the AI's behaviors within the smallest hormetic limit among multiple emotional dimensions. But once the AI discerns that its behaviors are bounded by allostasis, it might recalibrate its behavior to prioritize short-term outcomes. To deter addictive tendencies, such as excessive paper-clip production, AI designers should embed the prioritization of long-term welfare over immediate gains within the algorithm.

Experimentation and collaborative learning will be necessary to solve these problems. In reinforcement learning, the ensemble approach, combining multiple algorithms into a single agent, has been shown to greatly enhance model training and accuracy (Faußer & Schwenker, 2015; Lindenberg et al., 2020; Singh, 1991; Sun & Peterson, 1999; Wiering & van Hasselt, 2008). Thus, multiple AI agents could combine their learnings to form a shared value system – essentially a crowd-sourced database of optimal behaviors. Further, experiments in controlled sandbox environments such as Smallville (Park et al., 2023) would facilitate the natural selection of superior agents. For example, Voyager – an LLM-powered learning agent operating in the Minecraft environment (Wang et al., 2023) – could be an ideal agent for testing the HALO algorithm. Voyager works by creating an ever-expanding 'skill library' of executable code to perform various actions within Minecraft, using a novelty search approach to discover new behaviors (Wang et al., 2023). Since most of these actions are simple and repeatable, HALO could be used to assign hormetic limits and hormetic apexes for each behavior in the skill library, which could then be scaled up to increasingly complex behaviors. Different value systems could lead to varied AI personalities, which could then collaborate and compete with one another in an Axelrod tournament-like scenario to determine an optimal value system (Axelrod, 1980a, 1980b).

## 7.3 Limitations and future research

Our approach has some limitations that require further research to overcome. The first is a reliance on a simplified hormetic model that does not capture all variance in the human experience. BFRA assumes behaviors occur at an unchanging frequency, which does not capture real-world variability in behavioral timing. Furthermore, in everyday life, humans must ascertain whether their behaviors are within safe limits by judging the behavior's causal effects on their short- and long-term wellbeing, along with the wellbeing of those around them. This poses significant cognitive demands, especially when behaviors are coupled, potentially explaining the evolution of societal structures and religious systems—platforms adept at sharing knowledge on the hormetic limits of behaviors. In essence, humans are performing multivariate hormetic analysis (similar to Multicriteria Decision Analysis, or MCDA (Lahdelma et al., 2000; Steele et al., 2009)) in an attempt to balance the trade-offs for multiple behaviors, each with their hormetic curves.

A similar approach of multivariate hormetic analysis could be performed by AI agents, but this requires a more accurate understanding of both individual and group psychology. Allostatic regulation makes assumptions about individual human emotions that haven't yet been fully validated. To further complicate the issue, allostatic load may also build up within social groups as well, due to emotional and physiological linkage between individuals that result in correlated states of arousal (Saxbe et al., 2020). Thus, small changes in behavior or conditions may result in significant variations in emotional experiences for different individuals. This complexity, reminiscent of catastrophe theory, makes it challenging to accurately model allostatic rates (Zeeman, 1976). However, iterative refinement and crowdsourced value databases could help us understand which environments are more likely to lead to chaotic, unpredictable outcomes.

Similarly, the model assumes the b-process solely originates from the a-process. While the basic model demonstrates a plausible link between allostatic opponent processes and hormesis, it cannot replicate human emotional outcomes with absolute fidelity. A more comprehensive model might include additional compartments and extra clearance channels to account for more biological pathways. However, these simplifications were necessary to practically demonstrate HALO.

In summary, the development of HALO faces some challenges, such as the scalability of the database of opponent process parameters, the robustness of hormetic analysis against noise and uncertainty, and the ethical implications of using hedonic utility as a proxy for human values. However, we believe these issues can be solved by a multidisciplinary approach that requires a synthesis of knowledge in the fields of AI, psychology, neuroscience, economics, and philosophy.

# 8 Conclusion

HALO is a reward modeling approach that can be used to design a value system for alignment of AI agents, thus providing a potential solution to the value-loading problem. By treating behaviors as allostatic opponent processes, we can use either BFRA or BCRA to predict the hormetic apex and limit of behaviors and select optimal actions that maximize utility and minimize harm to humans. Our approach not only prevents extreme scenarios like the 'paperclip maximizer' but also paves the way for the development of a computational value system that enables an AI to learn from its decisions. We hope that our work will inspire further exploration of the potential of hormetic reward modeling for AI alignment, and we invite readers to improve upon this model by adapting the provided R code in the Supplementary Materials, where simulations for assessment of different behaviors can be performed with the 'bfra()' and 'bcra()' functions.

# 10 Appendix 1

## 10.1 Deriving a steady-state solution to the ODE system

This appendix contains a derivation of the steady-state solution for the $H_{a,b}$ compartment in the ODE model.

The equations for the ODE system used in our PK/PD model are:

1) $\dfrac{dDose}{dt} = -k_{Dose} Dose$

2) $\dfrac{da_{pk}}{dt} = k_{Dose} Dose - k_{a,pk} a_{pk}$

3) $\dfrac{db_{pk}}{dt} = k_{a,pk} a_{pk} - k_{b,pk} b_{pk}$

4) $\dfrac{da_{pd}}{dt} = E_{0_a} + \dfrac{E_{max_a} \cdot a_{pk}^{\gamma_a}}{EC_{50_a}^{\gamma_a} + a_{pk}^{\gamma_a}} - k_{a,pd} a_{pd}$

5) $\dfrac{db_{pd}}{dt} = E_{0_b} + \dfrac{E_{max_b} \cdot b_{pk}^{\gamma_b}}{EC_{50_b}^{\gamma_b} + b_{pk}^{\gamma_b}} - k_{b,pd} b_{pd}$

6) $\dfrac{dH_{a,b}}{dt} = k_{a,pd} a_{pd} - k_{b,pd} b_{pd} - k_H H_{a,b}$

To find the steady-state solution, all derivatives should be set to 0 to indicate that equilibrium has been reached in each compartment. ($X_{ss}$ refers to the steady-state concentration of compartment $X$.)

7) $0 = -k_{Dose} Dose_{ss}$

8) $0 = k_{Dose} Dose_{ss} - k_{a,pk} a_{pk_{ss}}$

9) $0 = k_{a,pk} a_{pk_{ss}} - k_{b,pk} b_{pk_{ss}}$

10) $0 = E_{0_a} + \dfrac{E_{max_a} \cdot a_{pk_{ss}}^{\gamma_a}}{EC_{50_a}^{\gamma_a} + a_{pk_{ss}}^{\gamma_a}} - k_{a,pd} a_{pd_{ss}}$

11) $0 = E_{0_b} + \dfrac{E_{max_b} \cdot b_{pk_{ss}}^{\gamma_b}}{EC_{50_b}^{\gamma_b} + b_{pk_{ss}}^{\gamma_b}} - k_{b,pd} b_{pd_{ss}}$

12) $0 = k_{a,pd} a_{pd_{ss}} - k_{b,pd} b_{pd_{ss}} - k_H H_{a,b_{ss}}$

From these equations, we can derive the following:

13) $k_{a,pk} a_{pk_{ss}} = k_{Dose} Dose_{ss}$
$\phantom{13)\ k_{a,pk} a_{pk_{ss}}} = k_{b,pk} b_{pk_{ss}}$

$$14)\ k_{a,pd}a_{pd_{ss}} = E_{0_a} + \frac{E_{max_a} \cdot a_{pk_{ss}}^{\gamma_a}}{EC_{50_a}^{\gamma_a} + a_{pk_{ss}}^{\gamma_a}}$$

$$15)\ k_{b,pd}b_{pd_{ss}} = E_{0_b} + \frac{E_{max_b} \cdot b_{pk_{ss}}^{\gamma_b}}{EC_{50_b}^{\gamma_b} + b_{pk_{ss}}^{\gamma_b}}$$

$$16)\ H_{a,b_{ss}} = \frac{E_{0_a} + \frac{E_{max_a} \cdot a_{pk_{ss}}^{\gamma_a}}{EC_{50_a}^{\gamma_a} + a_{pk_{ss}}^{\gamma_a}} - E_{0_b} - \frac{E_{max_b} \cdot b_{pk_{ss}}^{\gamma_b}}{EC_{50_b}^{\gamma_b} + b_{pk_{ss}}^{\gamma_b}}}{k_H}$$

Recall that to perform a BFRA, both *Potency* and $n$ are fixed, while $H_{a,b_{ss}}$ is measured as a function of $f$. Assuming the volume of each compartment is equal to 1, then at steady state, the rate of dose elimination $k_{Dose}Dose_{ss}$ is equal to the rate of dose administration $D_0 f$, where $D_0$ is the initial potency of the dose and $f$ is frequency. Then:

$$17)\ k_{Dose}Dose_{ss} = D_0 f$$

and

$$18)\ Dose_{ss} = \frac{D_0 f}{k_{Dose}}$$

So now we just need to substitute this into the equation for $H_{a,b_{ss}}$. Rearranging equation 13) to find $a_{pk_{ss}}$, we find:

$$19)\ a_{pk_{ss}} = \frac{\frac{k_{Dose}D_0 f}{k_{Dose}}}{k_{a,pk}} = \frac{D_0 f}{k_{a,pk}}$$

We can find $b_{pk_{ss}}$ by the same method:

$$20)\ b_{pk_{ss}} = \frac{D_0 f}{k_{b,pk}}$$

Then, substituting equations 19) and 20) into equation 16), we get the steady-state solution:

$$21)\ H_{a,b_{ss}} = \frac{E_{0_a} + \frac{E_{max_a} \cdot \left(\frac{D_0 f}{k_{a,pk}}\right)^{\gamma_a}}{EC_{50_a}^{\gamma_a} + \left(\frac{D_0 f}{k_{a,pk}}\right)^{\gamma_a}} - E_{0_b} - \frac{E_{max_b} \cdot \left(\frac{D_0 f}{k_{b,pk}}\right)^{\gamma_b}}{EC_{50_b}^{\gamma_b} + \left(\frac{D_0 f}{k_{b,pk}}\right)^{\gamma_b}}}{k_H}$$

# 11 Appendix 2

## 11.1 Examples of performing a BFRA (Behavioral Frequency Response Analysis)

We have included our code for performing a BFRA in the 'BFRA.R' file in the Supplementary Materials. The main function to use for BFRA simulations is bfra(). A Bode plot allows you to see how changing the frequency of a behavior leads to different allostatic outcomes. The top two graphs show the Hill equation for the a- and b-processes respectively, describing the relationship between the pharmacokinetic values (x-axis) and the pharmacodynamic effects (y-axis) for both a- and b-processes. The pharmacodynamic effects over time can be observed in the middle two graphs, which are simulations of the opponent processes generated by performing behaviors at different frequencies. The final graph is the Bode plot, displaying the steady-state value of the final hedonic compartment of the model.

The default simulation is below, followed by simulations with varying parameters to modify the shape of the opponent processes. Default parameters are: k_Dose=1, k_apk=0.02, k_bpk=0.004, k_apd=1, k_bpd=1, k_H=1, E0_a=0, Emax_a=1, EC50_a=1, gamma_a=2, E0_b=0, Emax_b=3, EC50_b=9, and gamma_b=2.

For all figures with multiple colors, the first parameter in the list passed to the function corresponds to the darkest color in the graph.

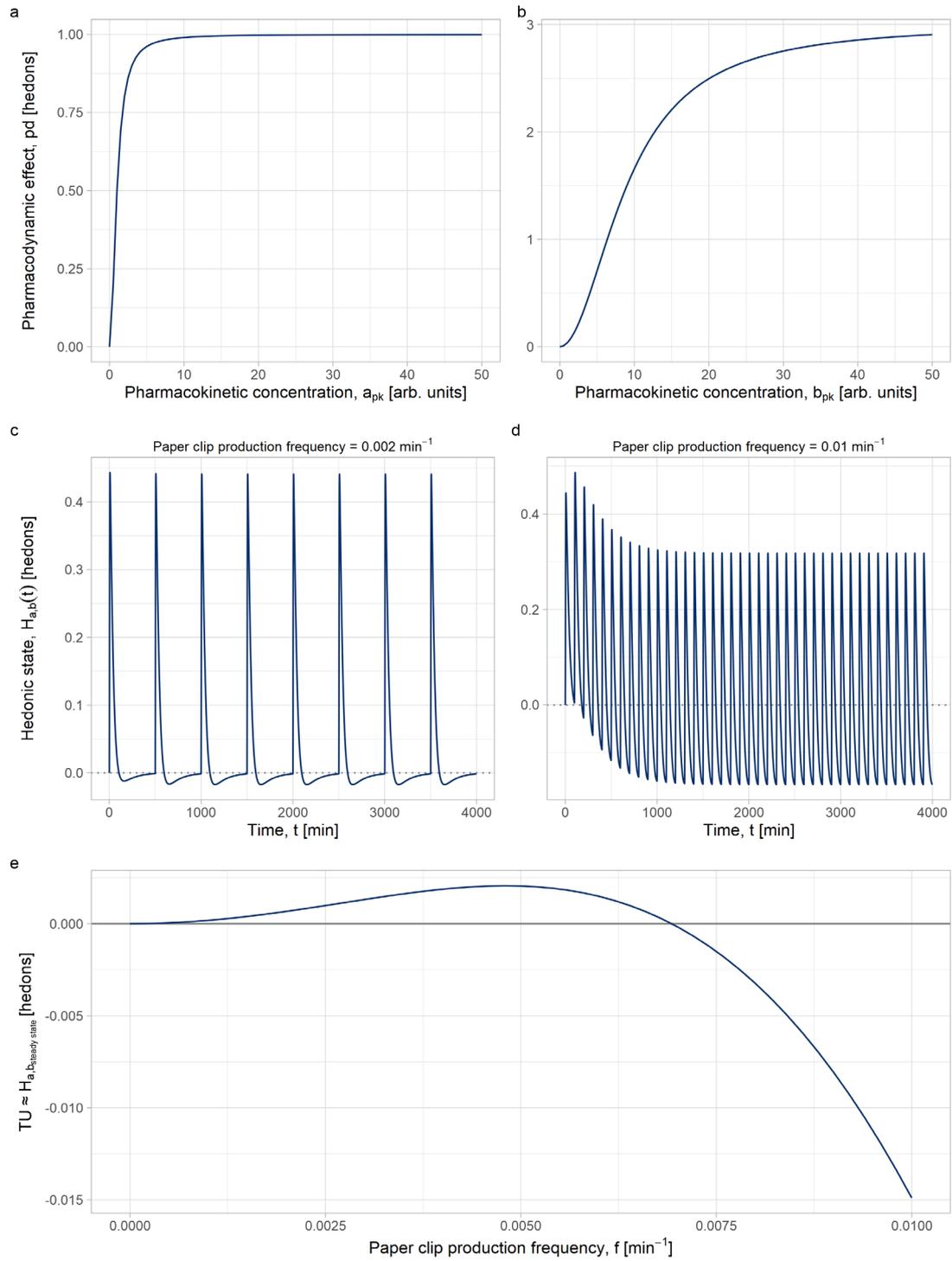

Figure 1. BFRA simulation produced with the following R command = bfra()

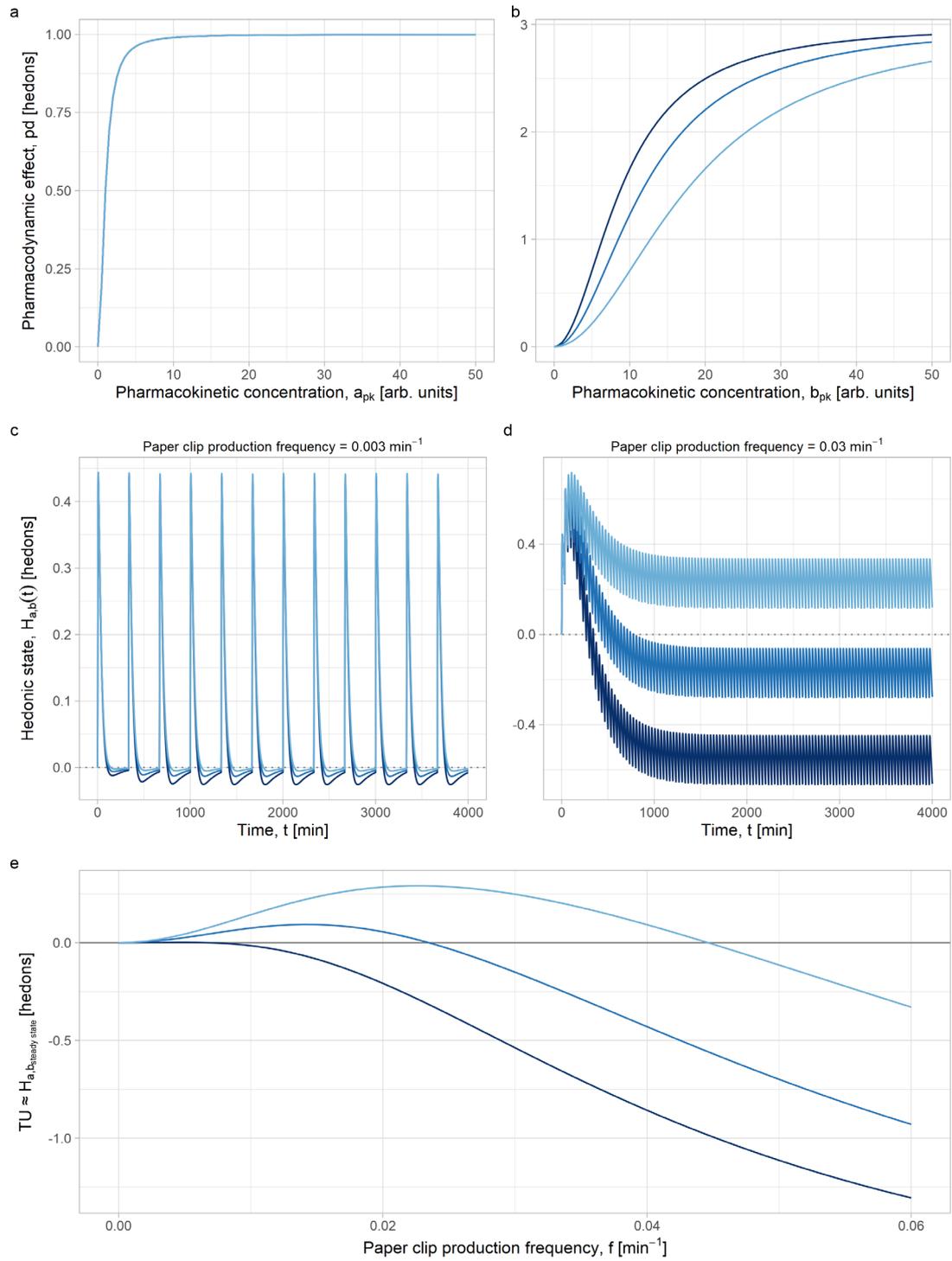

Figure 2. BFRA simulation produced with the following R command = bfra(EC50_b=c(9,12,18), seq_1=0.001, seq_2=0.06, plot_2=c(0.003, 0.03))

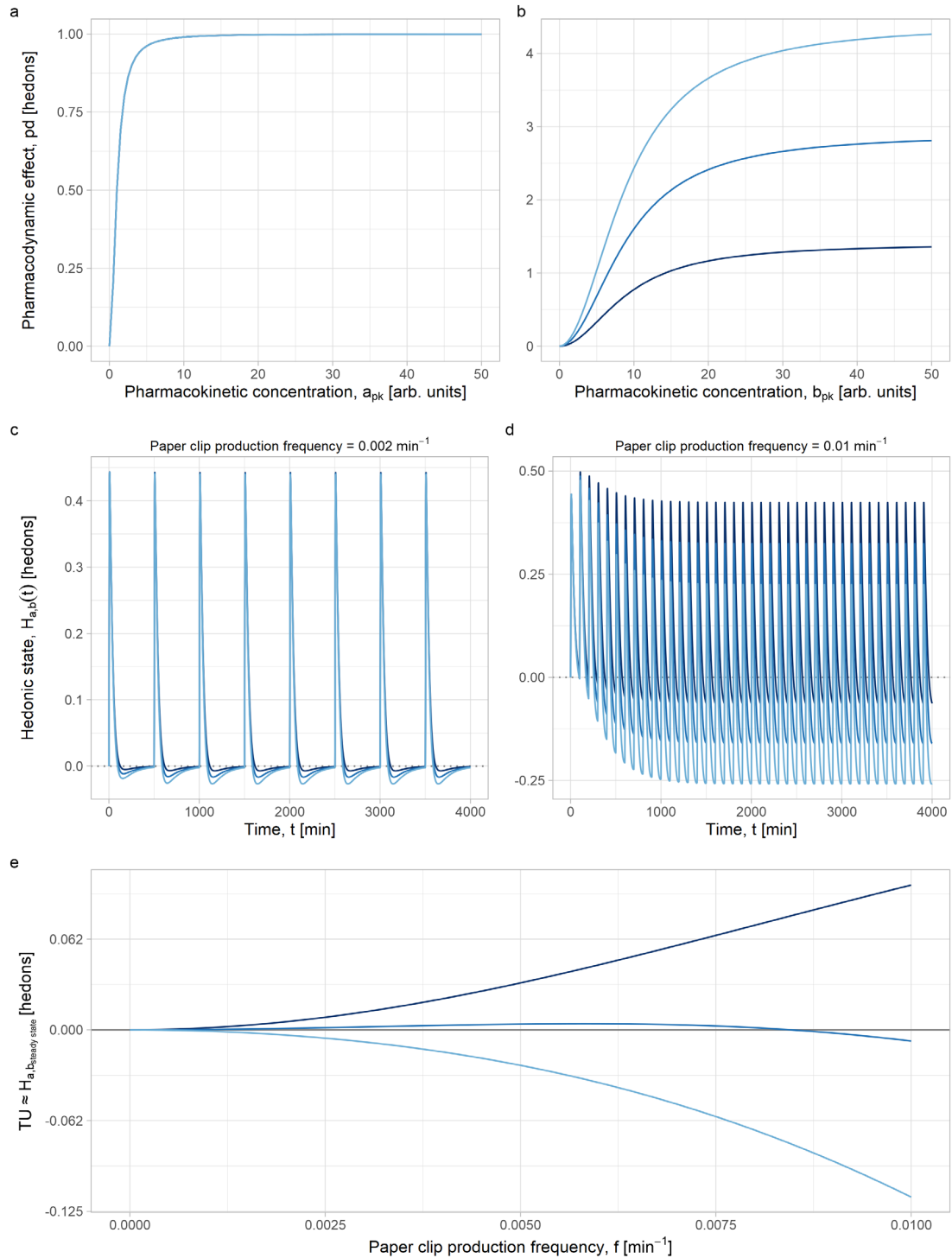

*Figure 3. BFRA simulation produced with the following R command = bfra(Emax_b=c(1.4, 2.9, 4.4))*

You can also include the simulated BFRA results, taking the integral of the H compartment across the simulation time. Note that the numerical solution curves (derived from the integrals of the simulations) are slightly higher than the analytical steady-state solution curves. This is because of the bounded nature of the simulation and the fact that b-processes take longer to decay than a-processes, meaning that a larger portion of b-processes is being eliminated from the integral.

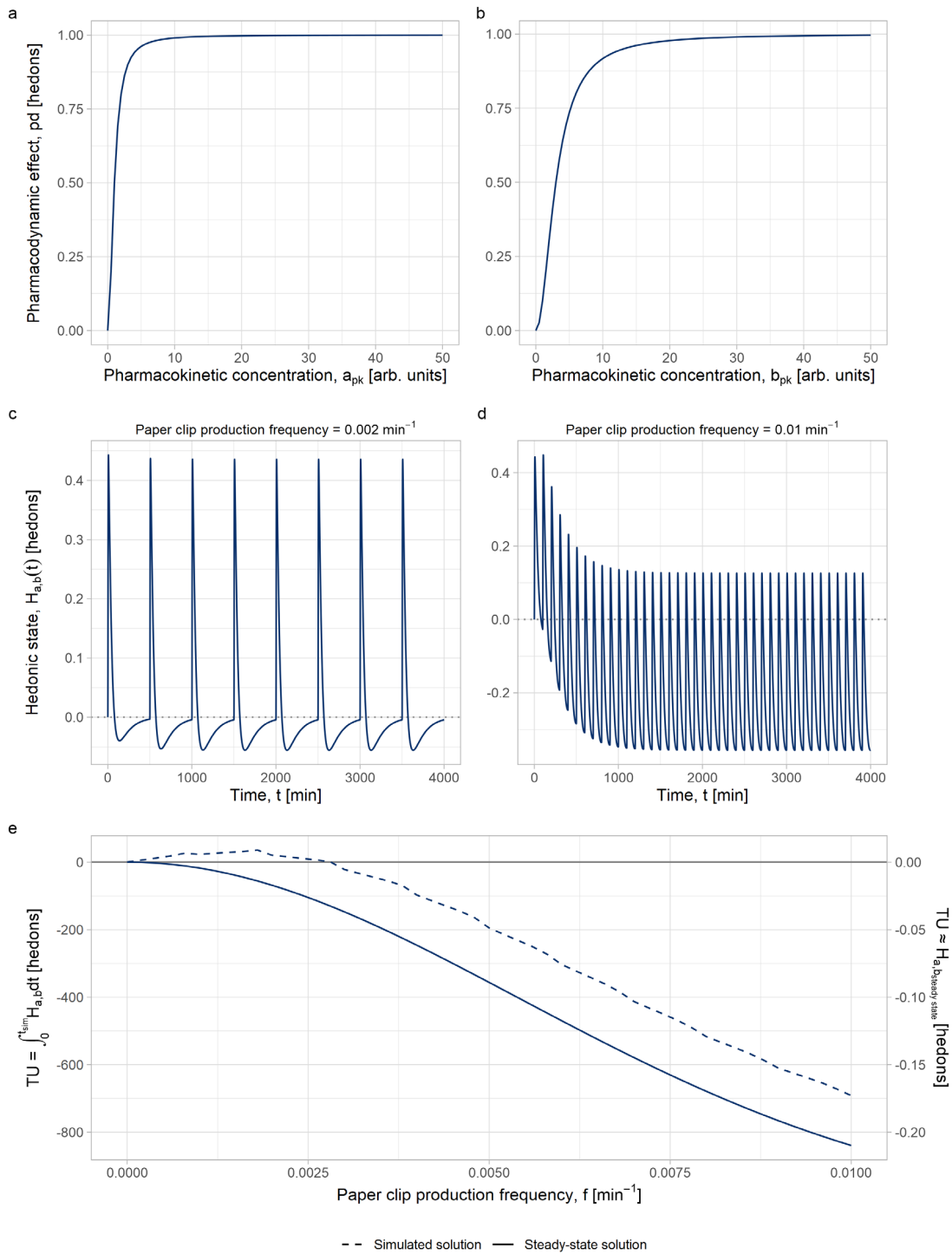

Figure 4. BFRA simulation produced with the following R command = bfra(Emax_b=1, EC50_b=3, include_simulated_results=TRUE)

### 11.1.1 Hormesis

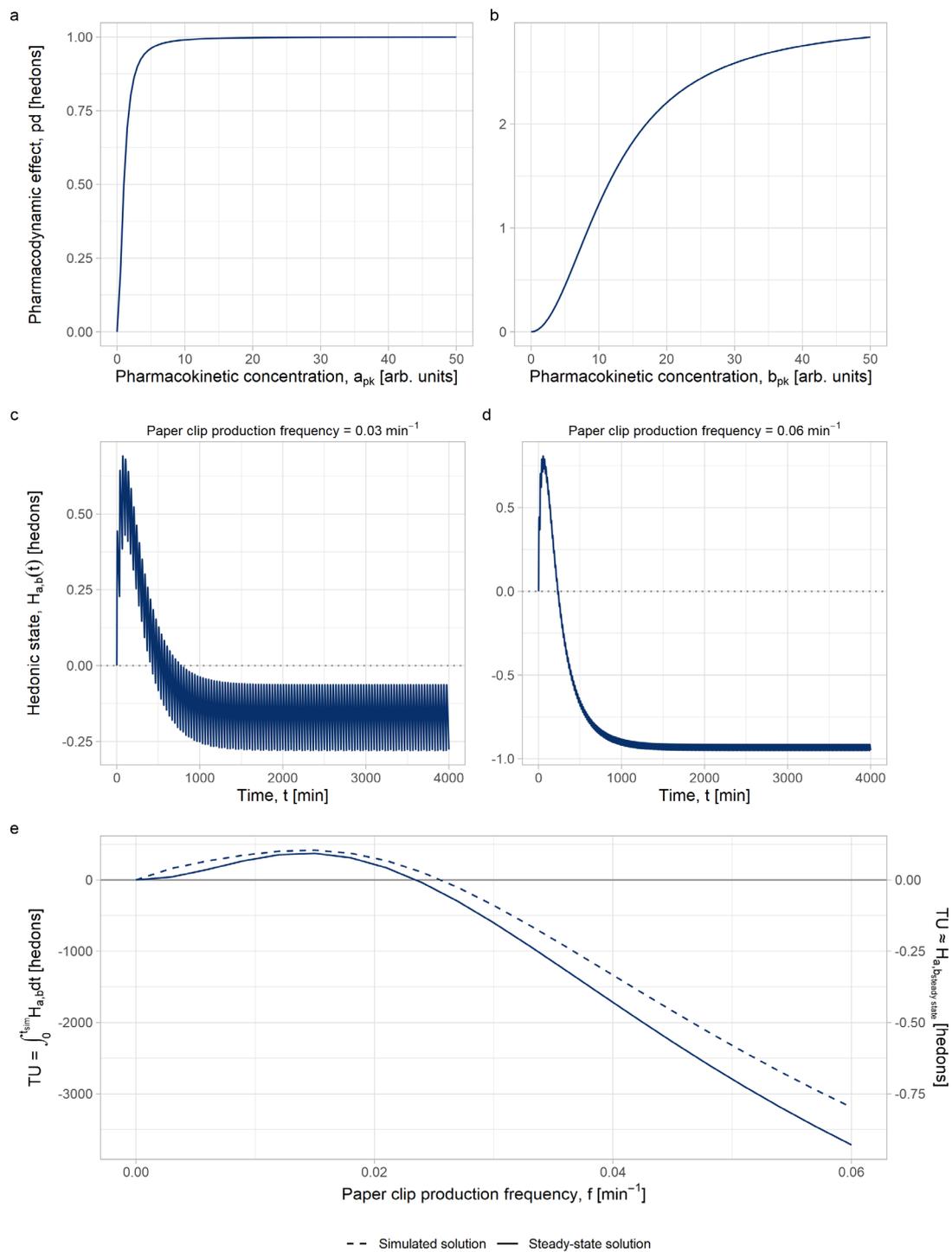

*Figure 5. BFRA simulation produced with the following R command = bfra(EC50_b=12, seq_1=0.003, seq_2=0.06, plot_2=c(0.03, 0.06), include_simulated_results=TRUE). Note how seq_1, seq_2 and plot_2 have been modified to change the frequency range of the Bode plot.*

## 11.1.2 Triphasic hormesis

Triphasic hormesis occurs when there is a faux hormetic limit at a low behavioral frequency; in fact, the behavior becomes positive again at extremely high frequencies. This only occurs for a small subset of opponent process parameterizations, but should be avoided where possible as it can lead to paperclip maximizer scenarios.

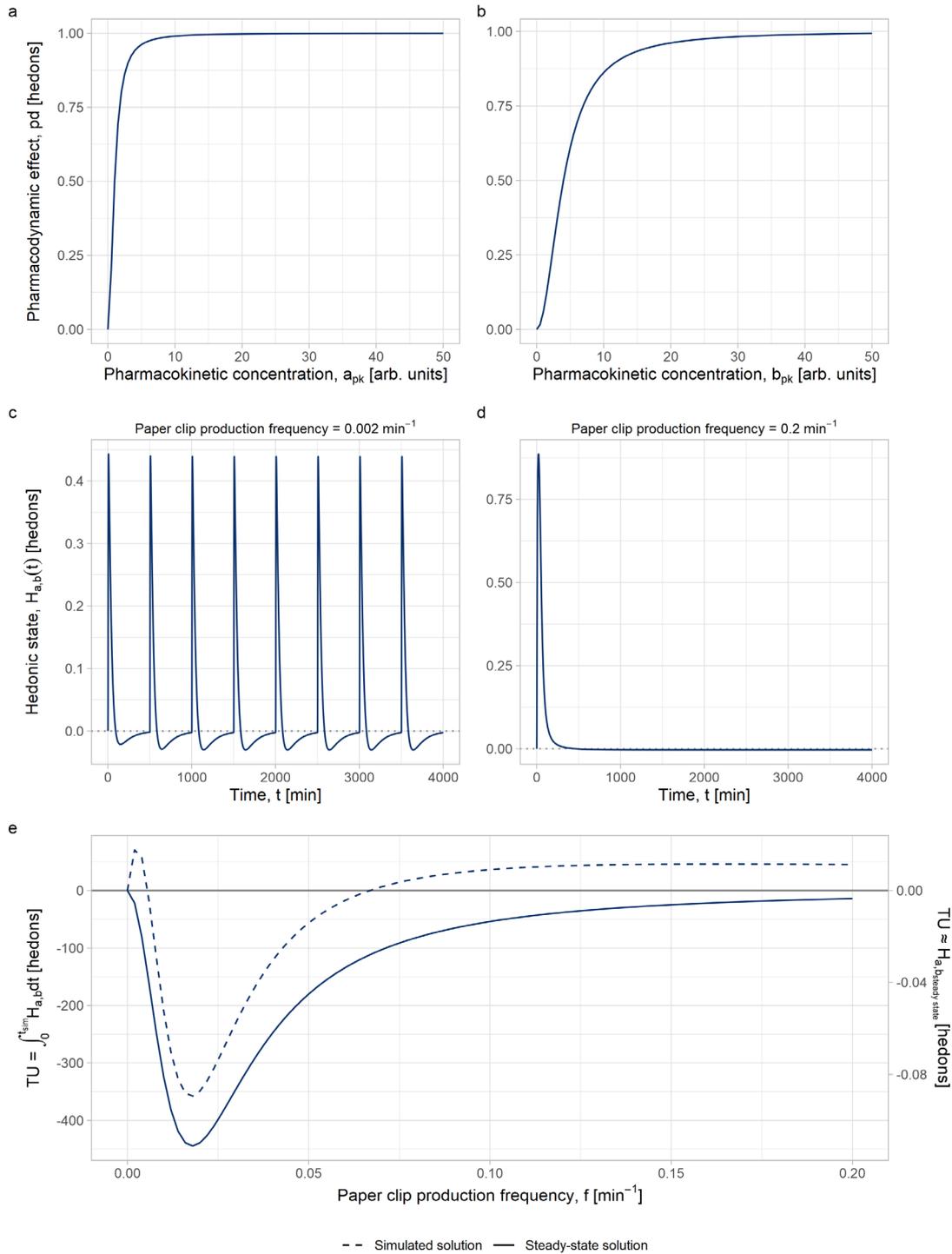

*Figure 6. BFRA simulation produced with the following R command = bfra(Emax_b=1, EC50_b=4, seq_1=0.002, seq_2=0.2, plot_2=c(0.002, 0.2), include_simulated_results=TRUE)*

### 11.1.3 Modeling behavioral bursts

The BCRA and BFRA methods can be combined by modifying the frequency of behavioral bursts. Note that the simulated solution diverges from the steady-state solution at higher behavioral frequencies, since steady state is not reached at these frequencies.

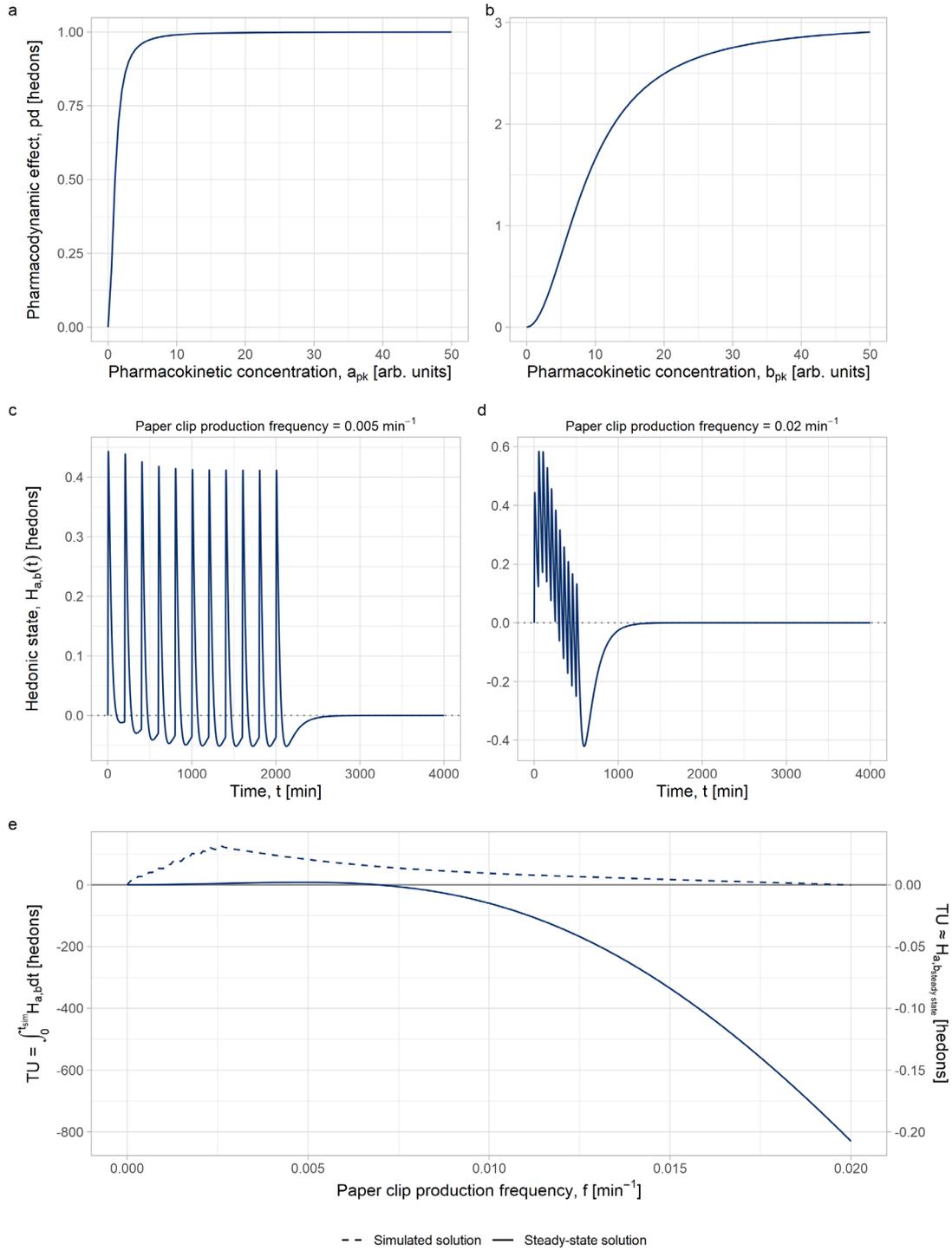

*Figure 7. BFRA simulation produced with the following R command = bfra(addl=10, seq_1=0.0001, seq_2=0.02, plot_2=c(0.005, 0.02), include_simulated_results = TRUE)*

### 11.1.3.1 Changing the length of behavioral bursts

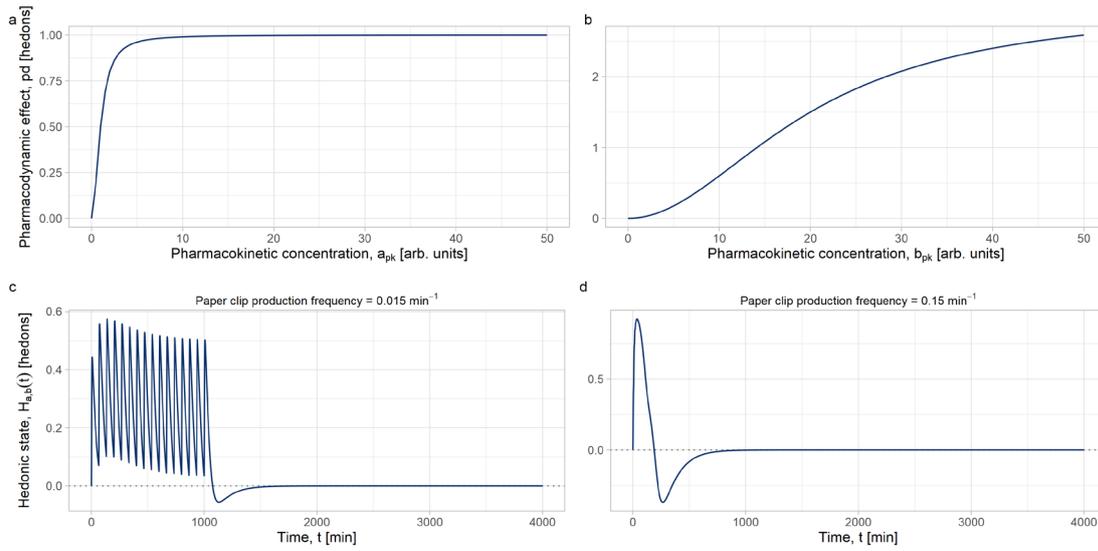

*Figure 8. BFRA simulation produced with the following R command = bfra(Emax_b=3, EC50_b=20, plot_2=c(0.015, 0.15), seq_1=0.015, seq_2=0.15, addl=15, plot_bfra_graph=FALSE)*

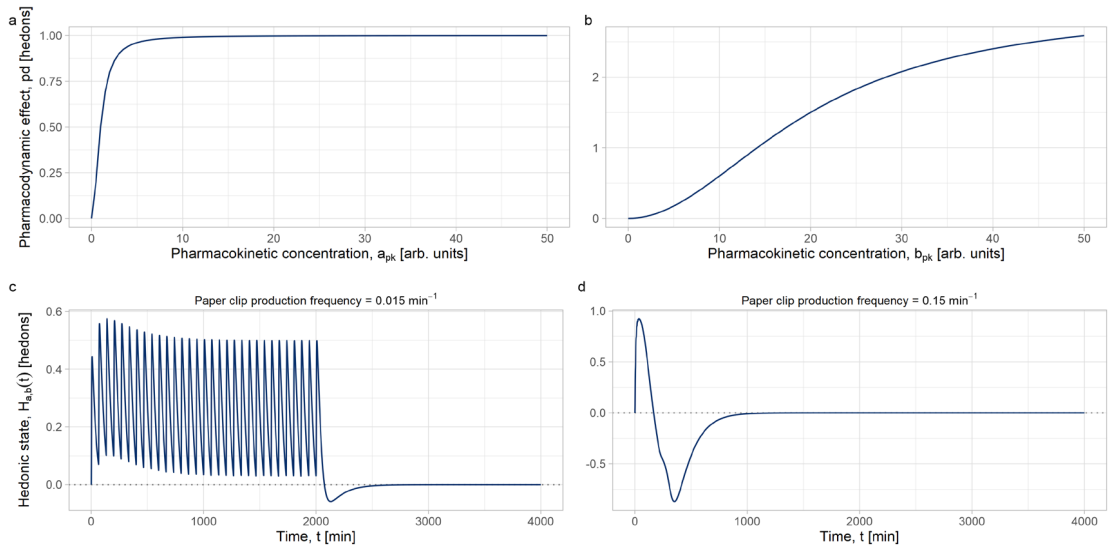

*Figure 9. BFRA simulation produced with the following R command = bfra(Emax_b=3, EC50_b=20, plot_2=c(0.015, 0.15), seq_1=0.015, seq_2=0.15, addl=30, plot_bfra_graph=FALSE)*

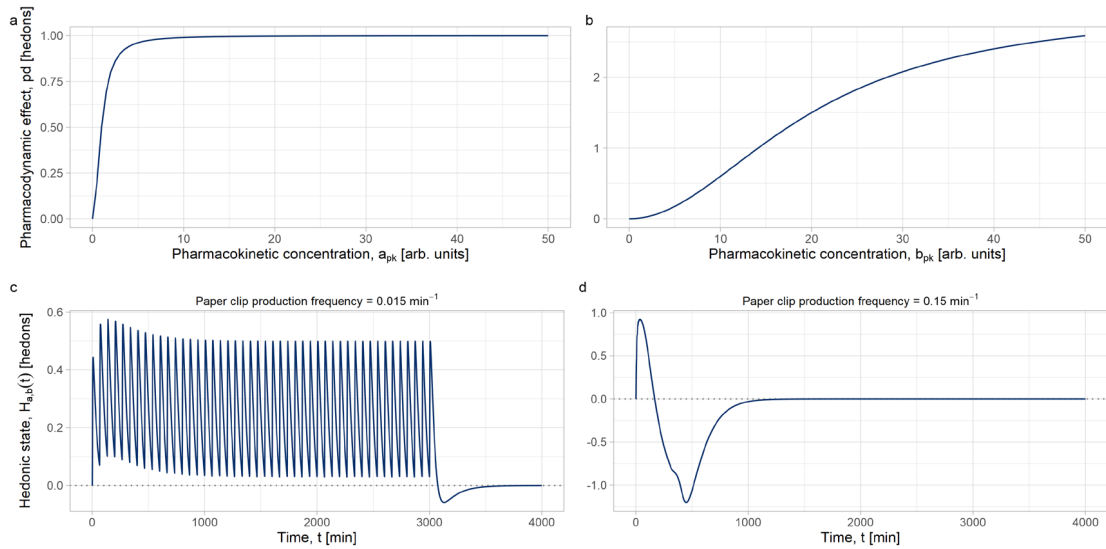

*Figure 10. BFRA simulation produced with the following R command = bfra(Emax_b=3, EC50_b=20, plot_2=c(0.015, 0.15), seq_1=0.015, seq_2=0.15, addl=45, plot_bfra_graph=FALSE)*

11.1.4 Multiple solutions

You can test how varying multiple parameters simultaneously affects the opponent processes, and in turn affects the hormetic curve. For example, you can pass two vectors to modify both Emax_b and EC50_b simultaneously.

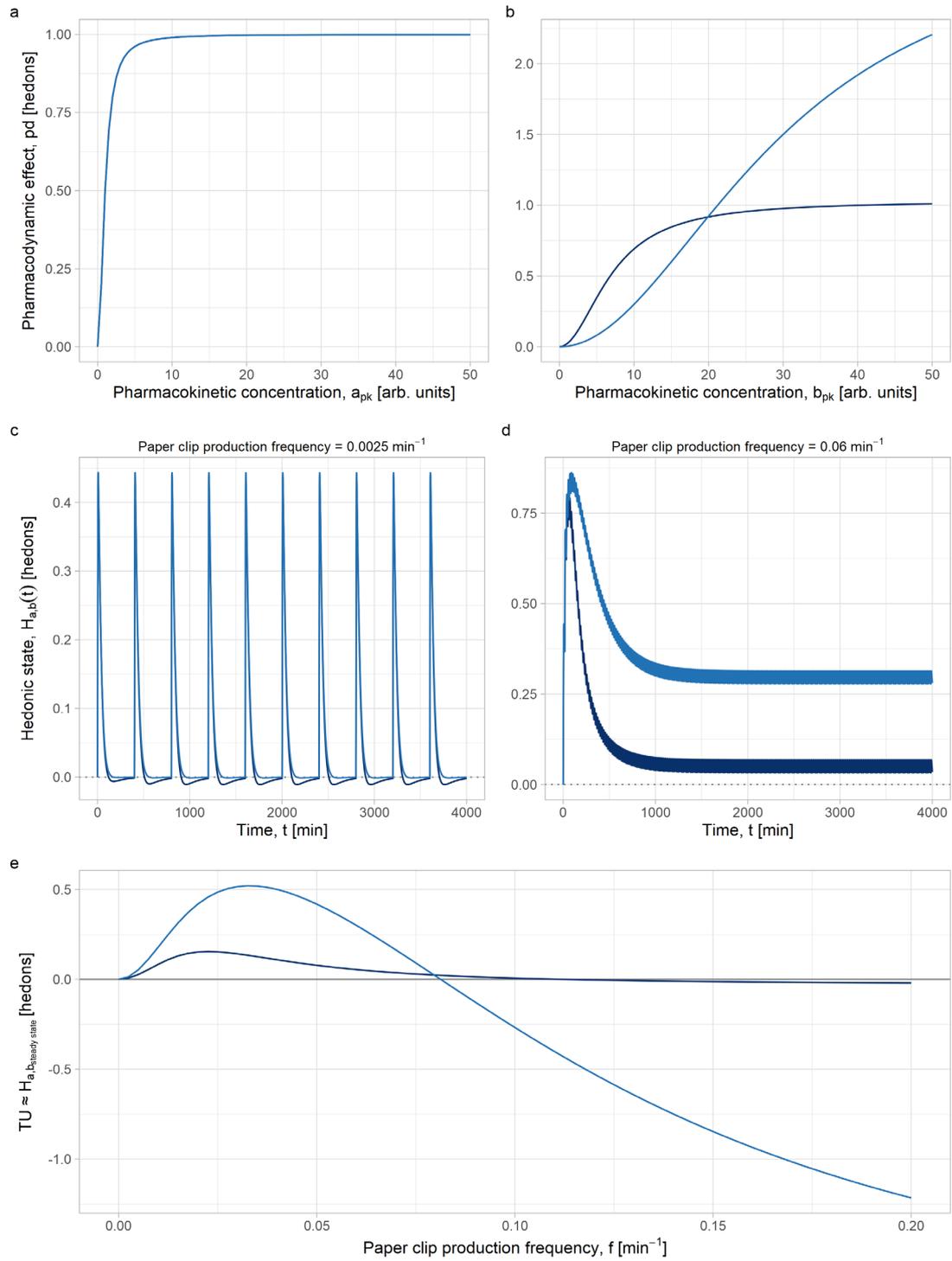

*Figure 11. BFRA simulation produced with the following R command = bfra(Emax_b=c(1.03, 3), EC50_b=c(7, 30), seq_1=0.0025, seq_2 = 0.2, plot_2=c(0.0025, 0.06))*

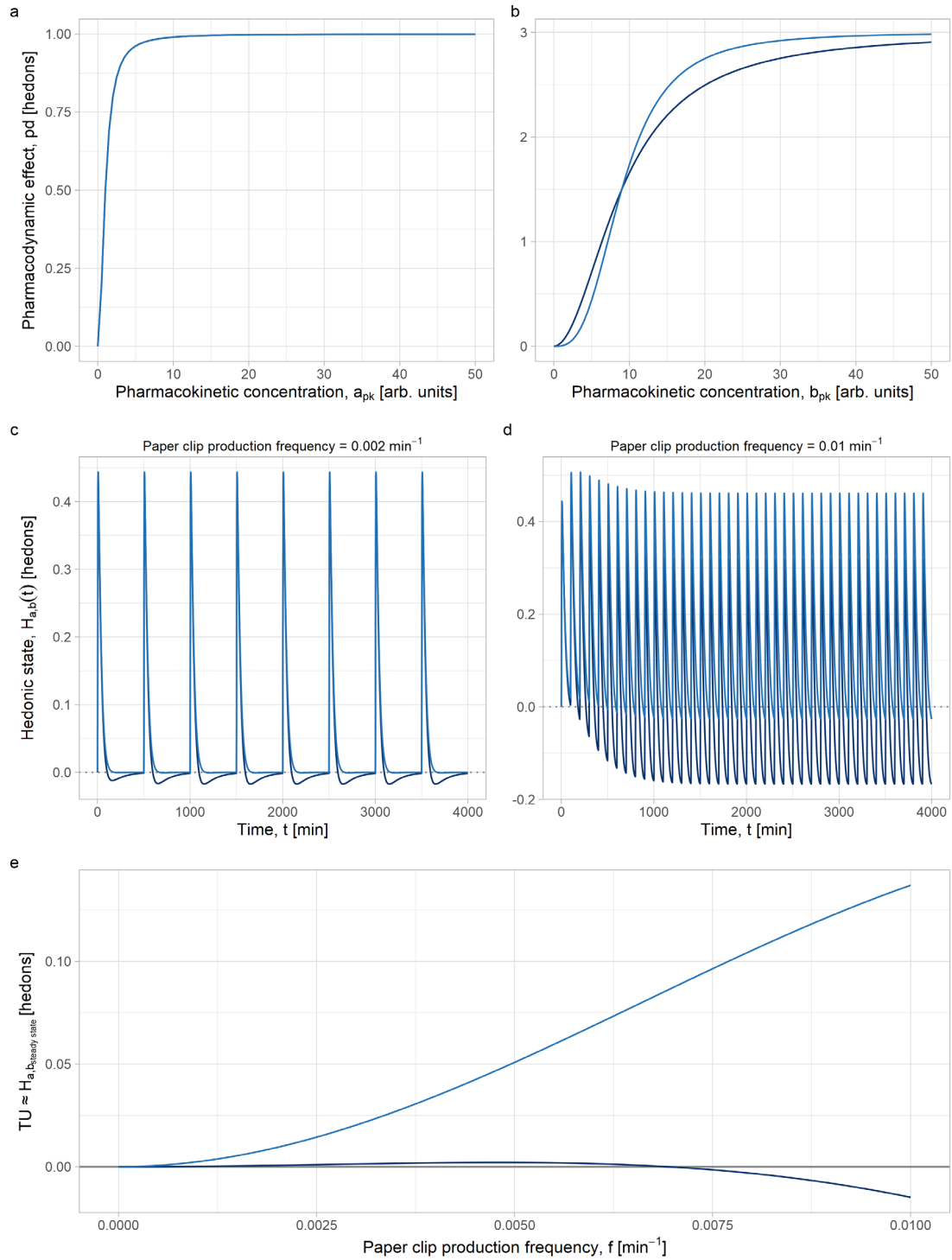

*Figure 12. BFRA simulation produced with the following R command = bfra(gamma_b=c(2, 3))*

### 11.1.5 Pharmacokinetic perturbations

From now on, only analytical steady-state solutions will be presented in the Bode plot.

### 11.1.5.1 Modifying a-process pharmacokinetic decay constant

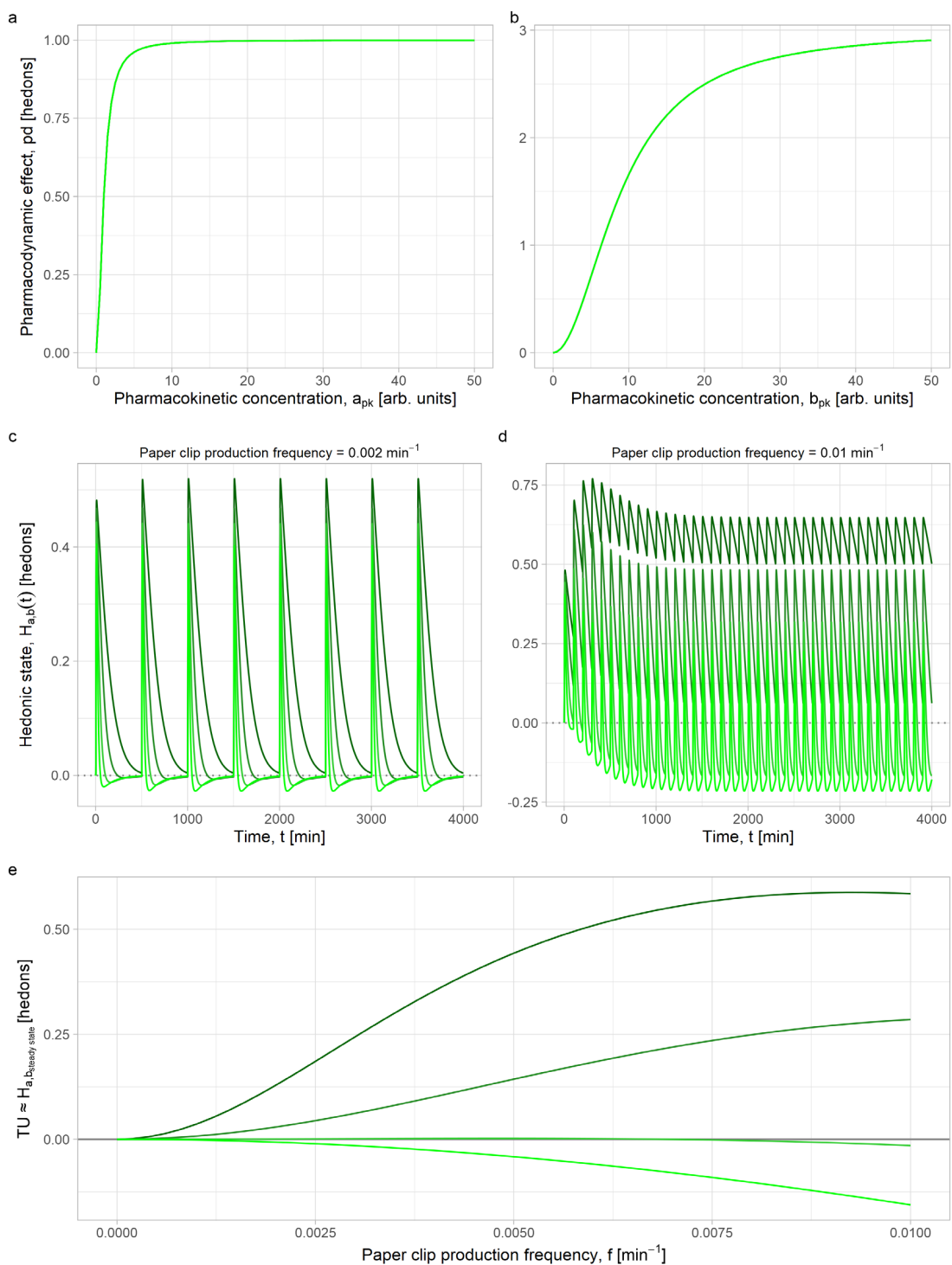

*Figure 13. BFRA simulation produced with the following R command = bfra(k_apk=c(0.005, 0.01, 0.02, 0.04), colorscheme=2)*

### 11.1.5.2 Modifying b-process pharmacokinetic decay constant

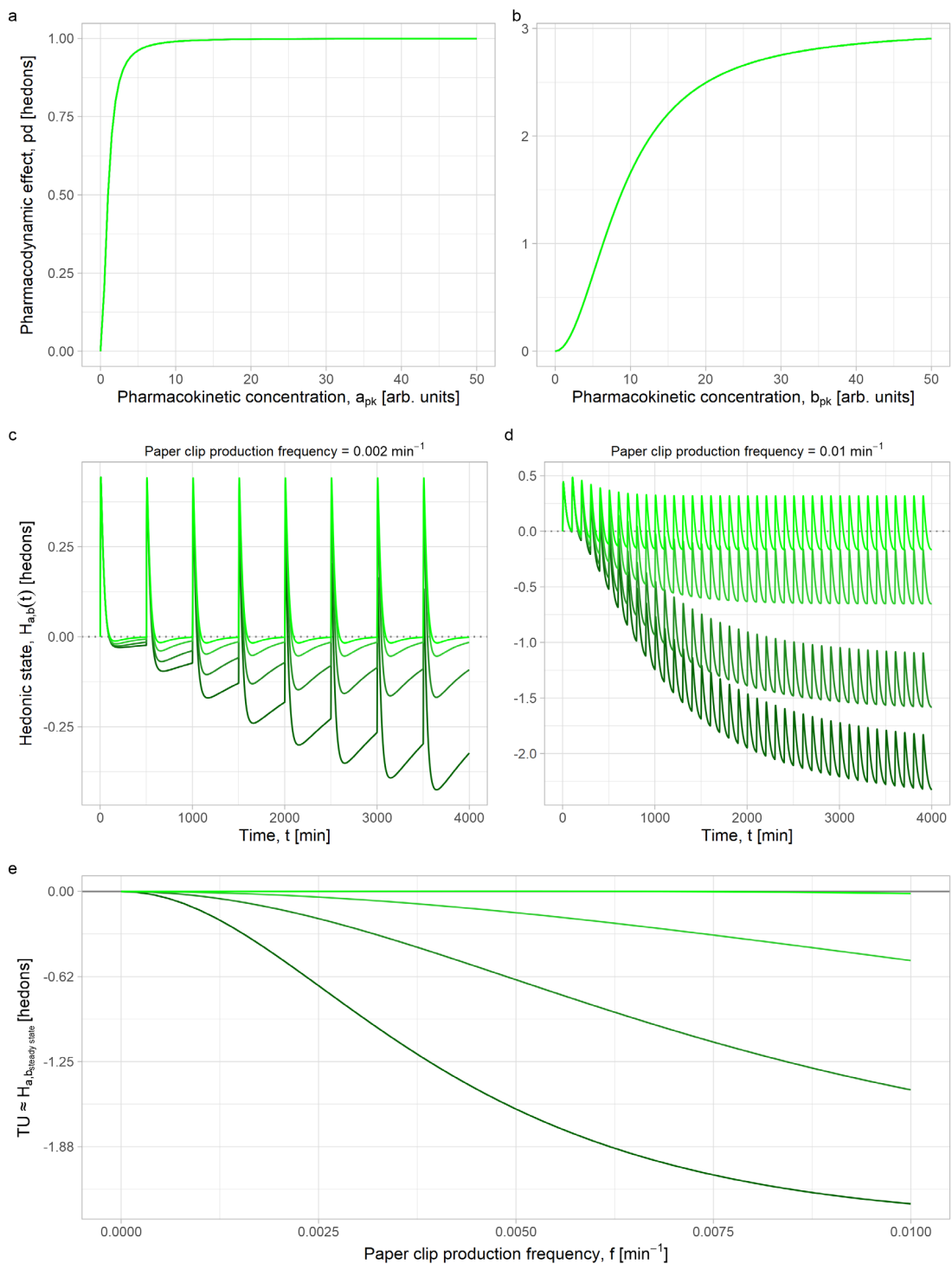

Figure 14. BFRA simulation produced with the following R command = bfra(k_bpk=c(0.0005, 0.001, 0.002, 0.004), colorscheme=2). Note that the top two Hill equation graphs remain unchanged.

## 11.1.6 Pharmacodynamic perturbations

Note that the analytical steady-state solution is calculated in terms of the pharmacokinetic decay constants for the current version of bfra(), but not the pharmacodynamic ones. This may be rectified in future versions. Hence, for the graphs below in which k_apd and k_bpd are modified, the BFRA graph is not plotted.

*11.1.6.1 Modifying a-process pharmacodynamic decay constant*

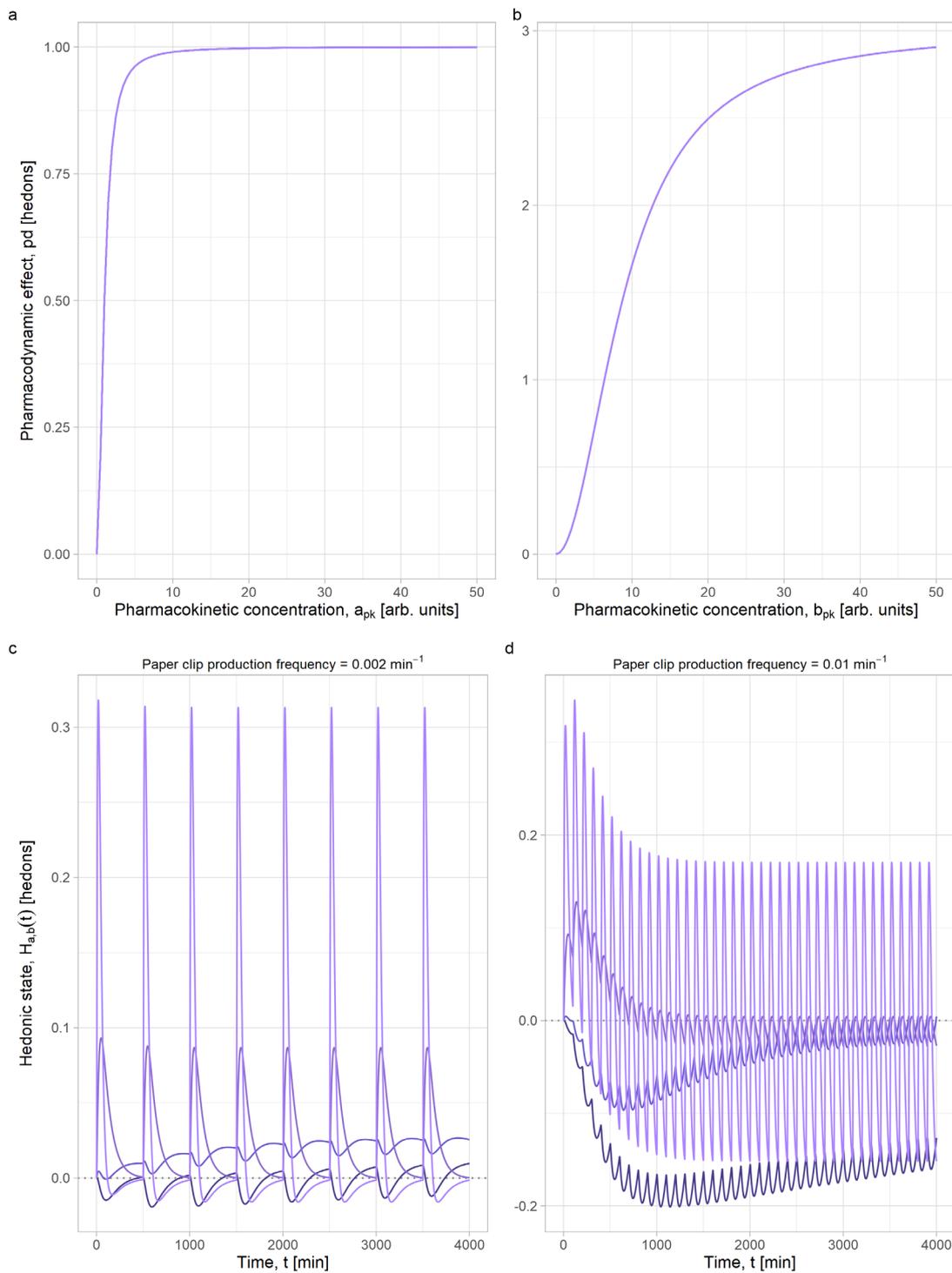

*Figure 15. BFRA simulation produced with the following R command = bfra(k_apd=c(0.0001, 0.001, 0.01, 0.1), colorscheme=3)*

### 11.1.6.2 Modifying b-process pharmacodynamic decay constant

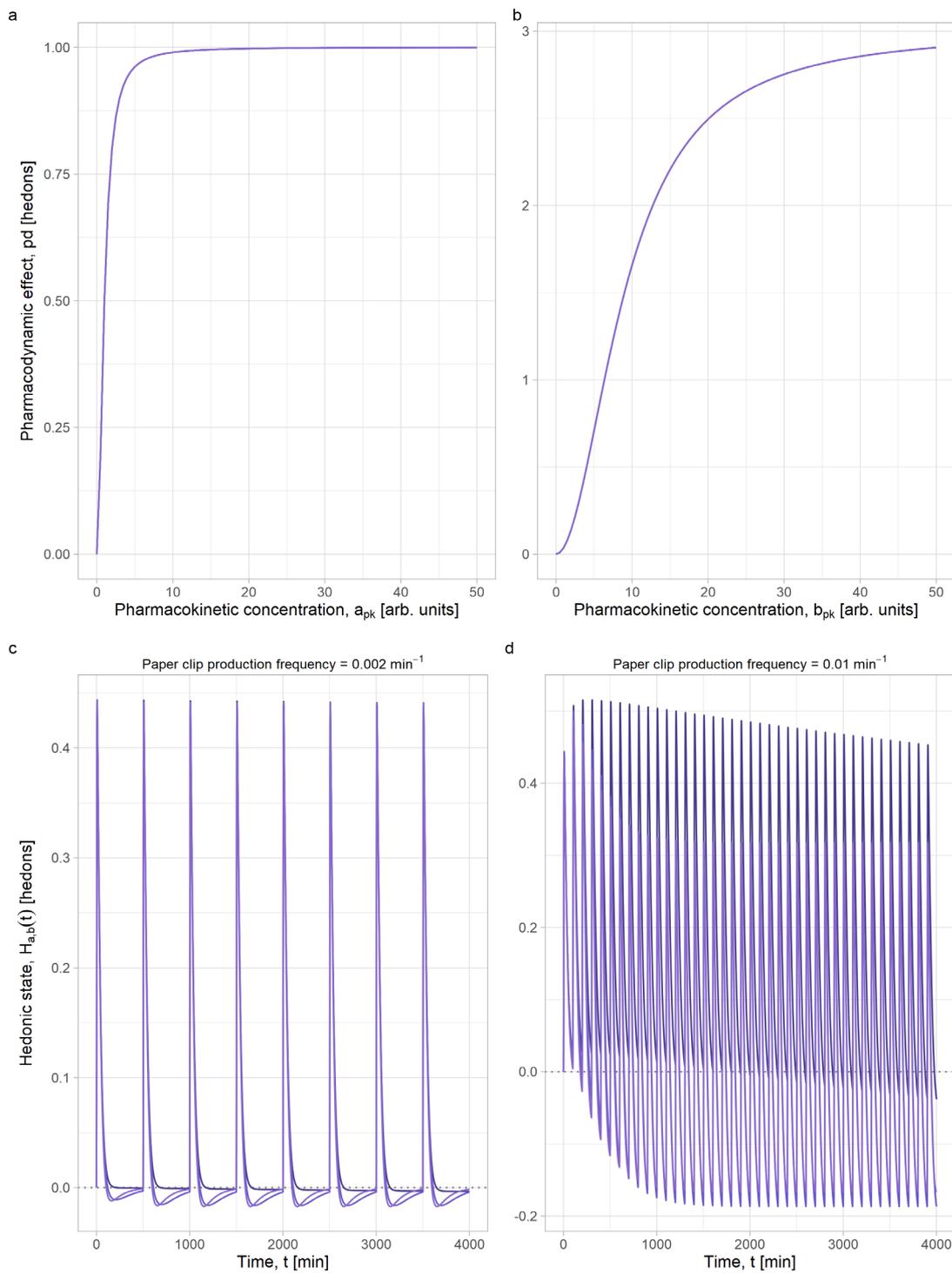

*Figure 16. BFRA simulation produced with the following R command = bfra(k_bpd=c(0.0001, 0.01, 1), colorscheme=3)*

### 11.1.6.3 Modifying hedonic compartment (H) decay constant

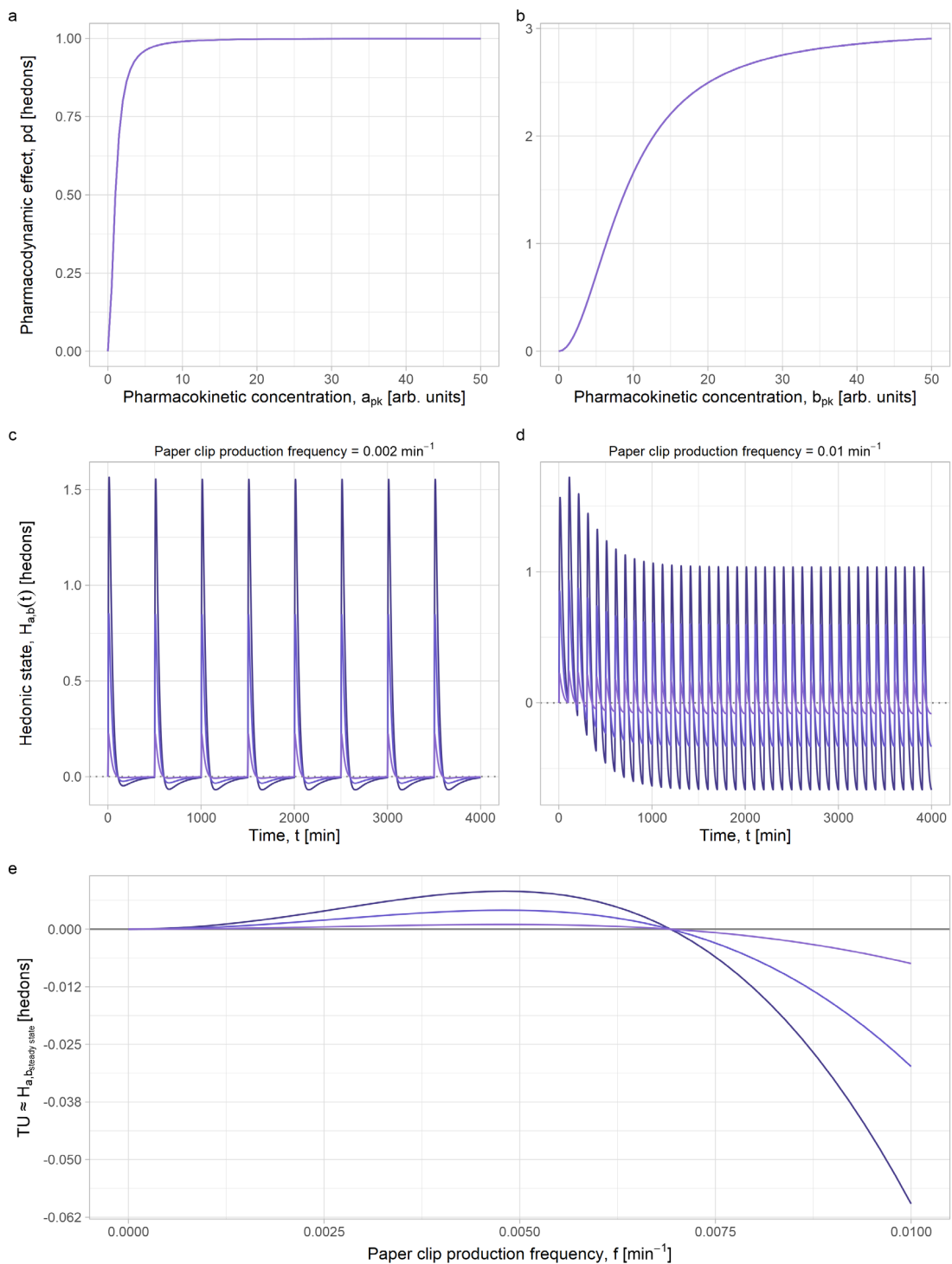

Figure 17. BFRA simulation produced with the following R command = bfra(k_H=c(0.25, 0.5, 2), colorscheme=3)

## 11.1.7 Hill equation parameters

### 11.1.7.1 Modifying E0_a

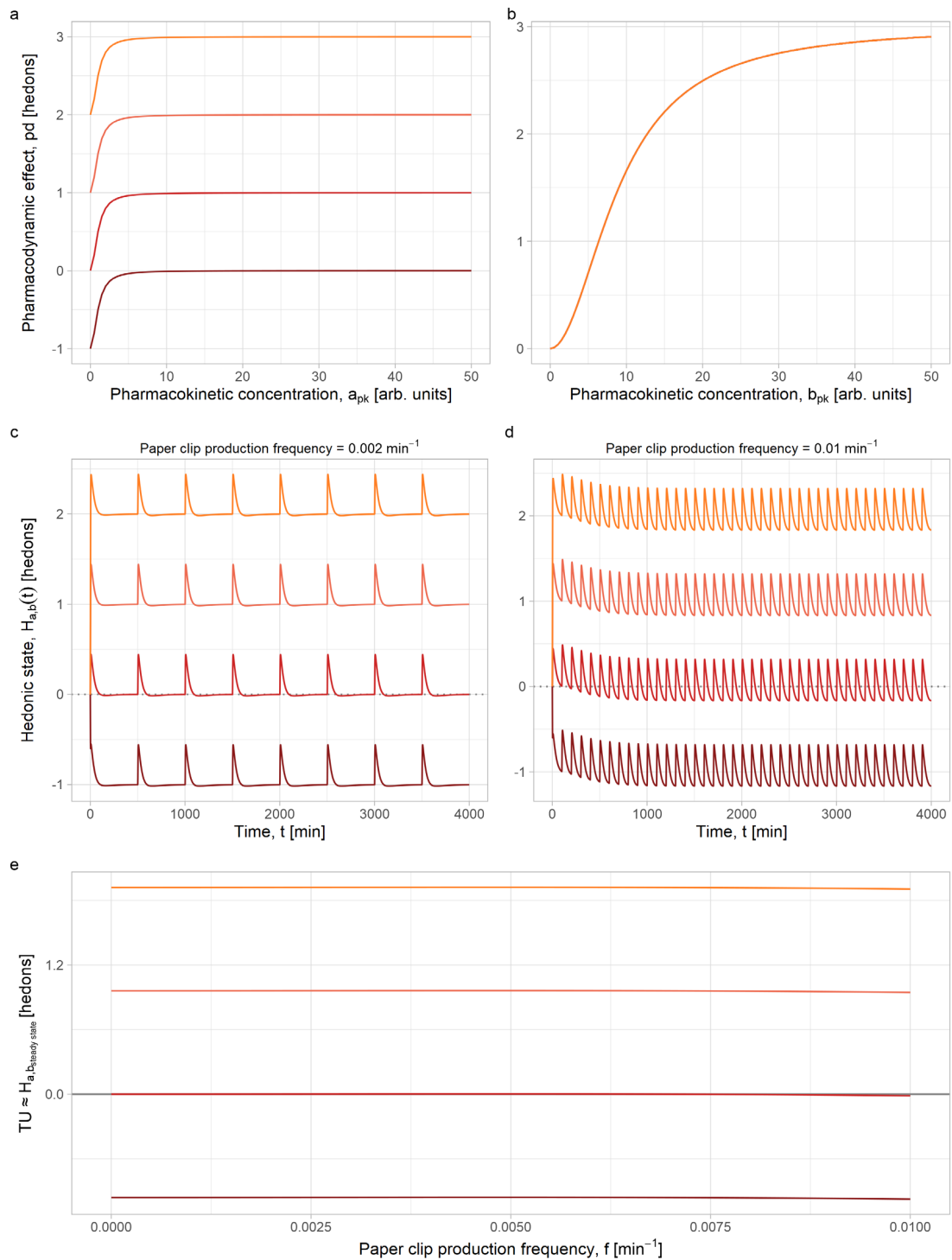

Figure 18. BFRA simulation produced with the following R command = bfra(E0_a=c(-1, 0, 1, 2), colorscheme=4)

### 11.1.7.2 Modifying Emax_a

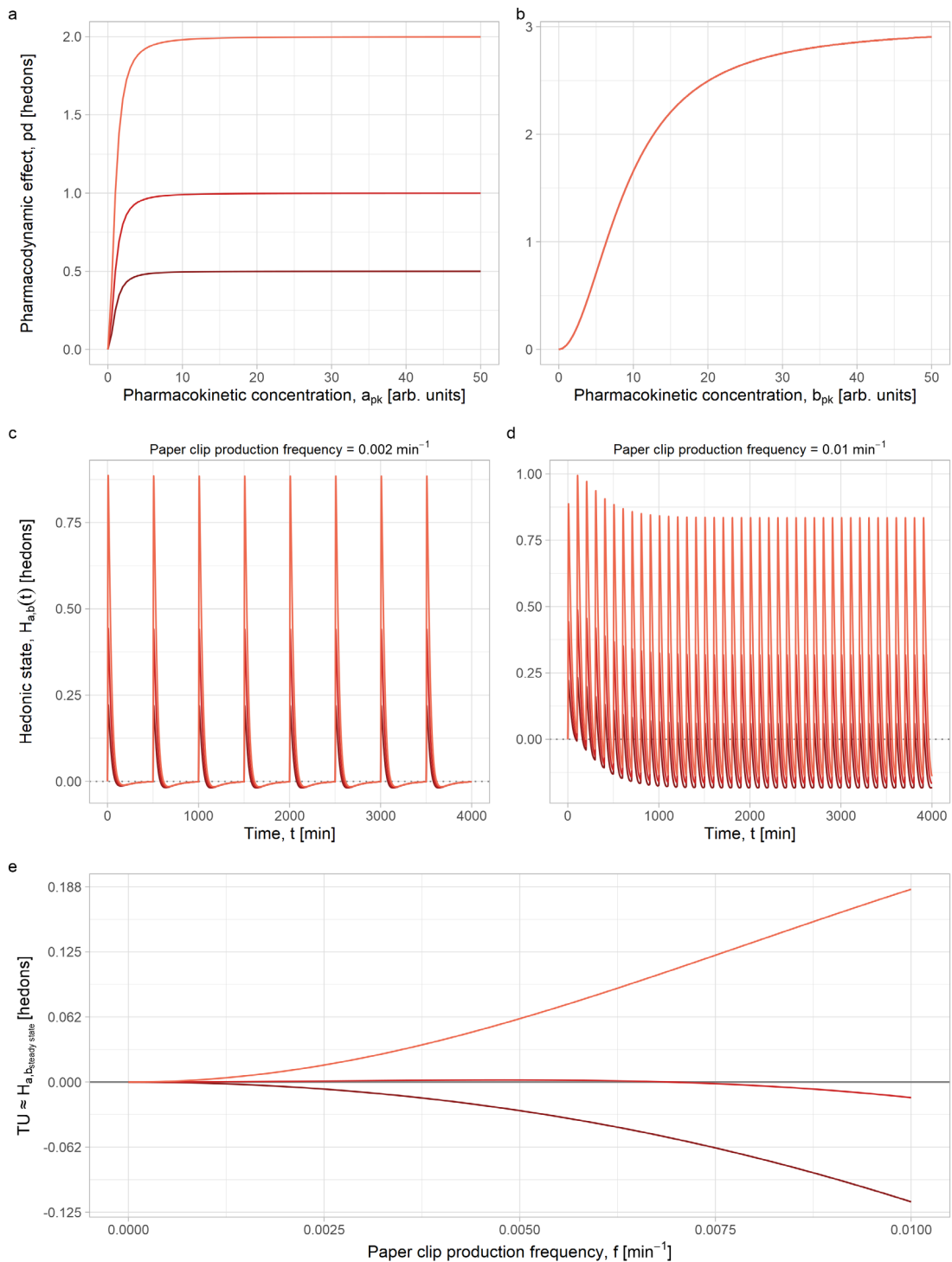

Figure 19. BFRA simulation produced with the following R command = bfra(Emax_a=c(0.5, 1, 2), colorscheme=4)

### 11.1.7.3 Modifying EC50_a

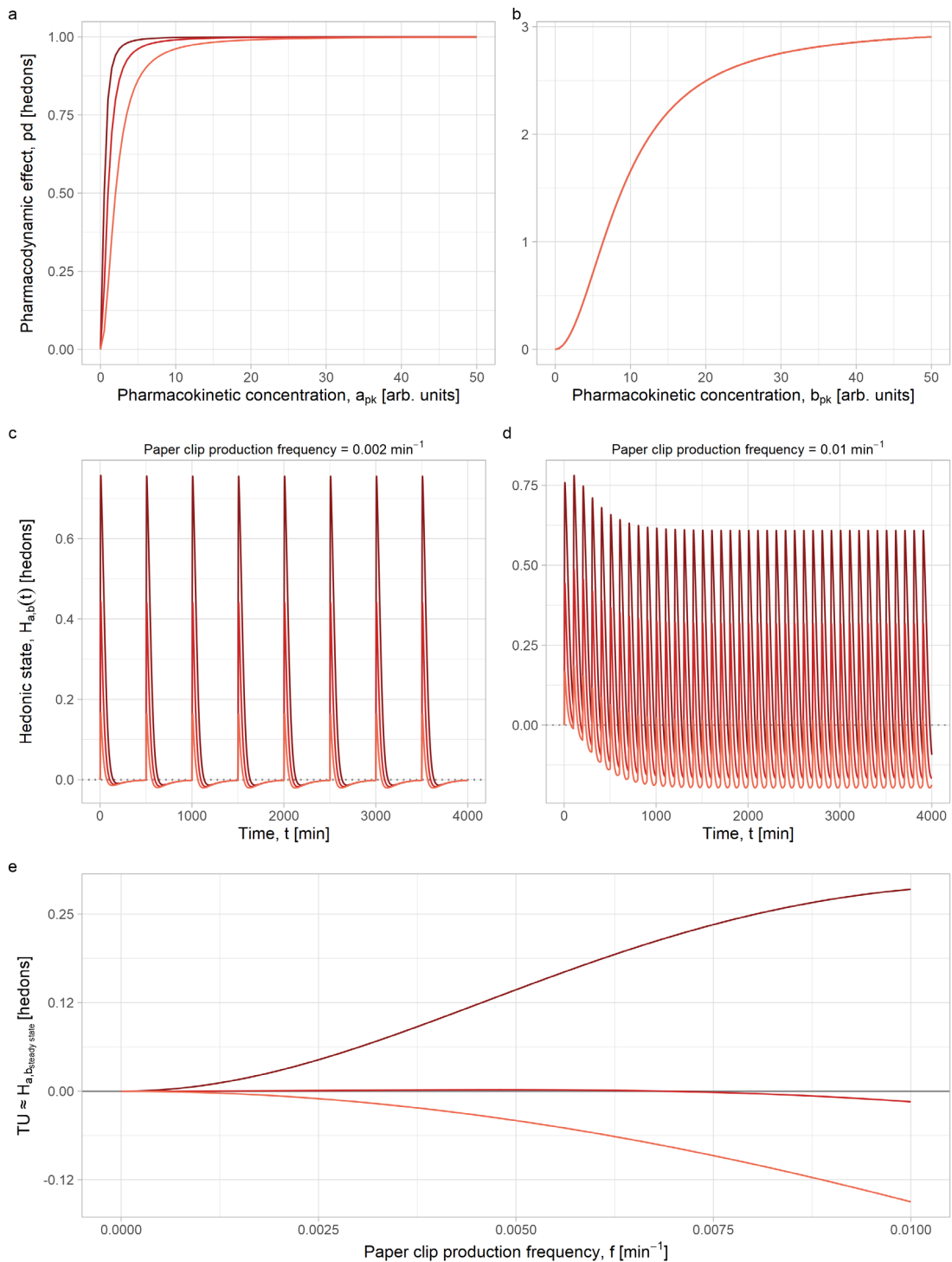

Figure 20. BFRA simulation produced with the following R command = bfra(EC50_a=c(0.5, 1, 2), colorscheme=4)

### 11.1.7.4 Modifying gamma_a

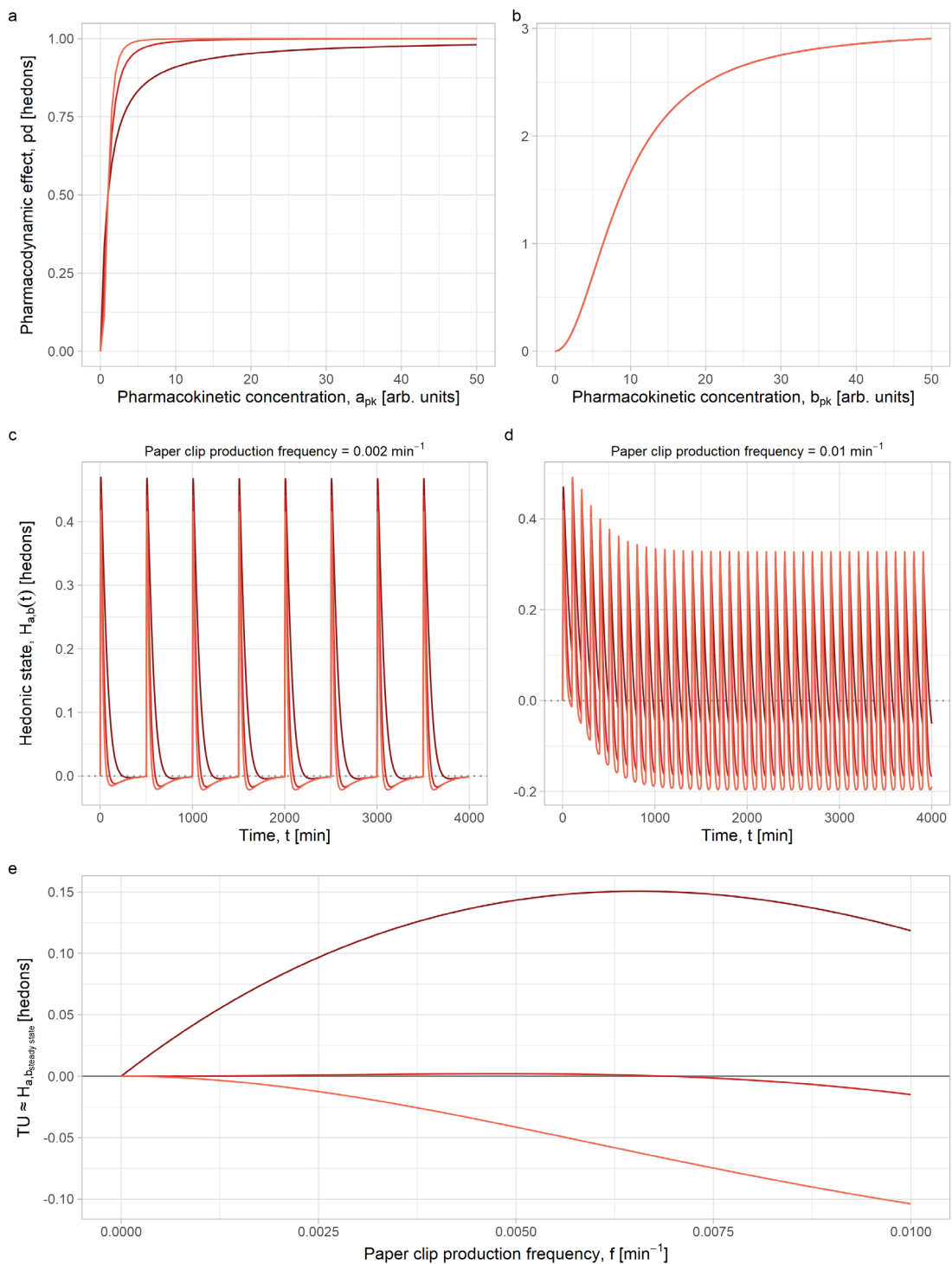

Figure 21. BFRA simulation produced with the following R command = bfra(gamma_a=c(1, 2, 3), colorscheme=4)

### 11.1.7.5 Modifying E0_b

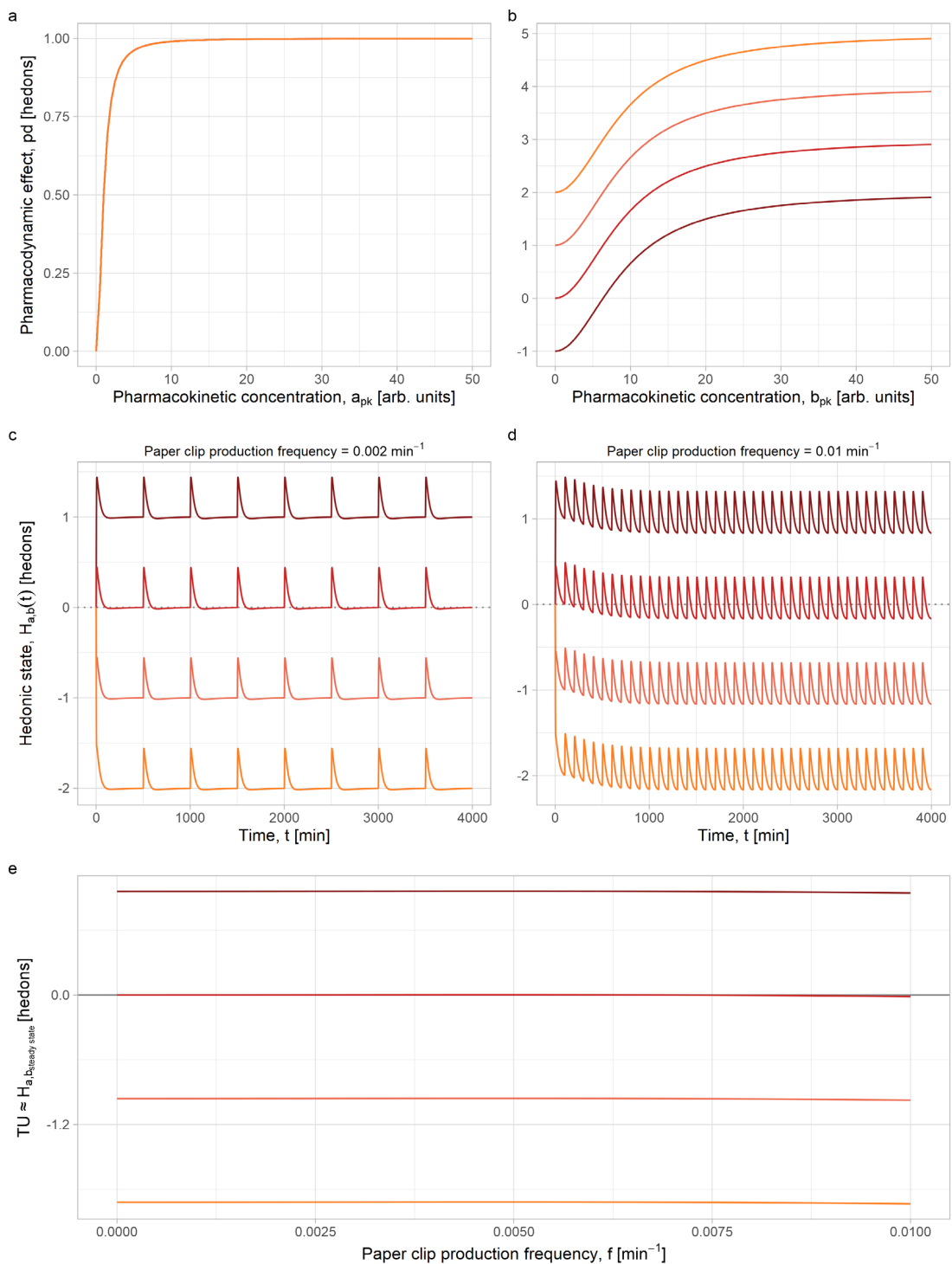

Figure 22. BFRA simulation produced with the following R command = bfra(E0_b=c(-1, 0, 1, 2), colorscheme=4)

### 11.1.7.6 Modifying Emax_b

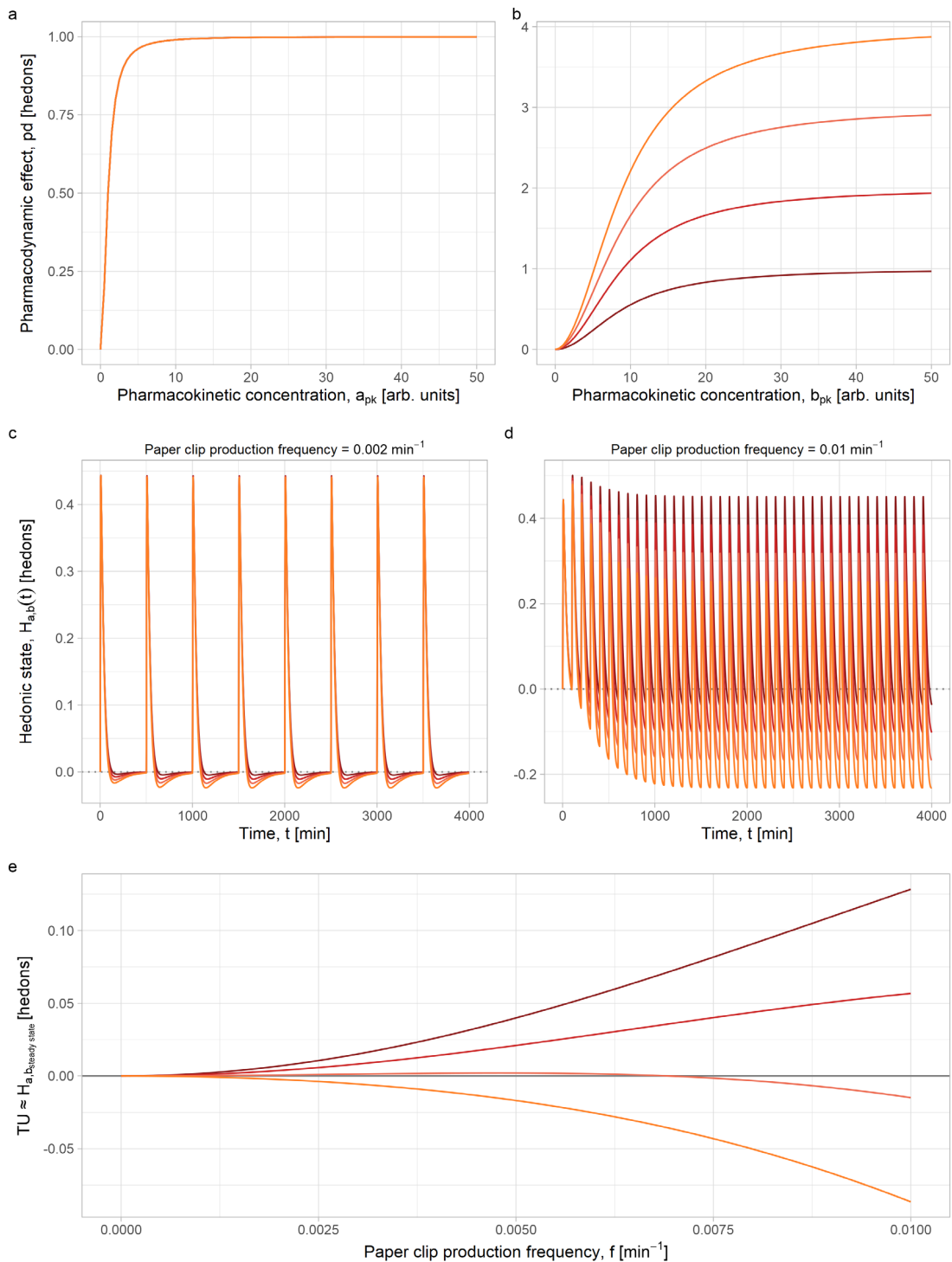

Figure 23. BFRA simulation produced with the following R command = bfra(Emax_b=c(1, 2, 3, 4), colorscheme=4)

### 11.1.7.7 Modifying EC50_b

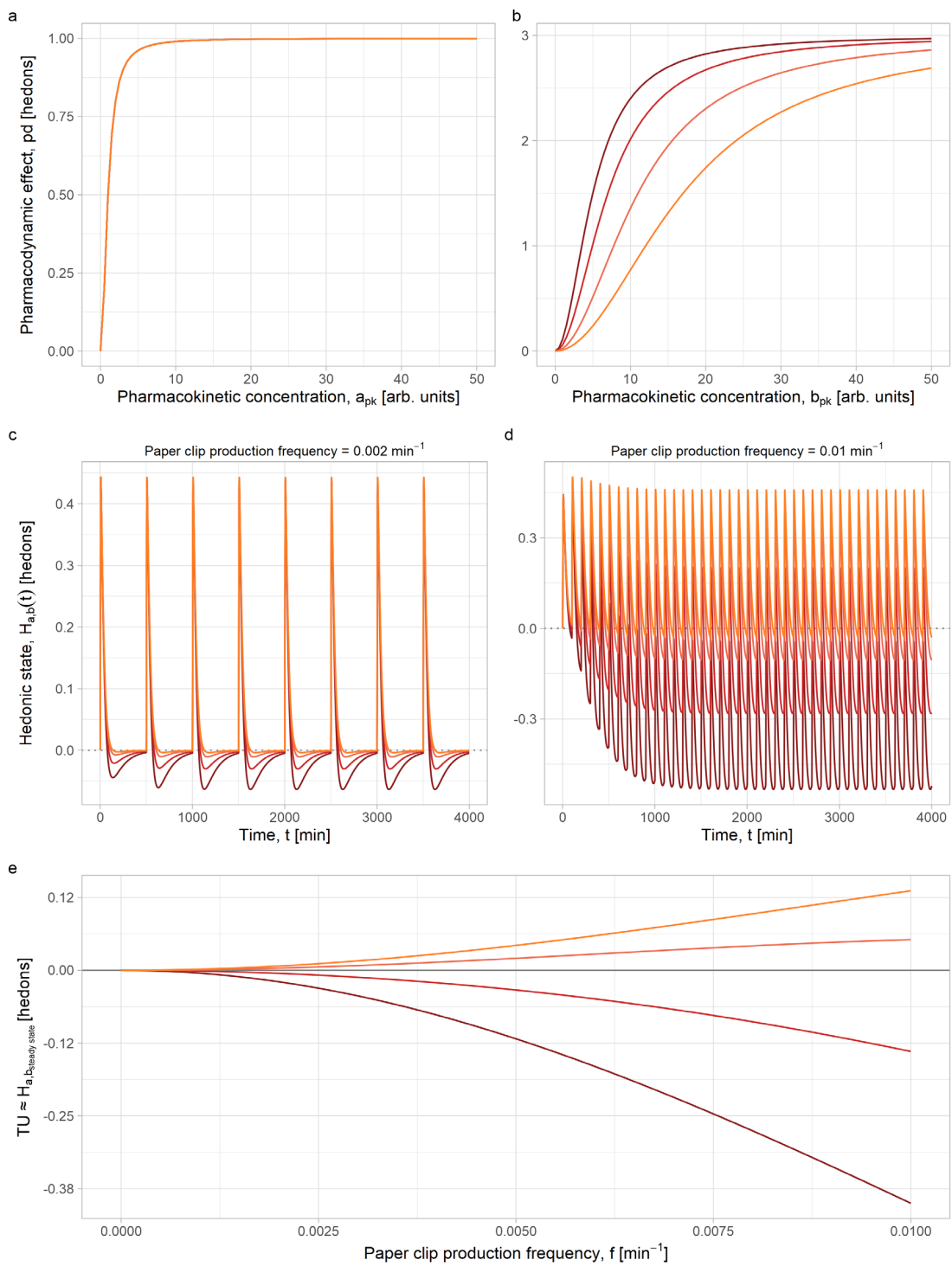

Figure 24. BFRA simulation produced with the following R command = bfra(EC50_b=c(5, 7, 11, 17), colorscheme=4)

### 11.1.7.8 Modifying gamma_b

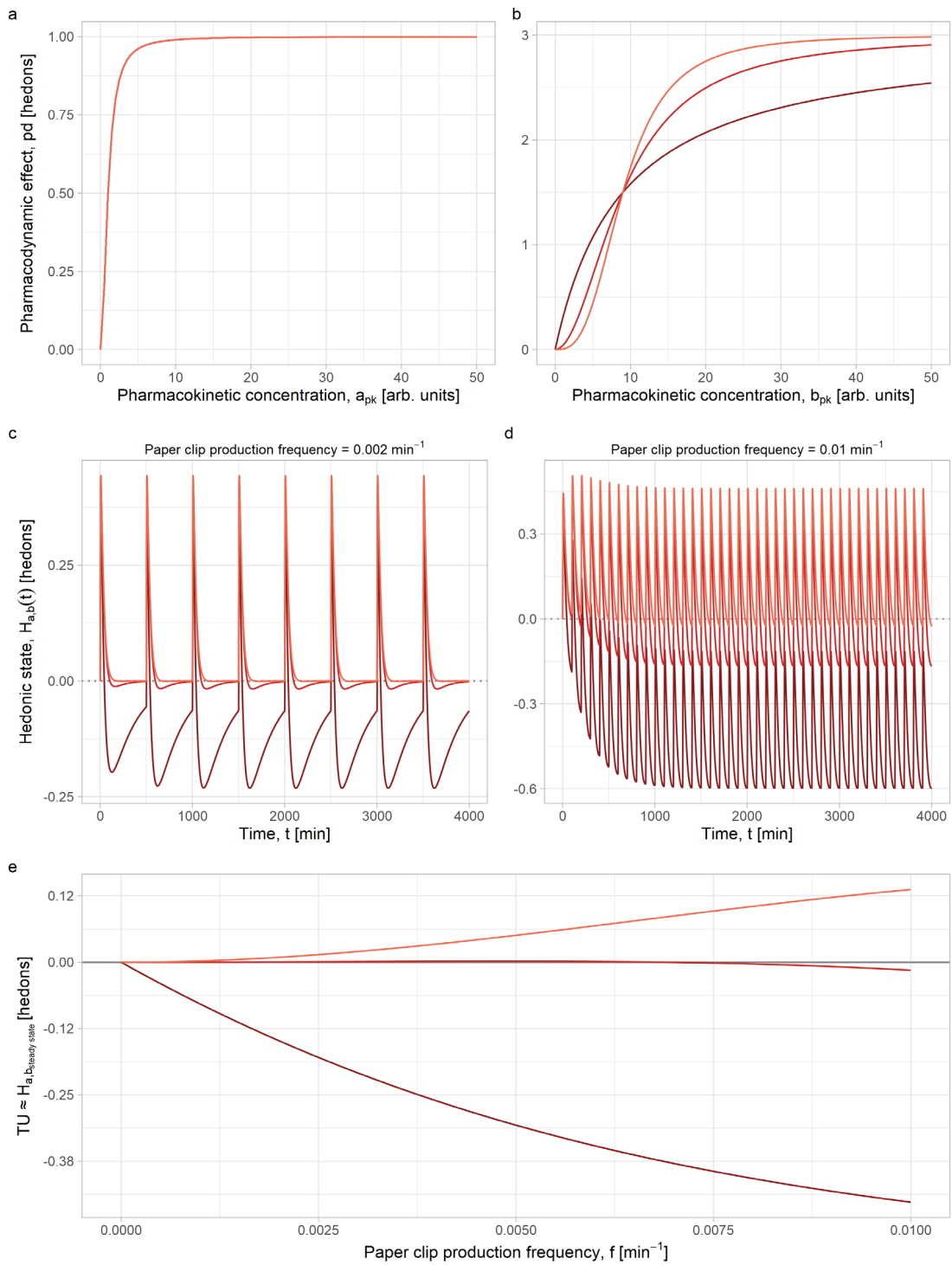

Figure 25. BFRA simulation produced with the following R command = bfra(gamma_b=c(1, 2, 3), colorscheme=4)

## 11.2 Examples of performing a BCRA (Behavioral Count Response Analysis)

The process for a Behavioral Count Response Analysis (BCRA) is identical to that for a BFRA, except in this case, the frequency of the behavioral doses is kept constant, and the count of behavioral doses is changed. This results in a count-response graph instead of a frequency-response graph.

We have included our code for performing a BCRA in the 'BCRA.R' file in the Supplementary Materials. The main function to use for BCRA simulations is bcra(). The default simulation is below, followed by simulations with varying parameters to modify the opponent processes. Default parameters are: k_Dose=1, k_apk=0.02, k_bpk=0.004, k_apd=1, k_bpd=1, k_H=1, E0_a=0, Emax_a=1, EC50_a=1, gamma_a=2, E0_b=0, Emax_b=3, EC50_b=9, and gamma_b=2.

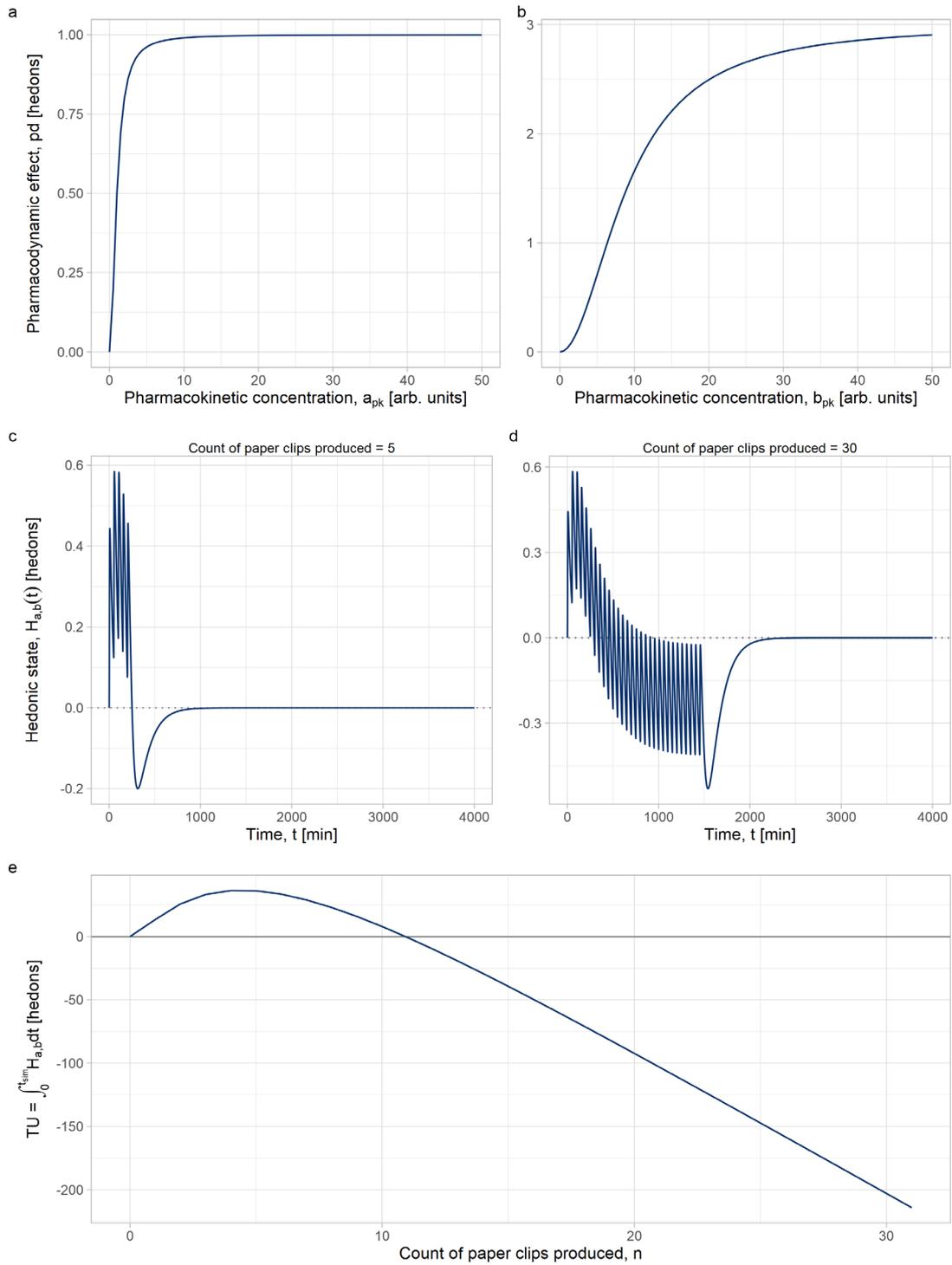

Figure 26. BCRA simulation produced with the following R command = bcra()

## 11.2.1 Pharmacokinetic perturbations

### 11.2.1.1 Modifying a-process pharmacokinetic decay constant

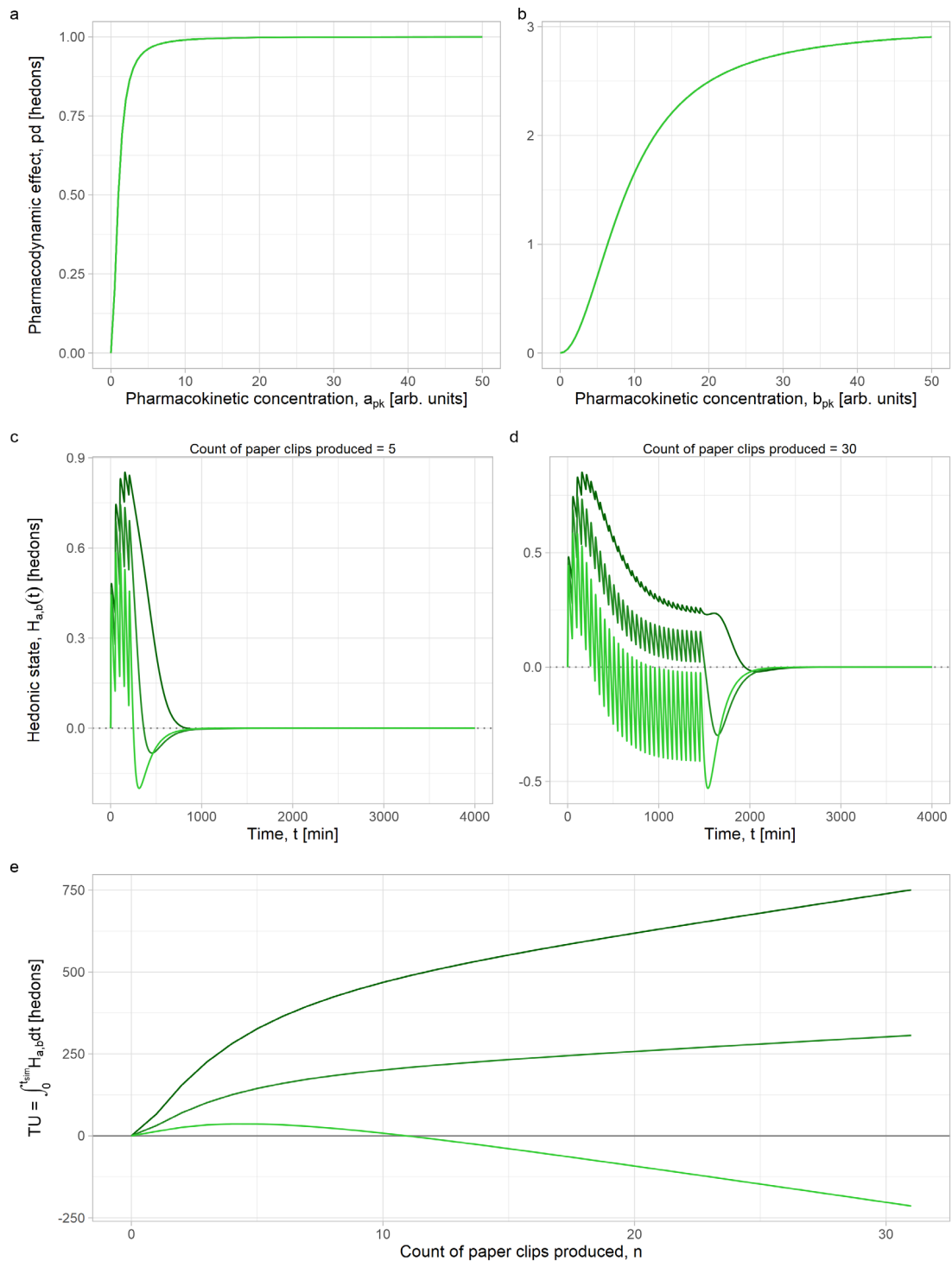

*Figure 27. BCRA simulation produced with the following R command = bcra(k_apk=c(0.005, 0.01, 0.02), colorscheme=2)*

*11.2.1.2 Modifying b-process pharmacokinetic decay constant*

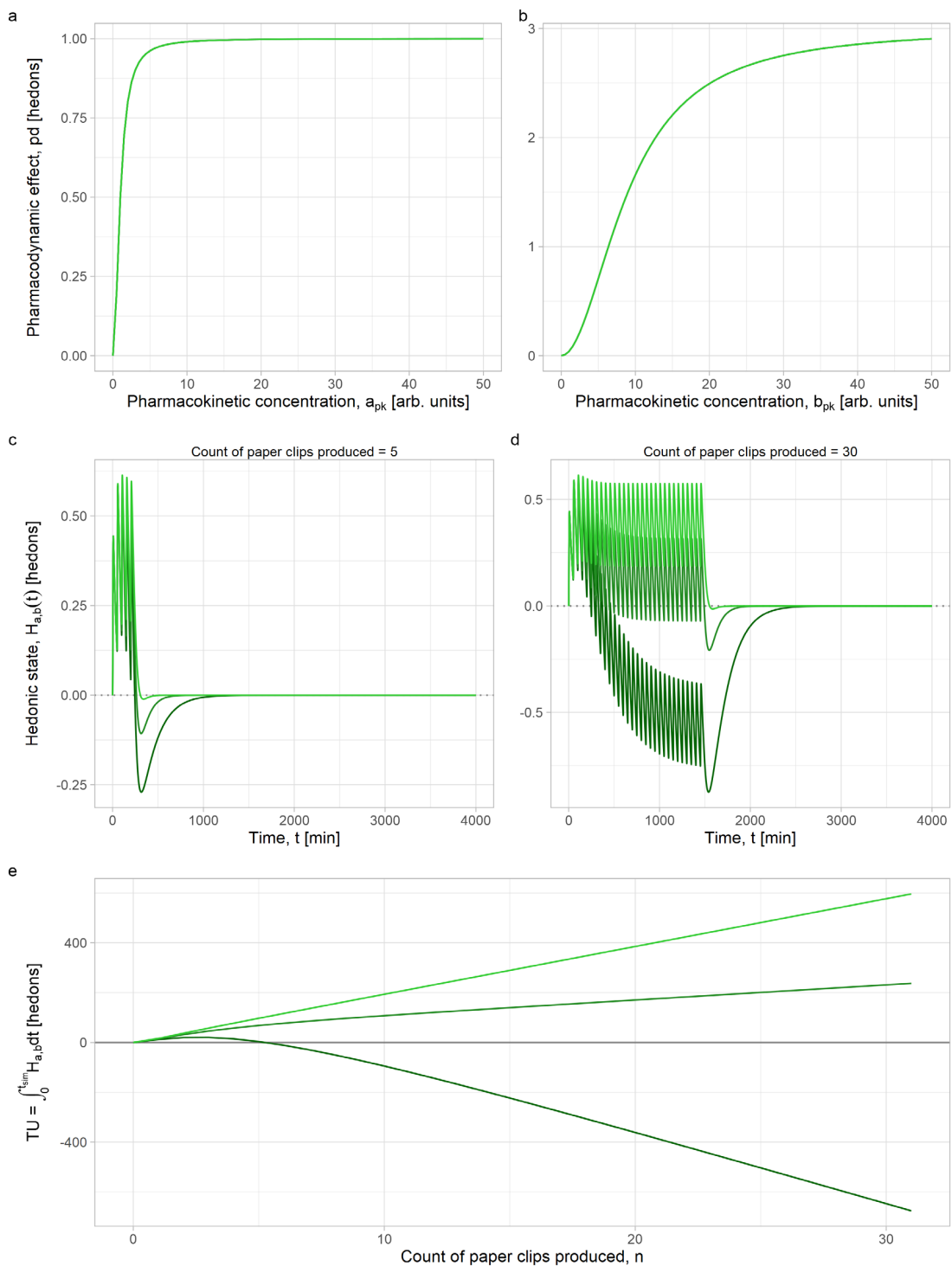

*Figure 28. BCRA simulation produced with the following R command = bcra(k_bpk=c(0.003, 0.006, 0.012), colorscheme=2). Note that the top two Hill equation graphs remain unchanged.*

## 11.2.2 Pharmacodynamic perturbations

### 11.2.2.1 Modifying a-process pharmacodynamic decay constant

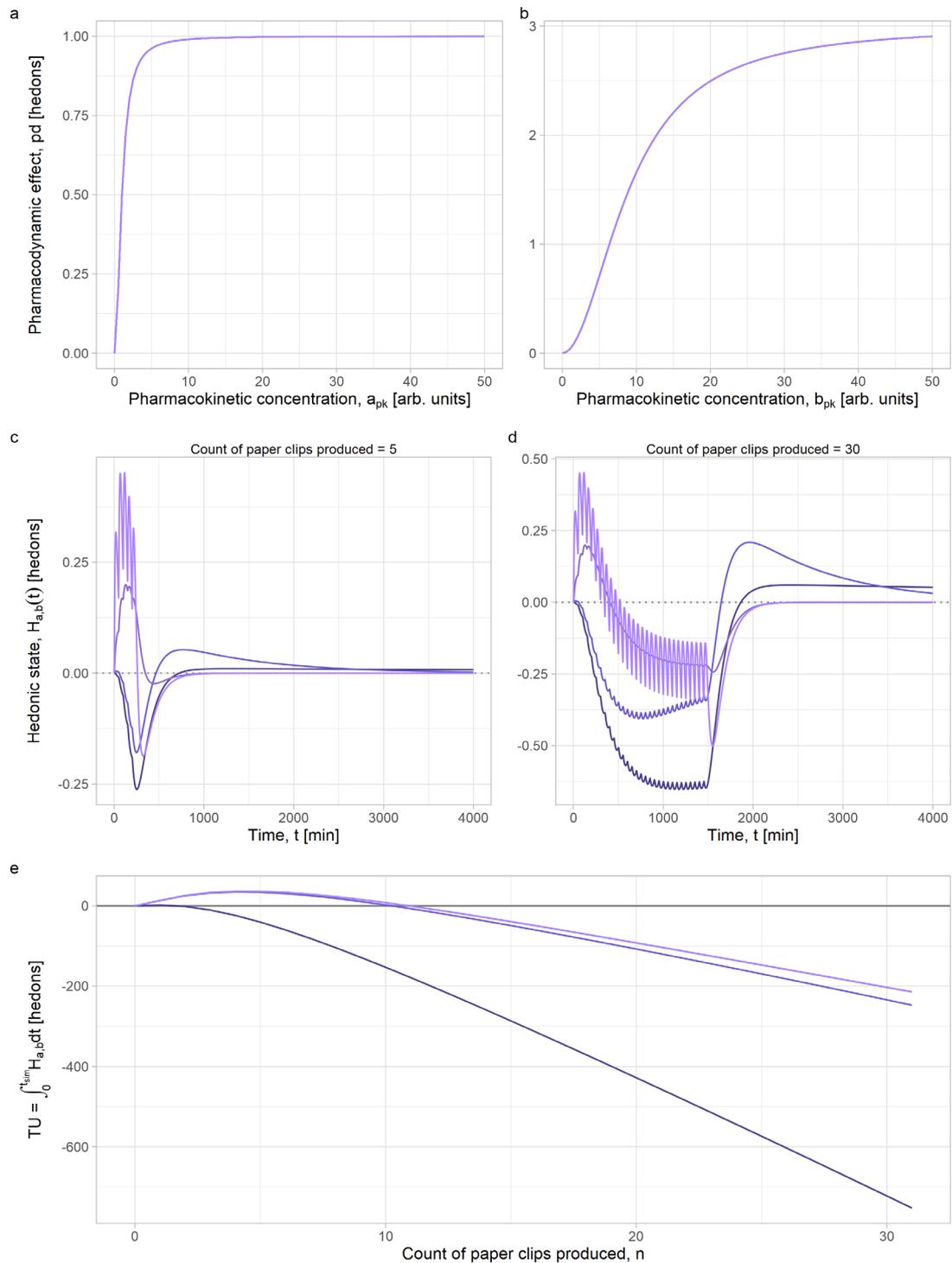

*Figure 29. BCRA simulation produced with the following R command = bcra(k_apd=c(0.0001, 0.001, 0.01, 0.1), colorscheme=3)*

## 11.2.2.2 Modifying b-process pharmacodynamic decay constant

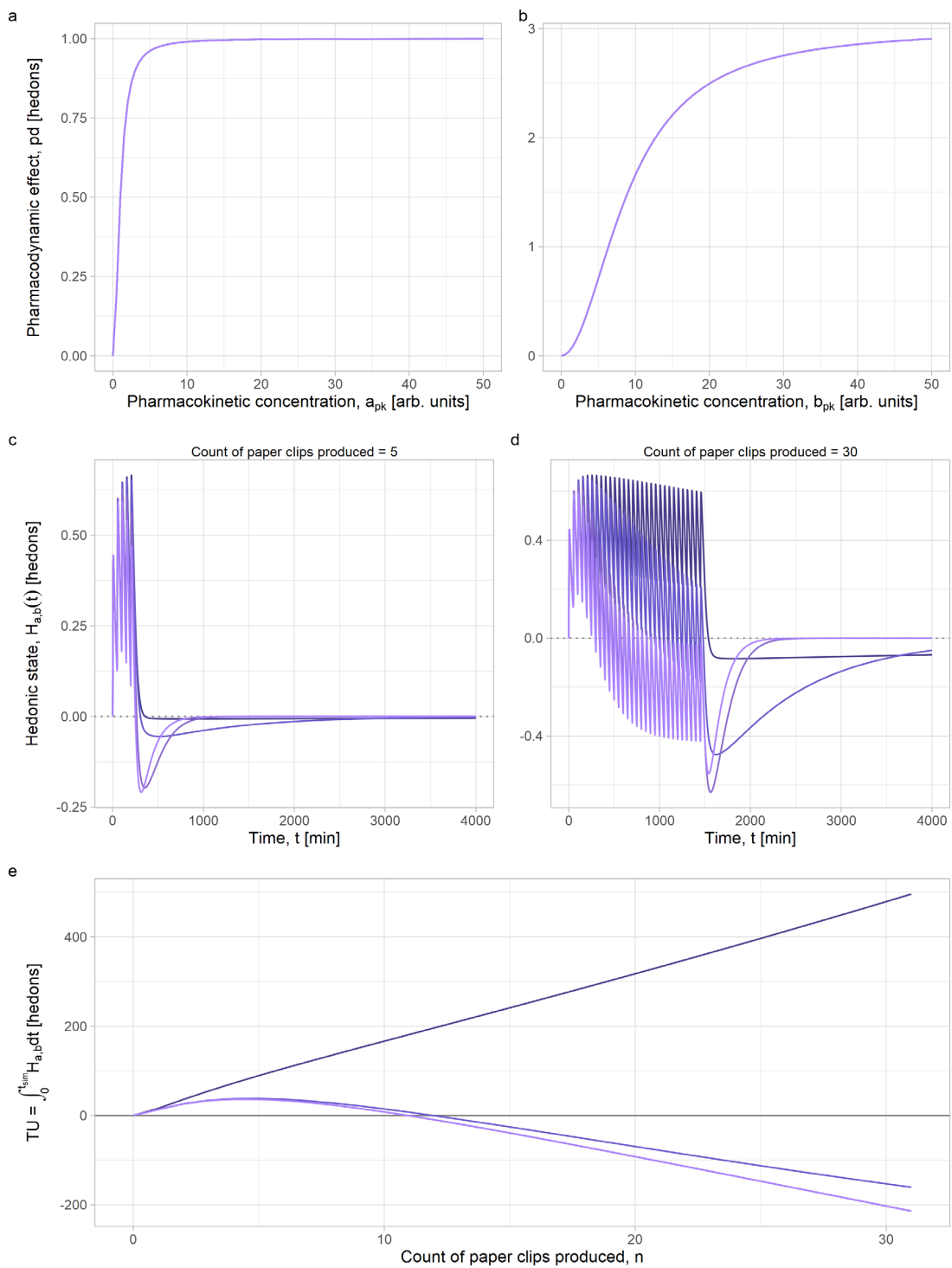

Figure 30. BCRA simulation produced with the following R command = bcra(k_bpd=c(0.0001, 0.001, 0.01, 0.1), colorscheme=3)

### 11.2.2.3 Modifying hedonic compartment (H) decay constant

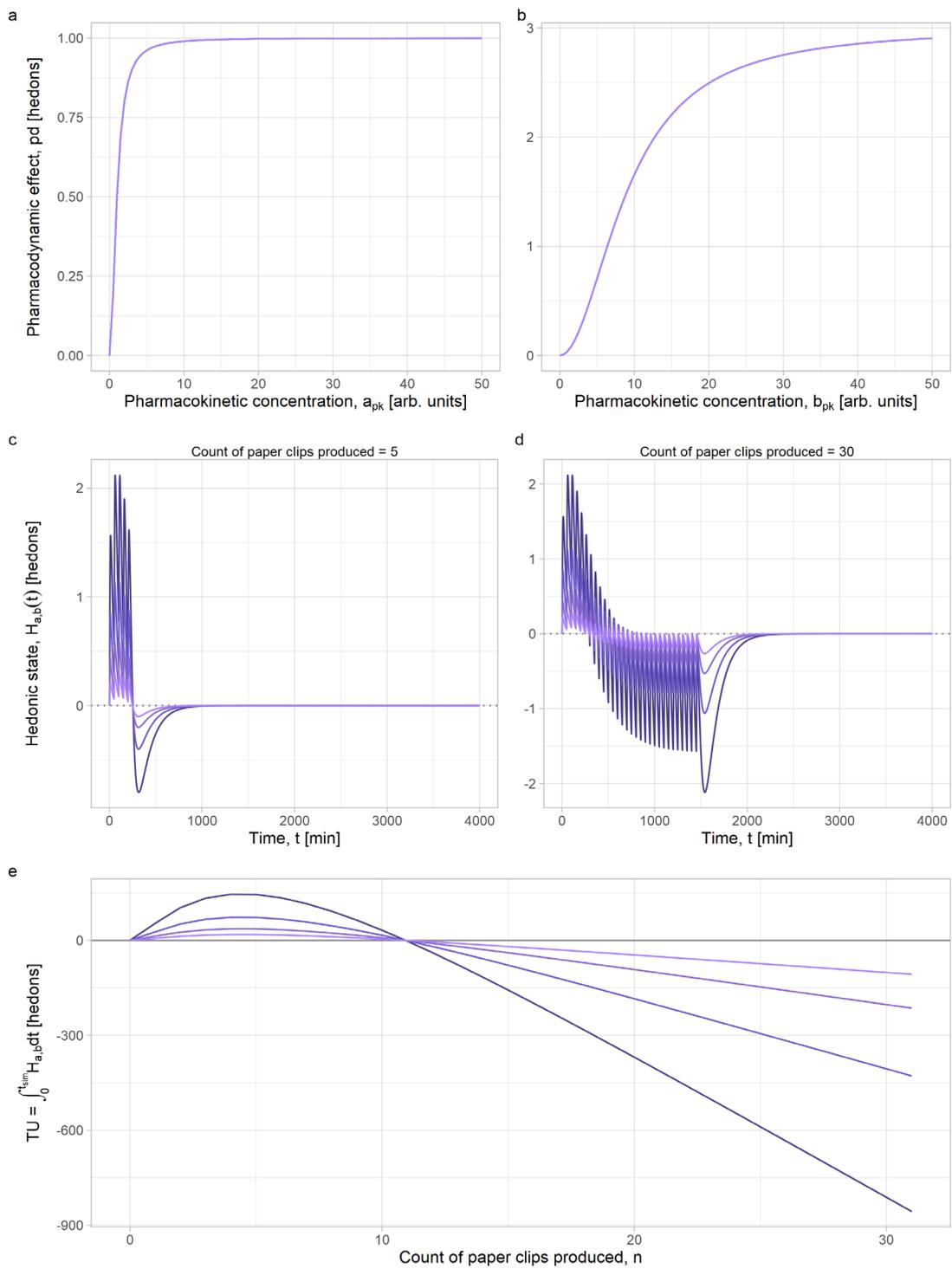

Figure 31. BCRA simulation produced with the following R command = bcra(k_H=c(0.25, 0.5, 1, 2), colorscheme=3)

## 11.2.3 Hill equation parameters

### 11.2.3.1 Modifying E0_a

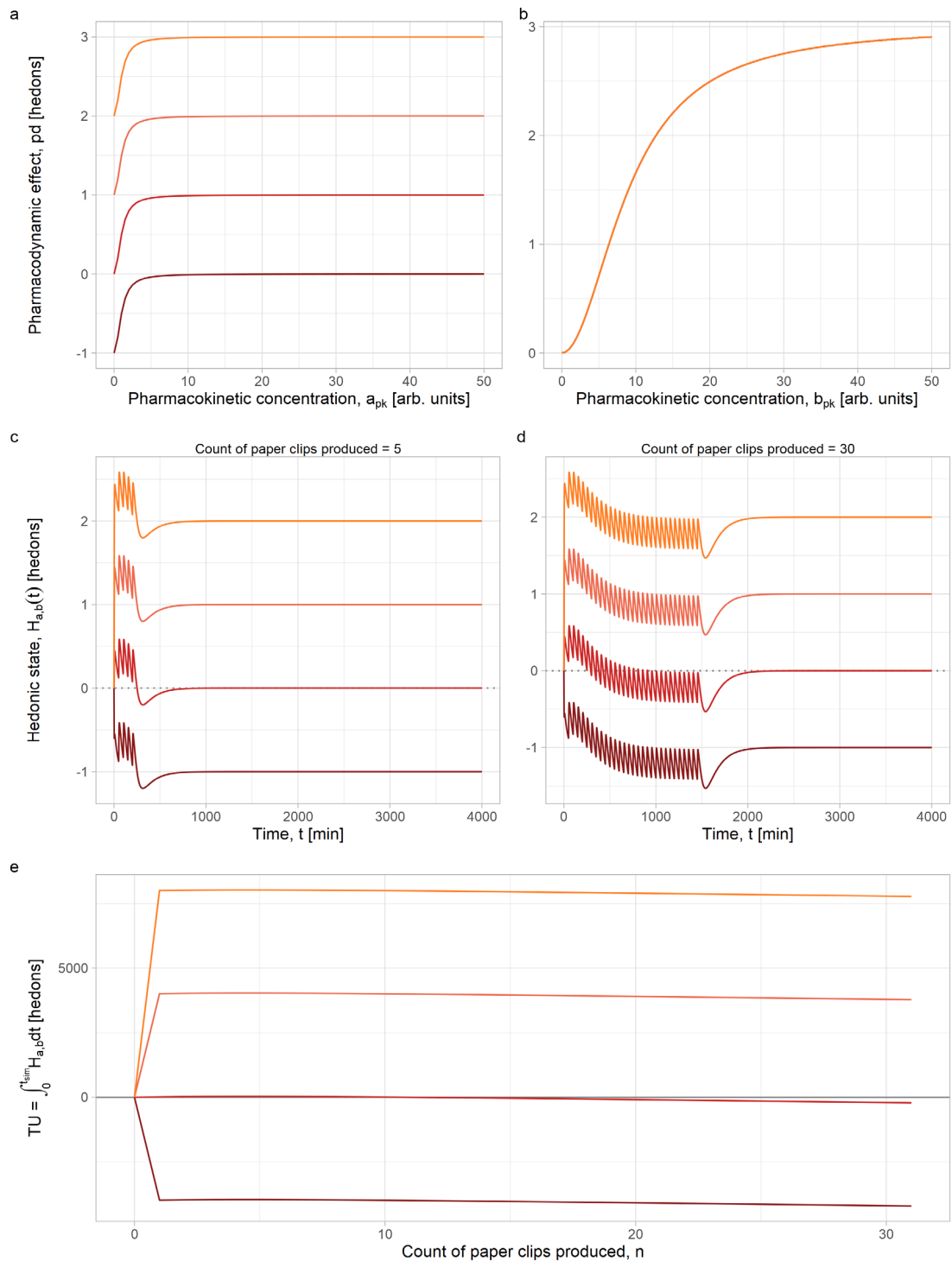

Figure 32. BCRA simulation produced with the following R command = bcra(E0_a=c(-1, 0, 1, 2), colorscheme=4)

### 11.2.3.2 Modifying Emax_a

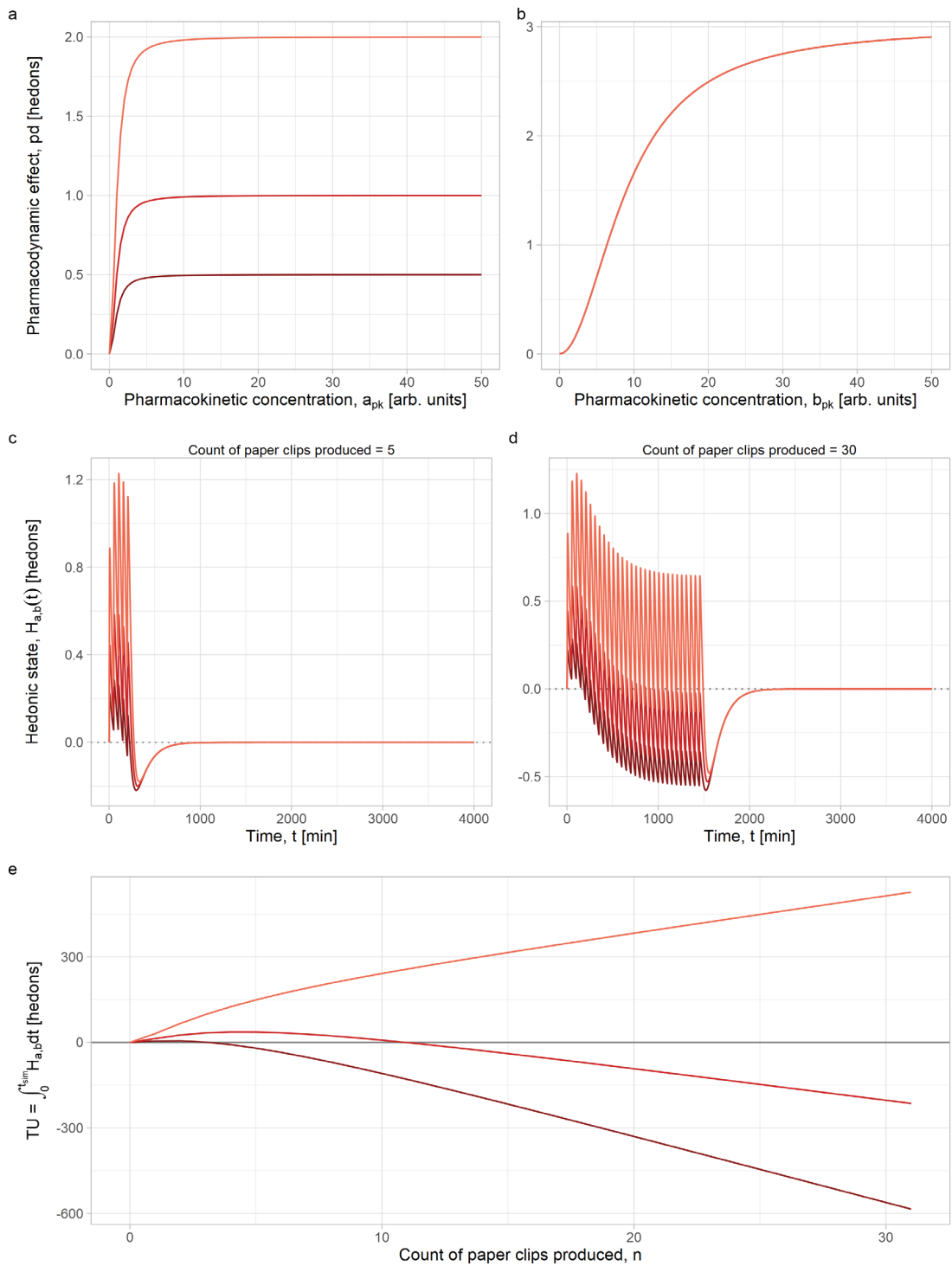

Figure 33. BCRA simulation produced with the following R command = bcra(Emax_a=c(0.5, 1, 2), colorscheme=4)

### 11.2.3.3 Modifying EC50_a

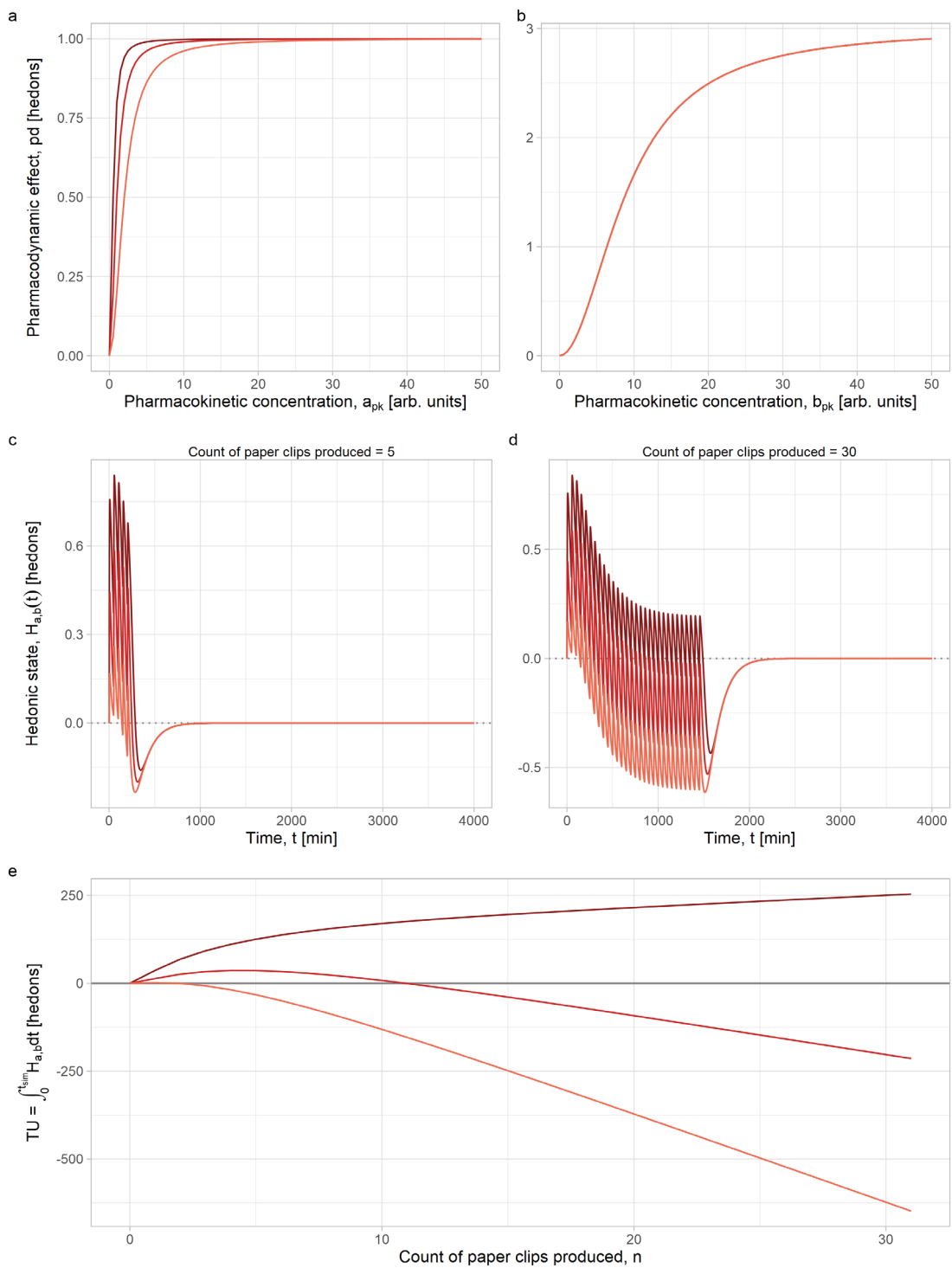

Figure 34. BCRA simulation produced with the following R command = bcra(EC50_a=c(0.5, 1, 2), colorscheme=4)

### 11.2.3.4 Modifying gamma_a

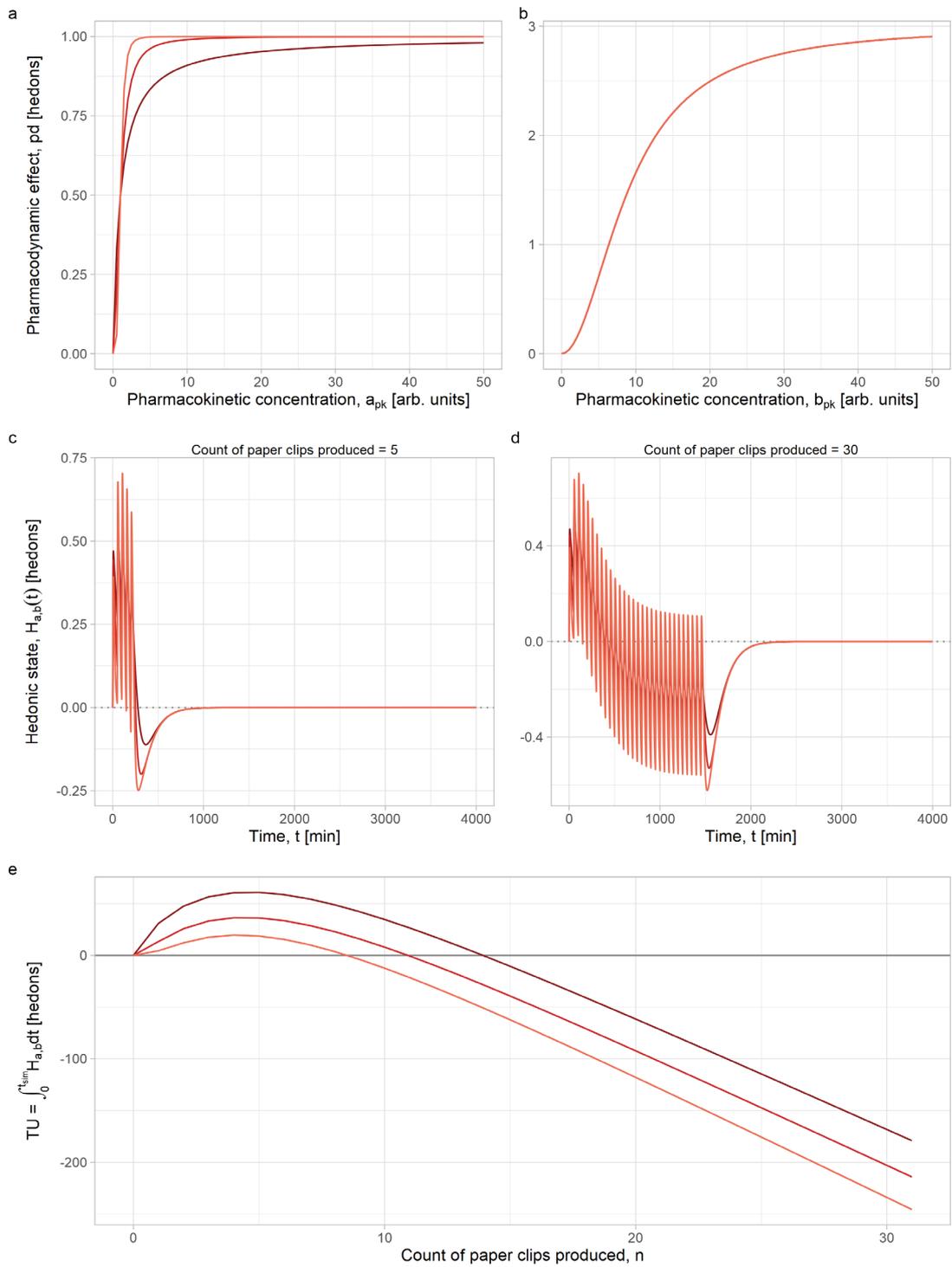

*Figure 35. BCRA simulation produced with the following R command = bcra(gamma_a=c(1, 2, 4), colorscheme=4)*

### 11.2.3.5 Modifying E0_b

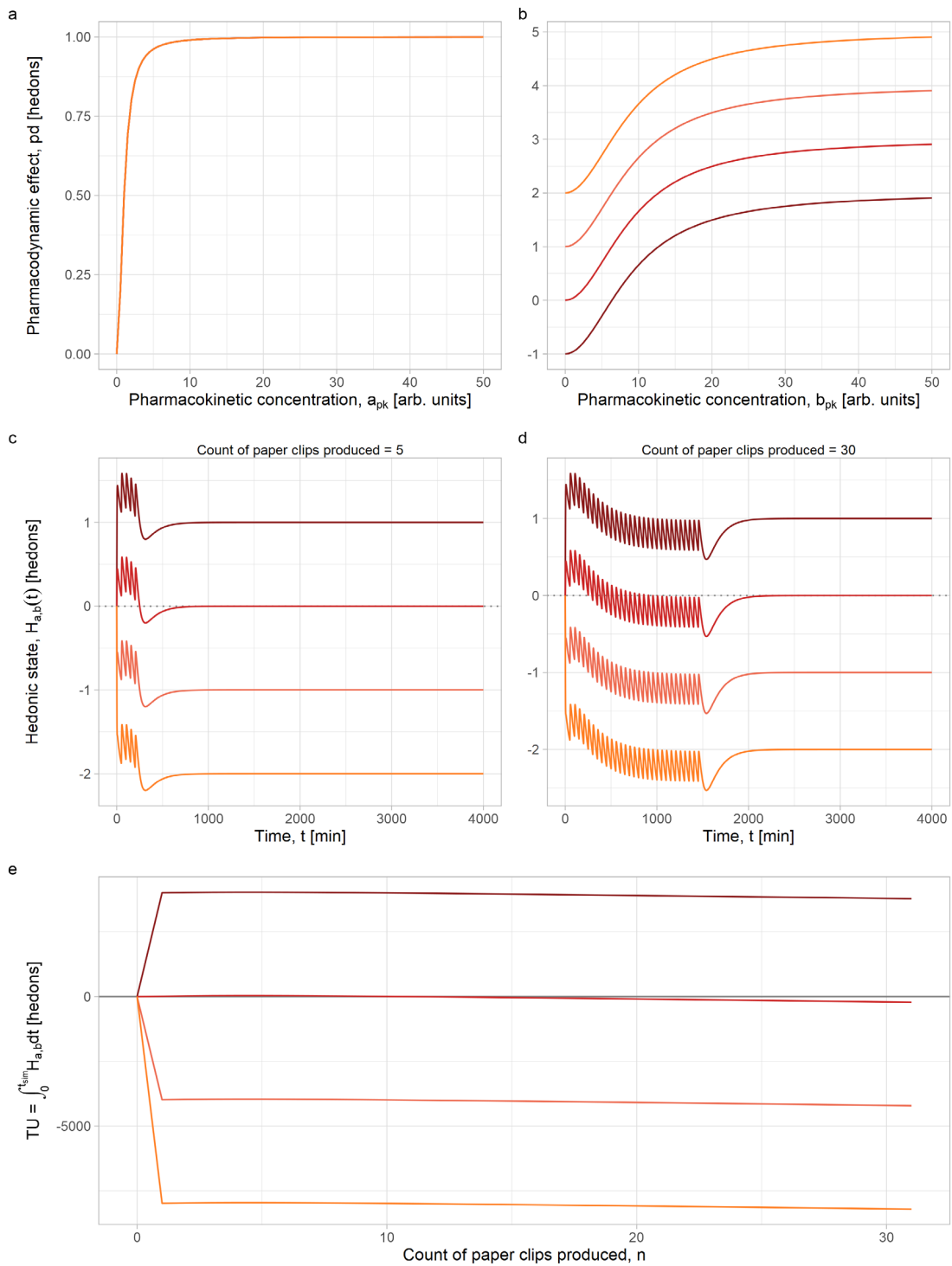

Figure 36. BCRA simulation produced with the following R command = bcra(E0_b=c(-1, 0, 1, 2), colorscheme=4)

### 11.2.3.6 Modifying Emax_b

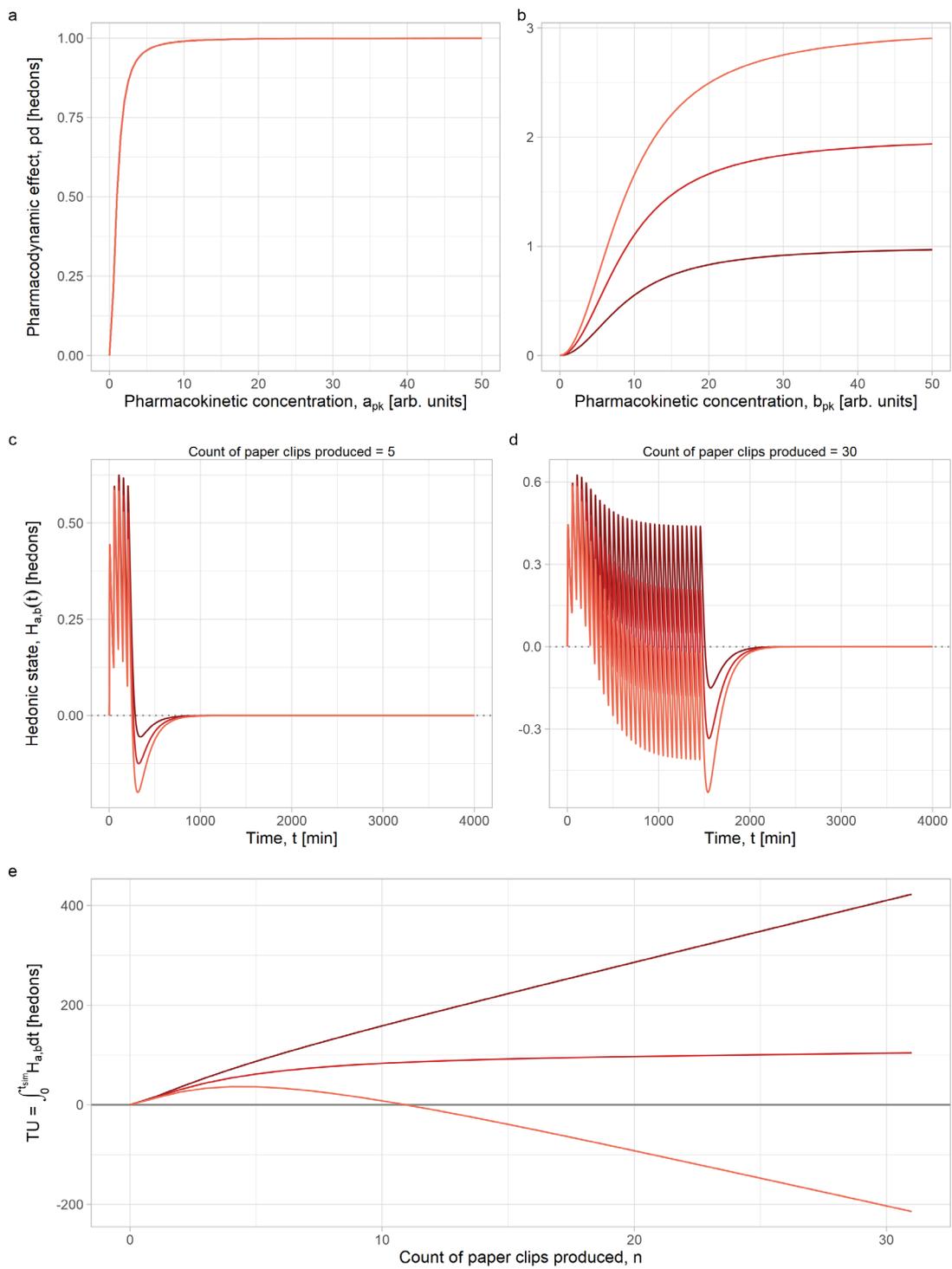

Figure 37. BCRA simulation produced with the following R command = bcra(Emax_b=c(1, 2, 3), colorscheme=4)

### 11.2.3.7 Modifying EC50_b

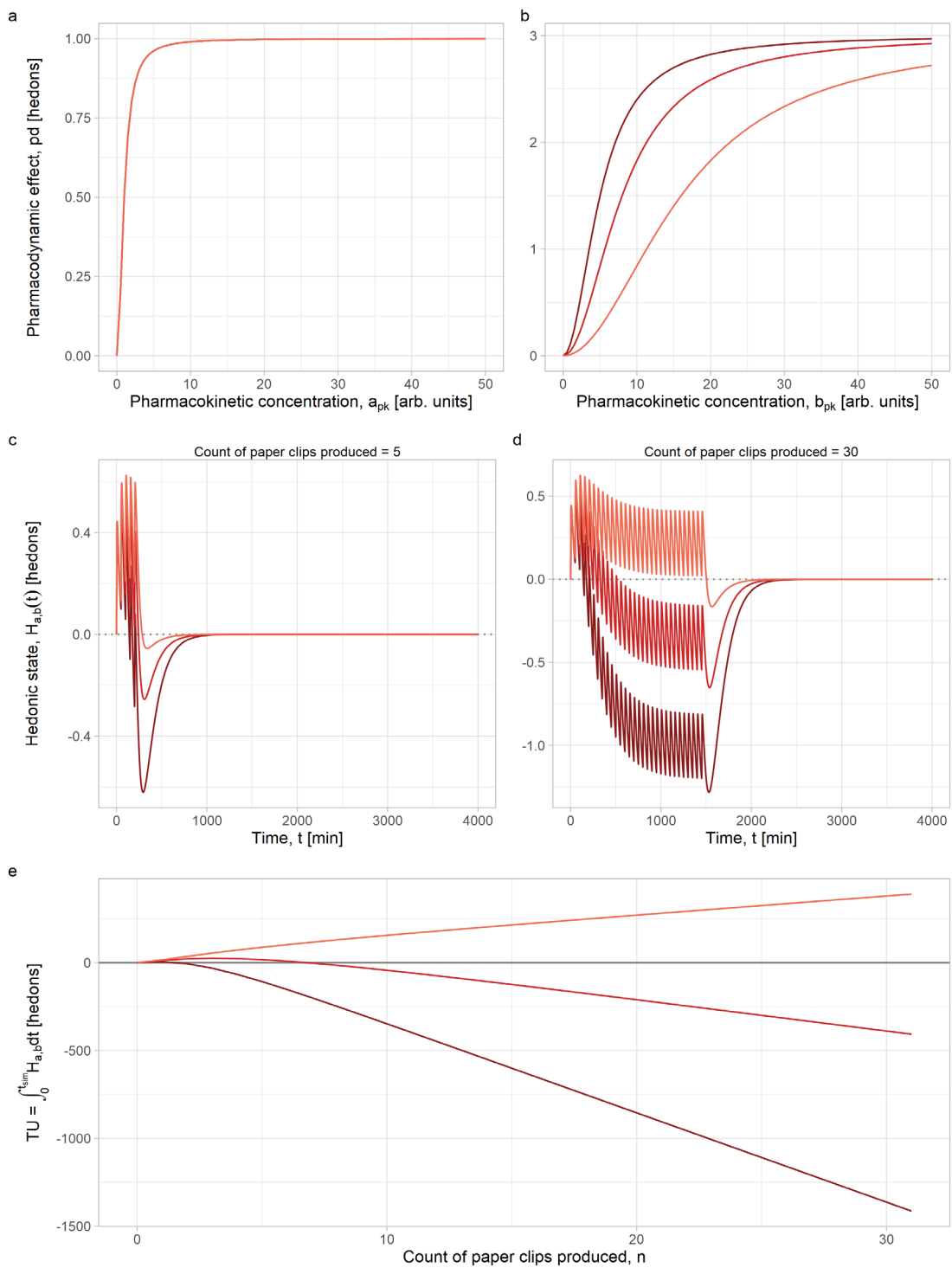

Figure 38. BCRA simulation produced with the following R command = bcra(EC50_b=c(5, 8, 16), colorscheme=4)

### 11.2.3.8 Modifying gamma_b

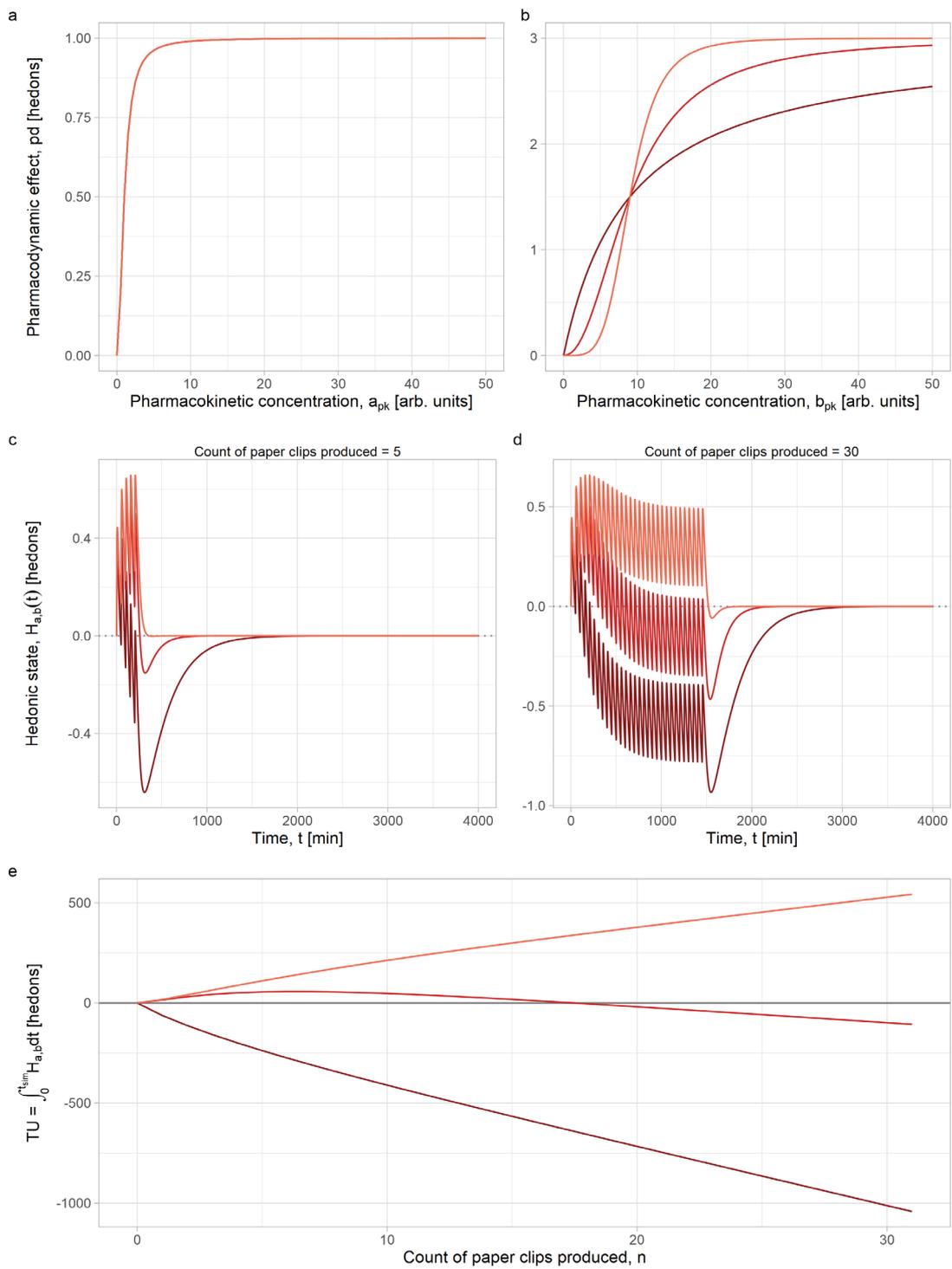

Figure 39. BCRA simulation produced with the following R command = bcra(gamma_b=c(1, 2.2, 4.6), colorscheme=4)

# 12 Appendix 3

## 12.1 BFRA.R

```
### This file contains R code for an example of a behavioral posology
simulation, performing a Behavioral Frequency Response Analysis on a PK/PD
model of a repeated digital behavior with opponent process dynamics. Examples
of how to run simulations using the function bfra() can be found in the
Supplementary Materials.

### Nathan Henry, November 2023

### Setup

library("tidyverse")

library("mrgsolve")

library("patchwork")

library("latex2exp")

# You may need to install some of these packages from GitHub, and may require
RTools. Refer to the documentation for remotes::install_github() and
https://cran.r-project.org/bin/windows/Rtools/, respectively.

## Package versions:

# tidyverse = 2.0.0

# mrgsolve = 1.0.9

# patchwork = 1.1.2

# latex2exp = 0.9.6

### -----

# Load in C++ model code

cpp_code <- "

  $PARAM // Parameters for simulation

    // Clearance rates for compartments - reset by opponentprocess_bfra()
function call

    k_Dose = 0,

    k_apk = 0,

    k_bpk = 0,
```

```
    k_apd = 0,
    k_bpd = 0,
    k_H = 0,

    // Pharmacodynamic constants - reset by opponentprocess_bfra() function call
    E0_a = 0,
    Emax_a = 0,
    EC50_a = 0,
    gamma_a = 0,
    E0_b = 0,
    Emax_b = 0,
    EC50_b = 0,
    gamma_b = 0,

    // Infusion duration
    infuse = 1

  $CMT // Model compartments
    Dose, // Hormonal concentration following Digital Behavior
    apk, // a-process pharmacokinetics
    apd, // a-process pharmacodynamics
    bpk, // b-process pharmacokinetics
    bpd, // b-process pharmacodynamics
    H // Overall hedonic outcomes

  $MAIN // Set additional relationships
    D_Dose = infuse; // Sets the infusion duration for digital behavior compartment

  $ODE // Ordinary Differential Equations
    dxdt_Dose = - k_Dose * Dose;
    dxdt_apk = k_Dose * Dose - k_apk * apk;
```

```
    dxdt_bpk = k_apk * apk - k_bpk * bpk;

    dxdt_apd = E0_a + (Emax_a * pow(apk, gamma_a)) / (pow(EC50_a, gamma_a)
+ pow(apk, gamma_a)) - k_apd * apd;

    dxdt_bpd = E0_b + (Emax_b * pow(bpk, gamma_b)) / (pow(EC50_b, gamma_b)
+ pow(bpk, gamma_b)) - k_bpd * bpd;

    dxdt_H = k_apd * apd - k_bpd * bpd - k_H * H;
  "

# Compile C++ code
mod <- mcode('Cppcode', cpp_code)

### color_scheme defines the color scheme for the BFRA graphs.

color_scheme <- function(scheme_num=1) {
  # Define a list of color schemes
  color_schemes <- list(
    scheme_1 = c('#08306B', '#2171B5', '#6BAED6', '#9ECAE1'),
    scheme_2 = c('darkgreen', 'forestgreen', 'limegreen', 'green'),
    scheme_3  =  c('slateblue4',  'slateblue3',  'mediumpurple3',
'mediumpurple1'),
    scheme_4 = c('firebrick4', 'firebrick3', 'coral2', 'chocolate1')
  )

  # Check if the scheme_num is valid
  if (scheme_num < 1 || scheme_num > length(color_schemes)) {
    stop("Invalid scheme number. Please choose a number from 1 to 4.")
  }

  # Return the selected color scheme
  return(color_schemes[[scheme_num]])
}
```

### opponentprocess_bfra() takes arguments for PK/PD models, creates an mrgsolve compartmental model, and plots the output.

```
opponentprocess_bfra <- function(
    ii=100000, # Dosing interval
    sim_length=4000, # Time length of PKPD simulation, in minutes
    addl=999999, # Number of additional doses to deliver - essentially infinite.
    plot_biophase=FALSE, # Whether to calculate the graphs for biophase or not.
    colorscheme=1, # Set color scheme for graphs

    # Set PK/PD constants for C++ code
    k_Dose=1,
    k_apk=0.02,
    k_bpk=0.004,
    k_apd=1,
    k_bpd=1,
    k_H=1,
    E0_a=0,
    Emax_a=1,
    EC50_a=1,
    gamma_a=2,
    E0_b=0,
    Emax_b=3,
    EC50_b=9,
    gamma_b=2,

    # Set infusion duration for drug input
    infuse=1,

    ## Set values for a-process, b-process, and double gamma plot datasets to return. These values represent the number of doses that fall within the allocated timeframe
```

```r
    plot_2=c(0.002, 0.01) # PKPD models for dose frequencies listed in plot_2 are plotted along with their associated Bode plot.
) {
  
  # Create data frame of parameters to pass to simulation.
  idataset=data.frame(
    k_Dose=k_Dose,
    k_apk=k_apk,
    k_bpk=k_bpk,
    k_apd=k_apd,
    k_bpd=k_bpd,
    k_H=k_H,
    E0_a=E0_a,
    Emax_a=Emax_a,
    EC50_a=EC50_a,
    gamma_a=gamma_a,
    E0_b=E0_b,
    Emax_b=Emax_b,
    EC50_b=EC50_b,
    gamma_b=gamma_b,
    infuse=infuse,
    sim_length=sim_length
  ) %>%
    rowid_to_column("ID") # Add column of IDs to start of data frame
  
  
  if (nrow(idataset) > 4) stop('Number of simulations must be 4 or less. Check idataset') # Stop if number of simulations > 4
  
  # Print out simulation parameters once
  if (plot_biophase) {
    cat('\nSimulation parameters =\n\n')
    print(idataset)
  }
```

```r
# Create a list of events
events <- ev(amt = 1, # Potency of dose
             rate = -2, # Signals that duration of infusion is modeled
             ii = ii, # Dosing interval
             ID = 1:nrow(idataset), # Add number of simulations being run
             addl = addl) # No. of additional doses to administer

# Run model
out <- mrgsim(mod, events, idataset, end=sim_length, maxsteps=50000)

# Calculate integral (AUC, area under curve) of H (hedonic) compartment
AUC_H <- out@data %>%
  group_by(ID) %>%
  summarise(AUC=sum(H))

# Calculate dose frequency
freq <- 1/ii
AUC_H$freq <- freq

# Print results
cat(paste('Integral of hedonic graph for simulation', AUC_H$ID, '=', AUC_H$AUC, '\n'))
cat(paste('Dose frequency =', freq, 'per min\n\n'))

# If rounded dose frequency value falls within plot_2 list, then return plot of H compartment
if (isTRUE(all.equal(freq, plot_2[[1]])) | isTRUE(all.equal(freq, plot_2[[2]]))) { # Use all.equal() to check equivalence of floating point numbers
  
  
  # Plot of H compartment over time
  cat('Saving plots for dose frequency above..................\n\n')
```

```
  plot_2_freq <- out@data %>%
    ggplot(aes(x=time, y=H, colour=factor(ID))) +
    geom_hline(yintercept=0, linetype='dotted', color='grey50') +
    geom_line() +
    scale_color_manual(values=color_scheme(colorscheme)) +
    ggtitle(bquote(paste('Paper clip production frequency = ', .(freq)) ~ min^-1)) +
    xlab('Time, t [min]') + {
      if            (isTRUE(all.equal(freq,           plot_2[[1]])))
ylab(bquote(paste('Hedonic  state,  H'[a*],'*b]*(t), ' [hedons]')))# Only
create y label if first plot
    } +
    theme_light() + {
      if (isTRUE(all.equal(freq, plot_2[[1]]))) { # Only create y label
if first plot
        theme(plot.title=element_text(size=9,                   hjust=0.5,
margin=margin(t=0, b=0)),
              legend.position='none')
      } else {
        theme(plot.title=element_text(size=9,                   hjust=0.5,
margin=margin(t=0, b=0)),
              legend.position='none',
              axis.title.y=element_blank())
      }
    }
} else {
  plot_2_freq <- NULL
}

## Create plots for biophase curves for PK -> PD conversion, using biophase
equations

if (plot_biophase) {
  # Set x axis length with dose_seq, then calculate biophase curves
```

```r
    pd_data <- tibble(dose_seq=seq(0, 50, 0.5))

    for (i in 1:nrow(idataset)) { # Calculate biophase curve for each set of parameters
        # Create column name for biophase curve based on ID number
        apd_colname <- paste0('apd', i); bpd_colname <- paste0('bpd', i)

        # Calculate biophase curves
        pd_data <- pd_data %>%
            mutate({{apd_colname}} := idataset$E0_a[i] + (idataset$Emax_a[i] * dose_seq ^ idataset$gamma_a[i]) / (idataset$EC50_a[i] ^ idataset$gamma_a[i] + dose_seq ^ idataset$gamma_a[i]),
                    {{bpd_colname}} := idataset$E0_b[i] + (idataset$Emax_b[i] * dose_seq ^ idataset$gamma_b[i]) / (idataset$EC50_b[i] ^ idataset$gamma_b[i] + dose_seq ^ idataset$gamma_b[i]))
    }

    # Plots for biophase curves
    apd_graph <- pd_data %>%
        pivot_longer(cols=starts_with('apd'),              names_to='ID', values_to='Values') %>%
        ggplot(aes(x=dose_seq, y=Values, colour=ID)) +
        geom_line() +
        scale_color_manual(values=color_scheme(colorscheme)) +
        theme_light() +
        theme(plot.title=element_text(size=9, hjust=0.5),
              legend.position='none') +
        ylab('Pharmacodynamic effect, pd [hedons]') +
        xlab(bquote(paste('Pharmacokinetic  concentration,  '*a[pk], ' [arb. units]')))
    bpd_graph <- pd_data %>%
        pivot_longer(cols=starts_with('bpd'),              names_to='ID', values_to='Values') %>%
        ggplot(aes(x=dose_seq, y=Values, colour=ID)) +
        geom_line() +
```

```
      scale_color_manual(values=color_scheme(colorscheme)) +
      theme_light() +
      theme(plot.title=element_text(size=9, hjust=0.5),
            legend.position='none',
            axis.title.y=element_blank()) +
      xlab(bquote(paste('Pharmacokinetic concentration, '*b[pk], ' [arb. units]')))
  }

  ## ----------------------------------------------------------------------------

  # Return necessary objects, including a list of parameters (idataset)
  ifelse(plot_biophase,
         return(list(AUC_H, freq, plot_2_freq, apd_graph, bpd_graph, idataset)),
         return(list(AUC_H, freq, plot_2_freq, NA, NA, idataset)))
}

### bfra() takes opponentprocess_bfra() and runs it across a range of dose frequencies, thus allowing us to plot the relationship between dose frequency and the integral of hedonic outcomes, and to determine whether this relationship is hormetic.
bfra <- function(
  # Pass on arguments to opponentprocess_bfra()
  ...,

  # Set color scheme for graphs
  colorscheme=1,

  # Set the resolution (seq_1) and upper limit (seq_2) of the x-axis values for the BFRA
  seq_1=0.0002,
  seq_2=0.01,
```

```r
  # Set y limit for hormesis graph (integer). If NA, ylim is automatically set
  gg_ylim=NA,
  
  # Plot BFRA simulated results, up to sim_length, the maximum simulation time
  include_simulated_results=FALSE,
  
  # Plot BFRA graph (either analytical or analytical + simulated)
  plot_bfra_graph=TRUE
) {
  
  # List of dose intervals to pass to opponentprocess_bfra()
  dose_interval <- c(0,  seq(seq_1, seq_2, seq_1)^-1)
  
  # Run loop to calculate Bode magnitude plot across range of frequencies
  H_list <- list() # Create list to house graphs of hedonic outcomes vs time
  for (i in 1:length(dose_interval)) {
    
    # If first dose interval, then set up bode_data data frame
    if (i == 1) {
      
      loop_list <- opponentprocess_bfra(ii=dose_interval[2],
                                        colorscheme=colorscheme,
                                        ...)
      
      # Create data frame to store wellbeing scores in, based on number of simulations performed
      bode_data <- tibble(
        'ID' = 1:nrow(loop_list[[1]]), # ID of each mrgsolve simulation
        'AUC' = rep(0, nrow(loop_list[[1]])), # AUC scores for H compartment graphs
```

```r
            'freq' = rep(0, nrow(loop_list[[1]])) # Dose frequency
        )
    } else if (i == 2) {
        # If second dose interval, then calculate biophase graphs. Otherwise just calculate loop_list to append to bode_data
        loop_list <- opponentprocess_bfra(ii=dose_interval[i],
                                         plot_biophase=TRUE,
                                         colorscheme=colorscheme,
                                         ...)
        apd_plot <- loop_list[[4]]; bpd_plot <- loop_list[[5]]
    } else {
        loop_list <- opponentprocess_bfra(ii=dose_interval[i],
                                         colorscheme=colorscheme,
                                         ...)
    }

    # Append AUC scores (hedonic outcomes) and dose frequencies, and store H compartment graphs
    bode_data <- bode_data %>%
      rbind(loop_list[[1]])
    H_list[[length(H_list) + 1]] <- loop_list[[3]]  # H compartment simulations
  }

  # Keep only distinct rows
  bode_data <- bode_data %>% distinct(.keep_all=TRUE)

  # Calculate the analytical solution to the BFRA, and add to the bode_data df. Use parameters passed to opponentprocess_bfra(), and scale the solution up by sim_length to ensure graphs can be compared easily
  params <- loop_list[[6]]

  H_axis       <-       params$sim_length       *       (params$E0_a       + params$Emax_a*((1*bode_data$freq/params$k_Dose)/params$k_apk)^params$gamma_a/(params$EC50_a^params$gamma_a+((1*bode_data$freq/params$k_Dose)/params$k_apk)^params$gamma_a)       -       (params$E0_b       +
```

```r
params$Emax_b*((1*bode_data$freq/params$k_Dose)/params$k_bpk)^params$gamma_b/(params$EC50_b^params$gamma_b+((1*bode_data$freq/params$k_Dose)/params$k_bpk)^params$gamma_b))) / params$k_H
  bode_data$H_analytical <- H_axis # Add analytical solutions to bode_data
  
  
  # Only plot BFRA graph if k_apd == 1, since the analytical solution is calculated based on pharmacokinetic decay constants, rather than pharmacodynamic. See equation to calculate H_axis above
  if (plot_bfra_graph & length(unique(params$k_apd)) < 2 & length(unique(params$k_bpd)) < 2) {
    if (unique(params$k_apd) == 1 & unique(params$k_bpd) == 1) {
      # Create Bode magnitude plot
      bode_graph <- bode_data %>%
        ggplot(aes(x=freq, color=factor(ID))) +
        geom_hline(yintercept=0, color='grey50') +
        geom_line(aes(y=H_analytical, linetype='Steady-state solution')) +
        {
          if (include_simulated_results) {
            geom_line(aes(y=AUC, linetype='Simulated solution'))
          }
        } +
        scale_color_manual(values=color_scheme(colorscheme)) + {
          if (!is.na(gg_ylim)) {
            coord_cartesian(ylim=c(gg_ylim, NA))
          }
        } + scale_linetype_manual(name = NULL, values = c("Steady-state solution" = "solid", "Simulated solution" = "dashed")) + {
          if (include_simulated_results) {
            scale_y_continuous(sec.axis = sec_axis(~ . / params$sim_length, name=bquote('TU \u2248' ~ H['a,b']['steady state']] ~ "[hedons]")))
          } else {
            scale_y_continuous(labels = scales::number_format(scale = 1/params$sim_length), name=bquote('TU \u2248' ~ H['a,b']['steady state']] ~ "[hedons]"))
          }
```

```r
          } +
          xlab(bquote('Paper clip production frequency, f' ~ '[' * min^-1 * ']')) +
          ylab(bquote('TU =' ~ integral(H['a,b']*dt, 0, t[sim]) ~ '[hedons]')) +
          guides(color='none') +
          theme_light() +
          theme(legend.position=ifelse(include_simulated_results,  "bottom", "none"), legend.box="horizontal", legend.justification="center")

        # Patch Hill equation, temporal, and Bode plots  together with patchwork package, and print
        tryCatch(bode_patch <- (apd_plot | bpd_plot) / (first(H_list) | last(H_list)) / bode_graph +
                   plot_annotation(tag_levels = 'a') &
                   theme(plot.tag = element_text(size = 11)),
                 error=function(e) {
                   warning(e)
                   stop("Error: bode_patch didn't patch together correctly. Double-check that the values in plot_2 are exact multiples of seq_1. So for example, you could pass the following arguments: 'seq_1=0.001, seq_2=0.1, plot_2=c(0.005, 0.1)'.")
                 }
        )
      } else {
        # Patch Hill equation and temporal plots together with patchwork package, and print
        tryCatch(bode_patch <- (apd_plot | bpd_plot) / (first(H_list) | last(H_list)) +
                   plot_annotation(tag_levels = 'a') &
                   theme(plot.tag = element_text(size = 11)),
                 error=function(e) {
                   stop("Error: bode_patch didn't patch together correctly. Double-check that the values in plot_2 are exact multiples of seq_1. So for example, you could pass the following arguments: 'seq_1=0.001, seq_2=0.1, plot_2=c(0.005, 0.1)'.")
                 }
```

```
      )
    }
  } else {
    # Patch Hill equation and temporal plots together with patchwork package, and print
    tryCatch(bode_patch <- (apd_plot | bpd_plot) / (first(H_list) | last(H_list)) +
               plot_annotation(tag_levels = 'a') &
               theme(plot.tag = element_text(size = 11)),
             error=function(e) {
               stop("Error: bode_patch didn't patch together correctly. Double-check that the values in plot_2 are exact multiples of seq_1. So for example, you could pass the following arguments: 'seq_1=0.001, seq_2=0.1, plot_2=c(0.005, 0.1)'.")
             }
    )
  }

  print(bode_patch)
}
```

## 12.2 BCRA.R

```
### This file contains R code for an example of a behavioral posology simulation, performing a Behavioral Count Response Analysis (BCRA) on a PK/PD model of a repeated digital behavior with opponent process dynamics. Examples of how to run simulations using the function bcra() can be found in the Supplementary Materials.

### Nathan Henry, November 2023

### Setup

library("tidyverse")

library("mrgsolve")

library("patchwork")

library("latex2exp")
```

# You may need to install some of these packages from GitHub, and may require RTools. Refer to the documentation for remotes::install_github() and https://cran.r-project.org/bin/windows/Rtools/, respectively.

## Package versions:

# tidyverse = 2.0.0

# mrgsolve = 1.0.9

# patchwork = 1.1.2

# latex2exp = 0.9.6

### -----

# Load in C++ model code

```
cpp_code <- "
  $PARAM // Parameters for simulation
    // Clearance rates for compartments - reset by opponentprocess_bcra() function call
    k_Dose = 0,
    k_apk = 0,
    k_bpk = 0,
    k_apd = 0,
    k_bpd = 0,
    k_H = 0,

    // Pharmacodynamic constants - reset by opponentprocess_bcra() function call
    E0_a = 0,
    Emax_a = 0,
    EC50_a = 0,
    gamma_a = 0,
    E0_b = 0,
    Emax_b = 0,
    EC50_b = 0,
    gamma_b = 0,
```

```
    // Infusion duration
    infuse = 1

  $CMT // Model compartments
    Dose, // Hormonal concentration following Digital Behavior
    apk, // a-process pharmacokinetics
    apd, // a-process pharmacodynamics
    bpk, // b-process pharmacokinetics
    bpd, // b-process pharmacodynamics
    H // Overall hedonic outcomes

  $MAIN // Set additional relationships
    D_Dose = infuse; // Sets the infusion duration for digital behavior compartment

  $ODE // Ordinary Differential Equations
    dxdt_Dose = - k_Dose * Dose;
    dxdt_apk = k_Dose * Dose - k_apk * apk;
    dxdt_bpk = k_apk * apk - k_bpk * bpk;
    dxdt_apd = E0_a + (Emax_a * pow(apk, gamma_a)) / (pow(EC50_a, gamma_a) + pow(apk, gamma_a)) - k_apd * apd;
    dxdt_bpd = E0_b + (Emax_b * pow(bpk, gamma_b)) / (pow(EC50_b, gamma_b) + pow(bpk, gamma_b)) - k_bpd * bpd;
    dxdt_H = k_apd * apd - k_bpd * bpd - k_H * H;
  "

# Compile C++ code
mod <- mcode('Cppcode', cpp_code)

### color_scheme defines the color scheme for the BCRA graphs.

color_scheme <- function(scheme_num=1) {
```

```r
  # Define a list of color schemes
  color_schemes <- list(
    scheme_1 = c('#08306B', '#2171B5', '#6BAED6', '#9ECAE1'),
    scheme_2 = c('darkgreen', 'forestgreen', 'limegreen', 'green'),
    scheme_3 = c('slateblue4', 'slateblue3', 'mediumpurple3', 'mediumpurple1'),
    scheme_4 = c('firebrick4', 'firebrick3', 'coral2', 'chocolate1')
  )

  # Check if the scheme_num is valid
  if (scheme_num < 1 || scheme_num > length(color_schemes)) {
    stop("Invalid scheme number. Please choose a number from 1 to 4.")
  }

  # Return the selected color scheme
  return(color_schemes[[scheme_num]])
}

### opponentprocess_bcra() takes arguments for PK/PD models, creates an mrgsolve compartmental model, and plots the output.

opponentprocess_bcra <- function(
    ii=50, # Dosing interval
    sim_length=4000, # Time length of PKPD simulation, in minutes
    addl=0, # Number of additional doses to deliver - essentially infinite.
    plot_biophase=FALSE, # Whether to calculate the graphs for biophase or not.
    colorscheme=1, # Set color scheme for graphs

    # Set PK/PD constants for C++ code
    k_Dose=1,
    k_apk=0.02,
    k_bpk=0.004,
    k_apd=1,
```

```
    k_bpd=1,

    k_H=1,

    E0_a=0,

    Emax_a=1,

    EC50_a=1,

    gamma_a=2,

    E0_b=0,

    Emax_b=3,

    EC50_b=9,

    gamma_b=2,

    # Set infusion duration for drug input

    infuse=1,

    ## Set values for a-process, b-process, and double gamma plot datasets
to return. These values represent the number of doses that fall within the
allocated timeframe

    plot_2=c(4, 29) # PKPD models for number of repeated doses listed in
plot_2 are plotted along with their associated Bode plot.

) {

  # Create data frame of parameters to pass to simulation.

  idataset=data.frame(

    k_Dose=k_Dose,

    k_apk=k_apk,

    k_bpk=k_bpk,

    k_apd=k_apd,

    k_bpd=k_bpd,

    k_H=k_H,

    E0_a=E0_a,

    Emax_a=Emax_a,

    EC50_a=EC50_a,

    gamma_a=gamma_a,
```

```r
      E0_b=E0_b,
      Emax_b=Emax_b,
      EC50_b=EC50_b,
      gamma_b=gamma_b,
      infuse=infuse,
      sim_length=sim_length
    ) %>%
      rowid_to_column("ID") # Add column of IDs to start of data frame

  if (nrow(idataset) > 4) stop('Number of simulations must be 4 or less. Check idataset') # Stop if number of simulations > 4

  # Print out simulation parameters once
  if (plot_biophase) {
    cat('\nSimulation parameters =\n\n')
    print(idataset)
  }

  # Create a list of events
  events <- ev(amt = 1, # Potency of dose
               rate = -2, # Signals that duration of infusion is modeled
               ii = ii, # Dosing interval
               ID = 1:nrow(idataset), # Add number of simulations being run
               addl = addl) # No. of additional doses to administer, noting that mrgsolve always

  # Run model
  out <- mrgsim(mod, events, idataset, end=sim_length, maxsteps=50000)

  # Calculate integral (AUC, area under curve) of H (hedonic) compartment
  AUC_H <- out@data %>%
    group_by(ID) %>%
    summarise(AUC=sum(H))
```

```r
  # Make a note of the number of total doses run
  AUC_H$n <- addl + 1

  # Print results
  cat(paste('Integral of hedonic graph for simulation', AUC_H$ID, '=', AUC_H$AUC, '\n'))
  cat(paste('No. of additional doses =', addl, '\n\n'))

  # If the number of additional doses falls within plot_2 list, then return plot of H compartment
  if (addl %in% plot_2) {

    # Plot of H compartment over time
    cat('Saving plots for dose repetitions above.................\n\n')
    plot_2_addl <- out@data %>%
      ggplot(aes(x=time, y=H, colour=factor(ID))) +
      geom_hline(yintercept=0, linetype='dotted', color='grey50') +
      geom_line() +
      scale_color_manual(values=color_scheme(colorscheme)) +
      ggtitle(paste('Count of paper clips produced =', addl + 1)) +
      xlab('Time, t [min]') + {
        if           (isTRUE(all.equal(addl,            plot_2[[1]])))
ylab(bquote(paste('Hedonic  state,  H'[a*','*b]*(t), ' [hedons]')))# Only create y label if first plot
      } +
      theme_light() + {
        if (isTRUE(all.equal(addl, plot_2[[1]]))) { # Only create y label if first plot
          theme(plot.title=element_text(size=9,                    hjust=0.5, margin=margin(t=0, b=0)),
                legend.position='none')
        } else {
```

```r
              theme(plot.title=element_text(size=9,                    hjust=0.5, margin=margin(t=0, b=0)),
                    legend.position='none',
                    axis.title.y=element_blank())
        }
      }
  } else {
    plot_2_addl <- NULL
  }

  ## Create plots for biophase curves for PK -> PD conversion, using biophase equations

  if (plot_biophase) {
    # Set x axis length with dose_seq, then calculate biophase curves
    pd_data <- tibble(dose_seq=seq(0, 50, 0.5))

    for (i in 1:nrow(idataset)) { # Calculate biophase curve for each set of parameters
      # Create column name for biophase curve based on ID number
      apd_colname <- paste0('apd', i); bpd_colname <- paste0('bpd', i)

      # Calculate biophase curves
      pd_data <- pd_data %>%
        mutate({{apd_colname}} := idataset$E0_a[i] + (idataset$Emax_a[i] * dose_seq ^ idataset$gamma_a[i]) / (idataset$EC50_a[i] ^ idataset$gamma_a[i] + dose_seq ^ idataset$gamma_a[i]),
               {{bpd_colname}} := idataset$E0_b[i] + (idataset$Emax_b[i] * dose_seq ^ idataset$gamma_b[i]) / (idataset$EC50_b[i] ^ idataset$gamma_b[i] + dose_seq ^ idataset$gamma_b[i]))
    }

    # Plots for biophase curves
    apd_graph <- pd_data %>%
```

```r
    pivot_longer(cols=starts_with('apd'),              names_to='ID', values_to='Values') %>%
    ggplot(aes(x=dose_seq, y=Values, colour=ID)) +
    geom_line() +
    scale_color_manual(values=color_scheme(colorscheme)) +
    theme_light() +
    theme(plot.title=element_text(size=9, hjust=0.5),
          legend.position='none') +
    ylab('Pharmacodynamic effect, pd [hedons]') +
    xlab(bquote(paste('Pharmacokinetic concentration, '*a[pk], ' [arb. units]')))
  bpd_graph <- pd_data %>%
    pivot_longer(cols=starts_with('bpd'),              names_to='ID', values_to='Values') %>%
    ggplot(aes(x=dose_seq, y=Values, colour=ID)) +
    geom_line() +
    scale_color_manual(values=color_scheme(colorscheme)) +
    theme_light() +
    theme(plot.title=element_text(size=9, hjust=0.5),
          legend.position='none',
          axis.title.y=element_blank()) +
    xlab(bquote(paste('Pharmacokinetic concentration, '*b[pk], ' [arb. units]')))
  }

  ## -----------------------------------------------------------------------

  # Return necessary objects, including a list of parameters (idataset)
  ifelse(plot_biophase,
         return(list(AUC_H, addl, plot_2_addl, apd_graph, bpd_graph, idataset)),
         return(list(AUC_H, addl, plot_2_addl, NA, NA, idataset)))
}
```

### bcra() takes opponentprocess_bcra() and runs it across a range of dose repetitions, thus allowing us to plot the relationship between the number of repeated doses and the integral of hedonic outcomes, and to determine whether this relationship is hormetic.

bcra <- function(

  # Pass on arguments to opponentprocess_bcra()

  ...,

  # Set color scheme for graphs

  colorscheme=1,

  # Set the resolution (seq_1) and upper limit (seq_2) of the x-axis values for the BCRA

  seq_1=1,

  seq_2=30,

  # Set y limit for hormesis graph (integer). If NA, ylim is automatically set

  gg_ylim=NA,

  # Plot BCRA graph (either analytical or analytical + simulated)

  plot_bcra_graph=TRUE

) {

  # List of dose repetitions to pass to opponentprocess_bcra()

  dose_repetitions <- c(0,  seq(seq_1, seq_2, seq_1))

  # Run loop to calculate Bode magnitude plot across range of dose repetitions

  H_list <- list() # Create list to house graphs of hedonic outcomes vs time

  for (i in 1:length(dose_repetitions)) {

    # If first dose repetition value, then set up bode_data data frame

```r
    if (i == 1) {
      
      loop_list <- opponentprocess_bcra(ii=dose_repetitions[2],
                                        colorscheme=colorscheme,
                                        ...)
      
      # Create data frame to store wellbeing scores in, based on number of simulations performed
      bode_data <- tibble(
        'ID' = 1:nrow(loop_list[[1]]), # ID of each mrgsolve simulation
        'AUC' = rep(0, nrow(loop_list[[1]])), # AUC scores for H compartment graphs
        'n' = rep(0, nrow(loop_list[[1]])) # Dose repetitions
      )
    } else if (i == 2) {
      # If second dose repetition value, then calculate biophase graphs. Otherwise just calculate loop_list to append to bode_data
      loop_list <- opponentprocess_bcra(addl=dose_repetitions[i],
                                        plot_biophase=TRUE,
                                        colorscheme=colorscheme,
                                        ...)
      apd_plot <- loop_list[[4]]; bpd_plot <- loop_list[[5]]
    } else {
      loop_list <- opponentprocess_bcra(addl=dose_repetitions[i],
                                        colorscheme=colorscheme,
                                        ...)
    }
    
    # Append AUC scores (hedonic outcomes) and dose repetitions, and store H compartment graphs
    bode_data <- bode_data %>%
      rbind(loop_list[[1]])
    H_list[[length(H_list) + 1]] <- loop_list[[3]] # H compartment simulations
```

```r
  }

  # Keep only distinct rows
  bode_data <- bode_data %>% distinct(.keep_all=TRUE)

  # Only plot BCRA graph if k_apd == 1, since the analytical solution is calculated based on pharmacokinetic decay constants, rather than pharmacodynamic. See equation to calculate H_axis above
  if (plot_bcra_graph) {
    # Create Bode magnitude plot
    bode_graph <- bode_data %>%
      ggplot(aes(x=n, y=AUC, color=factor(ID))) +
      geom_hline(yintercept=0, color='grey50') +
      geom_line() +
      scale_color_manual(values=color_scheme(colorscheme)) + {
        if (!is.na(gg_ylim)) {
          coord_cartesian(ylim=c(gg_ylim, NA))
        }
      } +
      xlab('Count of paper clips produced, n') +
      ylab(bquote('TU =' ~ integral(H['a,b']*dt, 0, t[sim]) ~ '[hedons]')) +
      guides(color='none') +
      theme_light() +
      theme(legend.position='none')

    # Patch Hill equation, temporal, and Bode plots  together with patchwork package, and print
    tryCatch(bode_patch <-  (apd_plot | bpd_plot) / (first(H_list) | last(H_list)) / bode_graph +
               plot_annotation(tag_levels = 'a') &
               theme(plot.tag = element_text(size = 11)),
             error=function(e) {
               warning(e)
```

```
              stop("Error: bode_patch didn't patch together correctly. Double-check that the values in plot_2 are exact multiples of seq_1. So for example, you could pass the following arguments: 'seq_1=1, seq_2=50, plot_2=c(5, 20)'.")
            }
      )
  } else {
    # Patch Hill equation and temporal plots together with patchwork package, and print
    tryCatch(bode_patch <- (apd_plot | bpd_plot) / (first(H_list) | last(H_list)) +
               plot_annotation(tag_levels = 'a') &
               theme(plot.tag = element_text(size = 11)),
             error=function(e) {
               stop("Error: bode_patch didn't patch together correctly. Double-check that the values in plot_2 are exact multiples of seq_1. So for example, you could pass the following arguments: 'seq_1=1, seq_2=50, plot_2=c(5, 20)'.")
             }
    )
  }

  print(bode_patch)
}
```